\documentclass{article}

\usepackage{a4}
\usepackage{graphicx}
\usepackage{subfigure}
\RequirePackage{amsthm,amsmath,amssymb,amsfonts,enumerate}

\usepackage{makeidx}         
\usepackage{graphicx}        
\usepackage{multicol}        

\usepackage{amsmath}
\usepackage{amsfonts}
\usepackage{amsbsy}
\usepackage{graphics}
\usepackage{epsfig}
\usepackage{amssymb}
\usepackage{mathrsfs}
\usepackage{dsfont}

\usepackage{url}

\usepackage[dvipsnames, table]{xcolor}

\def\Px{\mathsf{P}}

\def\eqd{\stackrel{{d}}{=}}

\newcounter{savedfn}
\newcommand{\myfootnote}[1]{%
  \footnotemark                     
  \setcounter{savedfn}{\value{footnote}}%
  \footnotetext[\value{savedfn}]{#1}%
}
\newcommand{\myref}{%
  \protect\footnotemark[\number\value{savedfn}]%
}

\def\Arg{\mathrm{Arg}}
\def\TT{^\top}
\def\Pr{\mathsf{Prob}}
\def\RR{\mathds{R}}

\def\SX{{\mathscr X}}
\def\Ib{\mathbf{I}}
\def\xb{\mathbf{x}}
\def\Xb{\mathbf{X}}
\def\Ex{\mathsf{E}}

\def\simd{\stackrel{d}{\sim}}

\def\ml{\lambda}

\def\PP{\mathbb{P}}
\def\QQ{\mathbb{Q}}
\def\TT{\mathbb{T}}
\def\SB{{\mathscr B}}

\def\SJ{{\mathcal J}}
\def\SS{{\mathcal S}}
\def\0b{\mathbf{0}}
\def\dt{\widetilde{d}}
\newcommand{\bea}{\begin{eqnarray*}}
\newcommand{\eea}{\end{eqnarray*}}
\newcommand{\be}{\begin{eqnarray}}
\newcommand{\ee}{\end{eqnarray}}
\def\Gb{\mathbf{G}}
\def\Rb{\mathbf{R}}
\def\bs{\mathbf{s}}

\def\Zb{\mathbf{Z}}
\def\ub{\mathbf{u}}
\def\wb{\mathbf{w}}
\def\Yb{\mathbf{Y}}
\def\yb{\mathbf{y}}
\def\zb{\mathbf{z}}
\def\CR{\mathsf{CR}}
\def\PR{\mathsf{PR}}
\def\mg{\gamma}
\def\dd{\mbox{\rm d}}
\def\rad{\stackrel{\rm d}{\ra}}
\def\ra{\rightarrow}
\def\SN{{\mathcal N}}
\def\ma{\alpha}
\def\mab{\mathbf{\alpha}}
\def\deltab{\mathbf{\delta}}

\def\e1{\mathrm{e}}
\newcommand{\fin} {\mbox{}~\hfill{\lower-0.3ex\hbox{$\triangleleft$}}}

\def\SC{\mathscr{C}}

\newtheorem{theorem}{Theorem}[section]

\newtheorem{lemma}{Lemma}[section]

\newtheorem{definition}{Definition}[section]
\newtheorem{remark}{Remark}[section]

\begin{document}

\title{Sparse designs via quantization of the uniform measure on high-dimensional hypercubes}

\author{Luc Pronzato\footnote{(corresponding author) Laboratoire I3S, CNRS-Universit\'e C\^ote d'Azur, Sophia Antipolis, France, {\tt pronzato@i3s.unice.fr}} \ and Anatoly Zhigljavsky\footnote{School of Mathematics, Cardiff University, UK, {\tt ZhigljavskyAA@cardiff.ac.uk}}}

\date{\today}

\maketitle

\begin{abstract}
While the notions of sphere packing and sphere covering underlie the most widely used criteria for space-filling design in computer experiments, namely, the maximin and minimax approaches, they suffer from major drawbacks in high dimensions. First, when the dimension $d$ is very large, deploying a space-filling design requires an extremely large number of design points, making the very notion of a space-filling design questionable.
Second, these two design characteristics are highly sensitive to boundaries, particularly when the design domain is a hypercube (the case considered in this paper), and especially to its vertices. This sensitivity is particularly acute when the design size is less than $2^d$. Instead of considering only the minimum distance between pairs of design points for the maximin (or packing-radius) criterion, or the maximum distance from a point $\xb$ in the space to the design for the minimax (or covering-radius) criterion, we consider here the quantization error of the uniform measure on the hypercube. This error has three key advantages:
it accounts for the entire distribution of distances from a random point $U$ to the design, is less sensitive to boundary effects than the maximin and minimax criteria, and, thanks to Zador’s celebrated theorem, enables a smooth and natural transition to uniform designs as the design size tends to infinity. Building on recent results on the construction of optimal random quantizers for spherically symmetric distributions \cite{PZ2026b-submitted}, we propose a method to generate random designs of a given target size $n$ with small $L_s$-mean quantization error. The support of the resulting distribution is then used to construct designs, possibly incrementally, with sizes up to $n$. We demonstrate that, for moderate $n$, designs supported on the vertices of a small hypercube perform exceptionally well---and significantly better than designs uniformly distributed over the entire hypercube. Furthermore, we show that selecting these vertices through a greedy packing algorithm, using a two-level fractional factorial design as the candidate set, is highly beneficial. We also explain how to account for projections onto lower-dimensional subspaces.
\end{abstract}

{\bf keywords:} quantization, distortion, Zador's theorem, computer experiments


\section{Introduction}
\label{sec:intr}

Let $\mu$ be a probability measure on a compact subset $\SX$ of $\RR^d$ (with particular emphasis on the case where $\SX$ is the hypercube $[-1,1]^d$), and let $\Xb_n=\{\xb_1,\ldots,\xb_n\}\in\RR^{n\times d}$ be a design of size $n$, i.e., an $n$-point set in $\SX$. By high dimension, we mean the situation where $n$ is typically smaller than $2^d$, the number of vertices of the hypercube. 
For any $s>0$, the $L_s$-mean quantization error of $\mu$ for $\Xb_n$ is $E_{\mu,s}(\Xb_n)=\left(\Ex_\mu\{ \min_{i=1,\ldots,n} \|U-\xb_i\|^s \}\right)^{1/s}$, where $U\simd\mu$. When $\mu$ has the density $\varphi$ with respect to the Lebesgue measure, Zador's celebrated theorem \cite{Zador82} states that the empirical distribution of an $n$-point optimal quantizer $\Xb_n^*$ minimizing $E_{\mu,s}(\Xb_n)$ converges weakly, as $n\to\infty$, to the probability measure with density proportional to $\varphi^{d/(d+s)}$. It also characterizes the asymptotic value of $n^{1/d}E_{\mu,s}(\Xb_n^*)$. Note the connection with transport theory: for $s\geq1$, $E_{\mu,s}(\Xb_n^*)$ is the minimum {Kantorovich (or Wasserstein) distance of order $s$ between $\mu$ and a measure with at most $n$ support points; see \cite[Lemma~3.4]{GrafL2000}. Designs with small quantization error are particularly relevant for numerical integration; see \cite{Pages97}.

The construction of an $n$-point optimal quantizer $\Xb_n^*$ is a difficult task. While in low dimensions greedy algorithms can be used for the incremental construction of designs with small quantization error \cite{LuschgyP2015,NogalesPR2021}, they become computationally infeasible for large $d$. When $\mu$ is uniform on $\SX$, the points of an optimal quantizer are  asymptotically uniformly distributed for all $s>0$, and the asymptotic considerations above do not help with the construction of designs with small quantization error. However, these asymptotics are effective only when $n^{1/d}$ is sufficiently large, which in large dimension requires astronomically large values of $n$. In fact, for large $d$ and reasonable $n$, random designs uniformly distributed on specific subsets of $\SX$ can achieve exceptional performance. This is proved in \cite{PZ2026b-submitted} for the case when $\mu$ is spherically symmetric; in this paper we consider the case where $\mu$ is uniform on $[-1,1]^d$. 
Related work studies covering and quantization in high-dimensional cubes, including non-lattice constructions and nested designs \cite{ZhigljavskyNoonan2020,NoonanZhigljavsky2021}; high-dimensional exploration and convergence of global random search algorithms are considered in \cite{NoonanZhigljavsky2024}.

The paper is organized as follows. Section~\ref{S:distancecdt-etc} introduces the main notions and recalls some results for spherically symmetric distributions. Section~\ref{S:cube}, which contains the main results of the paper, addresses the case where $\SX$ is the $d$-dimensional hypercube $\SC_d$. In Section~\ref{S:approximations-cube}, we construct spherically symmetric approximations of measures on $\SC_d$; this yields an approximation of $E_{\mu,s}^s(\Xb_n)$ for random designs $\Xb_n$ (Section~\ref{S:approximate-Edistortion-cube}). Distributions minimizing this approximate criterion are studied in Section~\ref{S:optimal-spherically-symmetric approximation}, with numerical illustrations in Section~\ref{S:Example-1}; Section~\ref{S:quality-approximate-Edistortion} compares the approximate criterion with its numerically computed value, showing that it overestimates the true value in the examples considered. Section~\ref{S:designs-in-cube} demonstrates the value of our approach through numerical examples. The support of the optimal distribution corresponds to the vertices of a small hypercube in $\SC_d$; these can be sampled randomly (Sections~\ref{S:sampling-without-R} and \ref{S:sampling-fract-fact}) or via a deterministic greedy algorithm (Section~\ref{S:greedy-packing-fract-fact}). Designs of size exceeding the number of vertices are treated in Section~\ref{S:n>2^d}, and Section~\ref{S:projections} illustrates the necessary trade-off between quantization in the full hypercube $\SC_d$ and in its lower-dimensional projections. Section~\ref{S:conclusion} briefly concludes. The examples presented in the paper use the MATLAB scripts available at
\url{https://github.com/lpronzato-prog/random-quantizers-high-d-cube}.

We denote by $\SB_d(\xb,r)$ the $d$-dimensional (closed) Euclidean ball with centre $\xb$ and radius $r$; we write $\SB_d(r)=\SB_d(\0b_d,r)$, and $\SS_{d-1}(r)$ denotes its $(d-1)$-dimensional boundary sphere; $\|\cdot\|$ is the $\ell_2$-norm; $\ub\TT$ is the transpose of the column vector $\ub$; $\Ib_d$ is the $d$-dimensional identity matrix.
\section{Distance c.d.f., quantization error and distortion}\label{S:distancecdt-etc}

\subsection{Definitions and basic properties}

For any $n$-point design $\Xb_n$ we denote by $d(\cdot,\Xb_n)$ the distance function defined by $\xb\in\RR^d \mapsto d(\xb,\Xb_n)=\min_{\xb_i\in\Xb_n}\|\xb-\xb_i\|$.

\begin{definition}\label{D:distance-cdf}
For $\Xb_n$ a fixed $n$-point set in $\RR^d$ and $\mu$ a probability measure on $\RR^d$, the \emph{distance c.d.f.} $F(\cdot\,;\Xb_n,\mu)$ is the c.d.f.\ of the random variable $d(U,\Xb_n)$ where $U \simd\mu$:
\bea 
F(t;\Xb_n,\mu) = \mu\left\{U\in \bigcup_{i=1}^n \SB_d(\xb_i,t)\right\}  = \mu\{d(U,\Xb_n) \leq t\}\,.
\eea
For $\Px_n$ a probability measure  on $\RR^{n\times d}$, the \emph{mean distance c.d.f.} $F(\cdot\,; \Px_n,\mu)$ is defined by 
$F(t; \Px_n,\mu) = \Ex_{\Px_n}\{F(t;\Rb_n,\mu) \}$,
where $\Rb_n$ is a random $n$-point design generated with $\Px_n$.  
\end{definition}

If $\mu$ is the delta measure $\delta_\ub$ concentrated at $\ub\in\RR^d$, the mean distance c.d.f.\ is simply $F(t; \Px_n,\delta_\ub)=\Px_n \left\{ d(\ub,\Rb_n)\leq t  \right\}$.
When $\mu$ is equivalent to the Lebesgue measure on $\SX$, the essential supremum of the random variable $d(U,\Xb_n)$ is equal to the covering radius $\CR(\Xb_n)=\max_{\xb\in\SX} d(\xb,\Xb_n)$ and $F(\cdot\,;\Xb_n,\mu)$ has a density $f(\cdot\,;\Xb_n,\mu)$ on $[0,\CR(\Xb_n)]$. 
When $\mu$ is the uniform measure on $\SX$, the distance c.d.f.\ contains all information about the filling of $\SX$ by $\Xb_n$. The $L_s$-mean quantization error of $\mu$ for $\Xb_n$ \cite{GrafL2000}, or equivalently the $(\mu,s)$-distortion related to $\Xb_n$ \cite{Pages97}, provides a scalar summary of this information.

\begin{definition}\label{D:quantization-error}
The \emph{$L_s$-mean quantization error} of a probability measure $\mu$ on $\RR^d$ with finite $s$-th moment $\Ex\{\|X\|^s\}$ for a given $n$-point design $\Xb_n$ is the $L_s(\mu)$-mean (and, when $s\geq1$, the $L_s(\mu)$-norm) of the distance function $d(\cdot,\Xb_n)$:
\bea 
E_{\mu,s}(\Xb_n) = \|d(\cdot,\Xb_n)\|_{L_s} = \left[\int_{\RR^d} d^s(\xb,\Xb_n)\,\mu(\dd\xb)\right]^{1/s}\,, \quad s> 0\,.
\eea
When $\Xb_n\simd\Px_n$ is random, we define the expected $(\mu,s)$-distortion as $D_{\mu,s}(\Px_n)=\Ex_{\Px_n}\{D_{\mu,s}(\Xb_n)\}$, 
where $D_{\mu,s}(\Xb_n)$ is the $(\mu,s)$-distortion related to $\Xb_n$, defined by $D_{\mu,s}(\Xb_n)=E_{\mu,s}^s(\Xb_n)$. 
\end{definition}

For any $s>0$, the $(\mu,s)$-distortion satisfies (see, e.g., \cite[p.~150]{Feller71-2}):
\be
D_{\mu,s}(\Xb_n) = \Ex_\mu\{d^s(U,\Xb_n)\} \!\!\!\!  &=& \!\!\!\!  s\, \int_0^\infty t^{s-1}\,[1-F(t;\Xb_n,\mu)] \,\dd t \,, \label{moment-ds-cdf} \\
&=& \!\!\!\!  \int_0^\infty [1-F(t^{1/s};\Xb_n,\mu)] \,\dd t \,, \nonumber
\ee
showing that $D_{\mu,s}(\Xb_n)$ corresponds to the area between the curve $t\mapsto F(t^{1/s};\Xb_n,\mu)$ for $t\in\RR^+$ and the horizontal asymptote $y=1$. As a direct consequence of properties of $L_s$ norms, for any $n$-point design and probability measure $\mu$, $E_{\mu,s}(\Xb_n) \leq E_{\mu,s'}(\Xb_n)$ for $0<s<s'$ (equality is possible when $d(U,\Xb_n)$ is $\mu$-almost surely constant). When $\mu$ is equivalent to the Lebesgue measure on $\SX$, $\lim_{s\to\infty}E_{\mu,s}(\Xb_n)=\CR(\Xb_n)$.
Moreover, Lemma~10.1 in \cite{GrafL2000} indicates that, for any $n$-point optimal quantizer $\Xb_n^*$, $E_{\mu,s}(\Xb_n^*)\to \min_{\Xb_n\in\RR^{n\times d}}\CR(\Xb_n)$ as $s\to\infty$ for any fixed $n$, showing that, for large $s$, optimal $n$-quantizers are almost minimax-optimal.

\begin{remark}[Partial covering]
Although the paper focuses on the quantization error, one could also consider the quantiles of the distance c.d.f., leading to the notion of {\em partial covering}. Let $\mu$ be the uniform probability measure on $\SX$. When $\SX$ is connected and coincides with the closure of its interior, the density $f(\cdot\,;\Xb_n,\mu)$ is then strictly positive on $(0,\CR(\Xb_n))$ and the $\mg$-quantile $q_\mg(\Xb_n)$ of $F(\cdot\,;\Xb_n,\mu)$ is defined by
$F[q_\mg(\Xb_n);\Xb_n,\mu]=\mg \ \mbox{ for any }\mg\in(0,1)$,
with $q_{0}(\Xb_n)=0$ and $q_{1}(\Xb_n)=\CR(\Xb_n)$. Any $n$-point design $\Xb_n$ with covering quantile $q_\mg(\Xb_n)=h$ is such that the $n$ balls $\SB_d(\xb_i,h)$ cover $100\,\mg$ \% of $\SX$, i.e., they provide a partial covering of $\SX$. In contrast, the $n$ balls $\SB_d(\xb_i,\CR(\Xb_n))$ cover the entire $\SX$: when $\mg$ is close to 1, partial covering ignores parts of $\SX$ that may be difficult to cover (such as corners when $\SX$ is a hypercube) and focuses on the main bulk of $\SX$. 

However, quantiles are more challenging to handle than the $L_s$-mean quantization error. Moreover, when considering the distance c.d.f.\ for the uniform measure on a hypercube, we introduce an approximation involving a non-uniform spherically symmetric distribution (see Section~\ref{S:approximations-cube}), and covering quantiles are then less meaningful. 
\fin
\end{remark}

\subsection{Quantization error of random quantizers}

Consider the expected $(\mu,s)$-distortion $D_{\mu,s}(\Px_n)$ defined in Definition~\ref{D:quantization-error}, which we want to minimize with respect to $\Px_n$.  
From \eqref{moment-ds-cdf}, we can write
\be\label{Ls-mean-meandcdf}
D_{\mu,s}(\Px_n) =  s \int_{t\geq 0} t^{s-1}[1-F(t;\Px_n,\mu)]\,\dd t\,,
\ee
where $F(\cdot\,; \Px_n,\mu) = \Ex_{\Px_n}\{F(\cdot\,;\Xb_n,\mu) \}$ with $\Xb_n\simd\Px_n$ is the mean distance c.d.f.\ of Definition~\ref{D:distance-cdf}; that is, the c.d.f.\ of the random variable $d(U,\Xb_n)$ where both $U$ and $\Xb_n$ are random. 
Note that $\Ex_{\Px_n}\{ E_{\mu,s}(\Xb_n) \}\leq [D_{\mu,s}(\Px_n)]^{1/s}$ for $s \geq 1$ by Jensen's inequality, which provides an  upper bound on the expected quantization error\footnote{In practice, we generally have $\Ex_{\Px_n}\{ E_{\mu,s}(\Xb_n) \} \approx [D_{\mu,s}(\Px_n)]^{1/s}$; see Section~\ref{S:quality-approximate-Edistortion}.}. 

By swapping the order of expectations in $F(\cdot\,; \Px_n,\mu) = \Ex_{\Px_n}\{F(\cdot\,;\Xb_n,\mu) \}$, we get
\be
F(t; \Px_n,\mu)  =  \Ex_\mu  \left\{F(t; \Px_n,\delta_U) \right\}\,, \label{eq:prod_partial_cdf}
\ee
where $F(t; \Px_n,\delta_\ub) = \Px_n  \left\{ d(\ub,\Rb_n)\leq t  \right\}$. 
From now on, we only consider random quantizers with i.i.d.\ points having some distribution $\PP$, so that $\Px_n(\dd\xb_1,\ldots,\dd\xb_n)=\prod_{i=1}^n \PP(\dd\xb_i)$. We then denote $\Px_n=\PP^{[n]}$.

For any fixed $\ub  \in \RR^d$, any $t \geq 0$ and any $\Xb_n$ with $n$ i.i.d.\ points $\xb_i\simd\PP$, an elementary calculation (see \cite{PZ2026b-submitted}) shows that 
\be
F(t; \PP^{[n]},\delta_\ub) = 1-\bigg(1-\PP \left\{ \|X - \ub\| \leq t \right\} \bigg)^n\,,
\label{eq:prod_partial}
\ee
where $X \simd \PP$.
Calculating the mean distance c.d.f.\ $F(t; \PP^{[n]},\mu)$ through \eqref{eq:prod_partial_cdf} and then calculating the expected $(\mu,s)$-distortion $D_{\mu,s}(\PP^{[n]})$ through \eqref{Ls-mean-meandcdf} yields
\be\label{distortion-general}
D_{\mu,s}(\PP^{[n]}) =  s \int_{t\geq 0} t^{s-1} \Ex_\mu \left\{ \bigg(1-\PP \left\{ \| X-U \| \leq t \right\} \bigg)^n \right\} \,\dd t \,.
\ee

As shown in \cite{PZ2026b-submitted}, when $\PP$ is spherically symmetric, $D_{\mu,s}(\PP^{[n]})$ admits an explicit representation involving three integrals. This result is summarized in the next section. The case where $\mu$ is uniform on a hypercube requires some approximations and is addressed in Section~\ref{S:cube}. 

\subsection{$\PP$ is spherically symmetric}\label{S:spherically-symm}

Assume that $\PP$ is spherically symmetric. By definition, the random vector $X=(X_1, \ldots,X_d)\simd\PP$ can be represented as $X \eqd  R \cdot Z^{(d)} $, where $R=\|X\|$, $Z^{(d)}=(Z_1,\ldots,Z_d)$ is uniformly distributed on the unit sphere $\SS_{d-1}(1)$, and $R$ and $Z^{(d)}$ are independent. Let $\Phi(\cdot)$ denote the c.d.f.\ of $R=\|X\|$ and  $\beta_{a,b}$ denote a random variable with the beta density
\bea
\varphi_{a,b}(t)=t^{a-1}(1-t)^{b-1}/B(a,b) \ \mbox{ for } 0\leq t \leq 1\,,
\eea
where $a,b>0$ and $B(a,b)=\Gamma(a)\Gamma(b)/\Gamma(a+b)$ is the beta function. The c.d.f.\ of $\beta_{a,b}$ is $\Pr\{\beta_{a,b} \leq t\}=I_t(a,b)$, where $I_t(a,b)$ is the regularized incomplete beta function, for which we use the convention
\be\label{convention}
I_t(\cdot,\cdot)=   \left \{\begin{array}{ll}
    0 &\;\;  {\rm for}\;\; t\leq 0 \\
  1 & \;\; {\rm for}\;\; t \geq 1 \, .\\
  \end{array}
\right.
\ee

As $\PP$ is spherically symmetric, the distribution of $\|X  - \ub \|$ for $X\simd \PP$ only depends on $\|\ub\|$ and $d$ (see \cite[Prop.~2.1]{PZ2026b-submitted}): when $d\geq 2$, for any fixed $\ub \in \RR^d$ we have
\be
 \|X  - \ub \|^2 \eqd (\|\ub\|-R)^2 +4\,\|\ub\|\,R \, \beta_{\dt,\dt}\, , \label{eq:quantile2}
\ee
where $\dt=(d-1)/2$ and the random variables $R=\|X \|$ and $\beta_{\dt,\dt}$ are independent. After some calculation, this gives
\be
D_{\mu,s}(\PP^{[n]}) = s \int_{t\geq 0} t^{s-1} \int_{r\geq 0} H(r,t;\Phi,n)\, \dd\Psi(r)\,\dd t \,, \label{Ls-mean-final}
\ee
where
\be \label{H}
H(r,t;\Phi,n) =\bigg(1-\int_{\rho\geq 0} I_\upsilon(\dt,\dt) \,\dd\Phi(\rho ) \bigg)^n 
\ee
with 
\be\label{v}
\upsilon=\upsilon(t,\rho,r)=[t^2 -(\rho-r)^2]/(4 \rho r) \,,
\ee
and where $\Psi(\cdot)$ is the c.d.f.\ of $\|U\|$ for $U\simd\mu$. Here $\mu$ does not need to be spherically symmetric, but $\Psi(\cdot)$ can be challenging to derive in general cases. For that reason, in Section~\ref{S:approximations-cube} we use a spherically symmetric approximation of $\mu$ uniform on $[-1,1]^d$. 

\section{Quantization error in $d$-dimensional hypercubes} \label{S:cube}

Without loss of generality, we restrict our attention to the unit cube $\SC_d=[-1,1]^d$. The case of an arbitrary cube can be addressed by applying a suitable linear rescaling. For any $\delta>0$, we denote by $\delta\,\SC_d$ the hypercube $[-\delta,\delta]^d$. We only consider the case where the measure $\mu$ is uniform on $\SC_d$.

\subsection{Spherically symmetric approximations of measures in $\SC_d$}\label{S:approximations-cube}

\subsubsection*{A family of product measures in $\SC_d$}
Product measures are well-suited for working in $\SC_d$, and we consider the family of symmetric distributions defined by the product measures formed from symmetric Beta$_\delta(\alpha,\alpha)$ distributions on $[-\delta, \delta]$ with $\alpha>0$ and $\delta\in(0,1]$, together with their limiting cases $\alpha=0$ and $\delta=0$. For $\alpha>0$, $\delta>0$, and $\xb=(x_1, \ldots, x_d)$, $\PP(\dd\xb)=\PP_{\alpha,\delta}(\dd\xb)= \prod_{i=1}^d p_{\alpha,\delta}(x_i)\,\dd x_i$,
where
\bea
p_{\alpha,\delta}(t)= \frac{(2\delta)^{1-2\alpha}}{\mbox{\rm B$(\alpha,\alpha)$}} [\delta^2-t^2]^{\alpha-1}\, , \;\;-\delta\leq t\leq\delta\, .
\eea
If $\Xb_n\simd\PP_{\alpha,\delta}^{[n]}$, then each point $\xb_i\in\Xb_n$ has $d$ i.i.d.\ coordinates with density $p_{\alpha,\delta}(\cdot)$.
The case $\alpha=0$ corresponds to a sampling scheme where each coordinate of $\xb_i \in \Xb_n$ takes the value $\pm \delta$ with probability 1/2. For practical applications, to avoid duplicate points, we shall assume that $n\leq 2^d$ and use sampling without replacement; other, more efficient sampling schemes will also be considered; see Section~\ref{S:designs-in-cube}.
The choice $\alpha=\delta=1$ corresponds to independent uniform sampling in $\SC_d$, with $\PP_{1,1}=\mu$. For fixed $d$ and $n\to\infty$, Zador’s theorem implies that the empirical measure of optimal quantizers converges to the uniform distribution. This concerns optimal quantizers; it does not by itself establish the asymptotic optimality of i.i.d.\ uniform random designs\footnote{For a uniform target measure, uniform sampling is asymptotically optimal within the class of i.i.d.\ sampling distributions, but not among all sequences of designs.}. For $\ma>1$, $p_{\alpha,\delta}(-\delta)=p_{\alpha,\delta}(\delta)=0$ and $\PP_{\alpha,\delta}$ places less mass near the boundary of $\SC_d$, so we restrict attention to $\ma\in[0,1]$.

Our objective is to minimize $D_{\mu,s}(\PP_{\alpha,\delta}^{[n]})$ with respect to $\alpha$ and $\delta$, with $\ma,\delta\in[0,1]$. Although this expected distortion is not available in closed form, it can be approximated using the results of Section~\ref{S:spherically-symm}. To do so, we first need to construct a spherically symmetric approximation of $\PP_{\alpha,\delta}$ (and in particular of $\PP_{1,1}=\mu$). 

\subsubsection*{Spherically symmetric substitutes}

Let $\ml_{SO(d)}$ denote the Haar probability measure on the special orthogonal group $SO(d)$ of rotations in $\RR^d$.  For any probability measure $\PP$ on $\SC_d$, consider the product measure $\PP\otimes \ml_{SO(d)}$ on $\SC_d \times SO(d)$ and 
the symmetrized measure $\overline \PP$ given by the pushforward of 
$\PP\otimes \ml_{SO(d)}$ under the transformation  $T(\Rb,\xb)=\Rb \xb$, $\Rb\in SO(d)$, $\xb\in\RR^d$:
\be\label{symmetrization}
\overline \PP = T_{\#} \left( \PP\otimes \ml_{SO(d)} \right) \,.
\ee

This spherical symmetrization applied to any $\PP_{\ma,\delta}$ yields a spherically symmetric measure $\overline\PP_{\ma,\delta}$. However, $\overline\PP_{\ma,\delta}$ is difficult to determine for general $\ma,\delta$ in $[0,1]$, and below we consider an approximation $\QQ_{\ma,\delta}$ of $\overline\PP_{\ma,\delta}$, based on moment-matching for $\|X-\ub\|^{2\,j}$, with $j=1,2,3$ and $\ub$ an arbitrary vector in $\RR^d$. It relies on the following result. 

\begin{theorem}{\rm \cite[Ths.~1.2 \& 1.3]{noonanhigh2024}}\label{th:third}
Assume that  $d\geq 3$ and  $X =(X_1, \ldots,X_d) \in \RR^d$  is a  random vector satisfying either of the two models:

(\textit{i})  the components $X_i$ of $X $ are i.i.d.\ with a symmetric distribution having finite moments $\Ex\{X_1^\ell \}$ for $\ell  \leq 6$;

(\textit{ii}) $X$ has a  spherically symmetric distribution with $\Ex\{\|X \|^6\}<\infty$.

Then, for any $\ub\in\RR^d$, we have:
\bea
\Ex \{\|X  - \ub \|^2\}  &=& \Ex \{\|X  \|^2\} +  \| \ub \|^2\, , \\
\Ex \{\|X  - \ub \|^4\} &=&
\Ex \{\|X  \|^4\} +\|\ub \|^4 +\smash{\left(2+4/{d}\right)}\; \|\ub \|^2\, \Ex\{\|X  \|^2\}  \, ,
\\
\Ex\{\|X -\ub\|^6\} &=& \Ex\{\|X \|^6\}+ \|\ub\|^6 \\
&& + \left(3+12/d\right) \left[  \|\ub\|^2\, \Ex\{\|X \|^4\} + \|\ub\|^4\,  \Ex\{\|X \|^2\} \right]\,.
\eea
\end{theorem}

The assumptions $d \geq 3$, $\Ex\{X_1^6 \}<\infty$ and $\Ex\{\|X \|^6\}<\infty$ in Theorem~\ref{th:third} are required for the validity of the expression of $\Ex\{\|X -\ub\|^6\}$, but the expressions for $\Ex\{\|X -\ub\|^{2}\}$ and $\Ex\{\|X -\ub\|^{4}\}$ are valid for all $d \geq 1$ and $d \geq 2$, respectively, when the requirements on the moments are relaxed accordingly.
Expressions for $\Ex\{\|X -\ub\|^8\}$ are also available for product and spherically symmetric measures, but they do not coincide; see \cite[Th.~ 1.4]{noonanhigh2024}.

When $X\simd\PP_{\ma,\delta}$, the first three moments of $\|X\|^2$ are known explicitly from the formulas for the moments of beta-distributed random variables; they are given in the following lemma. 

\begin{lemma} \label{lem:5a2}
Let  $d\geq 3$ and  $X\simd \PP_{\alpha,\delta}$. Then
\bea
\Ex\{\|X  \|^2\} &=& d \nu_2 \,, \\
\Ex\{\|X  \|^4\} &=&  d[\nu_4+(d\!-\!1)\nu_2^2]\,,\\
\Ex\{\|X  \|^6\} &=&  d[ \nu_6+ 3(d\!-\!1) \nu_2 \nu_4 \!+\! (d\!-\!1)(d\!-\!2)\nu_2^3]\,,
\eea
where
$\nu_2 ={\delta^2}/({1+2\alpha})$, 
$\nu_4 = 3{\delta}^{4}/ [\left( 1+2\,{\alpha} \right)
 \left( 2\,{\alpha}+3 \right) ]$, and
 $\nu_6 =   15 \delta^{6} / [ ( 1+2{\alpha} )  ( 2{\alpha}+3 )  ( 2{\alpha}+5 )  ]$.
\end{lemma}

Our family of spherically symmetric approximations is constructed as follows. Let $\|X\|^2$ have the beta density $\varphi(\cdot\,;a,b|M)$ on the interval $(0,M)$,
\be \label{eq:beta_general}
\varphi(t;a,b|M)= \frac{t^{a-1}(M\!-\!t)^{b-1}}{M^{a+b-1}B(a,b)}\, \ t \in (0, M)\,, \ a,b >0\,,
\ee
where $B(\cdot, \cdot)$ is the beta function. 
The first three moments of $\|X\|^2$ are known in closed form. Matching them with those of $\PP_{\ma,\delta}$ yields explicit expressions for $a,b$ and $M$ in terms of $\ma$ and $\delta$:
\begin{subequations} \label{eq:parameters_alpha_delta}
  \be
a_* \!\!\!&=& \!\!\!
d\, \frac{4(d+2)\alpha^2 + 4(4d- 1)\alpha + 3(5d - 4)}{8\alpha^2(d+ 2)+ 8\alpha(4d- 1) +  6d}  \,,
\\
M_* \!\!\!&=& \!\!\! \delta^2\, \frac{4(d + 2)\alpha^2 + 4(4d - 1)\alpha +  3d }{3(2\alpha + 1)^2}\,,
\\
b_* \!\!\!&=& \!\!\! a_* \left( \frac{M_*(2\alpha+1)}{\delta^2 d}-1 \right) \,.
\ee
\end{subequations}
We denote by $\QQ_{\ma,\delta}$ the corresponding spherically symmetric approximation of $\PP_{\ma,\delta}$. 
The density of $\|X\|$ for $\QQ_{\ma,\delta}$ is 
$\phi_{\ma,\delta}(t) = {2t}\varphi(t^2;{a_*,b_*|M_*})$, $t\in [0, \sqrt{M_*}]$, with $\varphi(\cdot\,;a,b|M)$ given by \eqref{eq:beta_general}.
Note that $a_*$ and $b_*$ do not depend on $\delta$, which only affects the support of $\phi_{\ma,\delta}(\cdot)$. 

The left panel of Figure~\ref{F:d20_beta} shows the evolutions of $a_*$, $b_*$ and $M_*$ with $d$ for two choices of $\alpha$ and $\delta$; the right panel plots $\phi_{\ma,\delta}(t)$ for the same choices of $\alpha$ and $\delta$ when $d=20$. 
When $\ma=0$, $a_*=(5d-4)/2$, $M_*=d \delta^2$ and $b_*=0$, so that the distribution of $\|X\|$ for $\QQ_{0,\delta}$ has the Dirac delta measure concentrated at $\sqrt{M_*}=\delta \sqrt{d}$. Note that $a_*$ is decreasing in $\ma$ for all fixed $d\geq 2$, with $\lim_{\ma\to\infty} a_*=d/2$, so that $a_*>d/2$ for all $\ma\geq 0$. Also, for any $d\geq 2$, $b_*=0$ for $\ma=0$ and $b_*>1$ for $\ma=1$, and there exists $\ma^*\in(0,1)$ such that $b_*=1$ for $\ma=\ma^*$. For $a_*>1/2$ and $b_*>1$, $\phi_{\ma,\delta}(t)$ is maximum at $\widehat t=\sqrt{M_*(2a_*-1)/[2(a_*+b_*)-3]}$ (which is increasing in $a_*$ and decreasing in $b_*$).

This moment matching approach applied to a mixture $\PP(\wb,\mab,\deltab)=\sum_{i=1}^q w_i \PP_{\ma_i,\delta_i}$ (where $\mab=(\ma_1,\ldots,\ma_q)$, $\deltab=(\delta_1,\ldots,\delta_q)$ and $\wb=(w_1,\ldots,w_q)$, with $w_i\geq 0$ and $\sum_{i=1}^q w_i=1$) yields the mixture $\QQ(\wb,\mab,\deltab)=\sum_{i=1}^q w_i \QQ_{\ma_i,\delta_i}$. For general $\ma,\delta$ in $[0,1]$, $\QQ_{\ma,\delta}$ does not coincide with the symmetrized measure $\overline \PP_{\ma,\delta}$ defined by \eqref{symmetrization}. However, for any $\delta>0$, we have 
$\QQ_{0,\delta} = \overline {\PP_{0,\delta}}$, and this property extends to mixtures: $\QQ(\wb,\0b_q,\deltab) = \overline {\PP(\wb,\0b_q,\deltab)}$.

\subsubsection*{Spherically symmetric approximation of $\PP_{1,1}=\mu$}
A case of particular importance is when $\alpha=\delta=1$, for which $\PP_{1,1}=\mu$, uniform on $\SC_d$. We then have
\bea
\Ex\{\|X \|^2\} = \frac{d}{3}\, ;\;\;
\Ex\{\|X \|^4\} =  \frac{d^2}{9}+ \frac{4d}{45} ;\;\;
\Ex\{\|X \|^6\} = \frac{d^3}{27} +{\frac {4\,{d}^{2}}{45}}+ {\frac {16\,d}{945}}\,,
\eea
and we approximate $\PP_{1,1}$ by a spherically symmetric $\mathbb{Q}_{1,1}$ such that $\|X\|^2$ has a density of the form \eqref{eq:beta_general}. Denoting by $\phi_{1,1}(\cdot)$ the density of $\|X\|$, we get 
\be\label{phi11}
\phi_{1,1}(t) = {2t}\varphi(t^2;{a_d,b_d|M_d})\,, \ t\in [0, \sqrt{M_d}]\,, 
\ee
where
\be \label{eq: matching three moments}
  a_d = {\frac {d \left( 35d\!-\!8 \right) }{2(23d\!+\!4)}}\,, \ 
  b_d={
\frac { \left( 7d\!+\!2 \right)  \left( 35d\!-\!8 \right) }{9(23d\!+\!4)}}\,, \
M_d=
{\frac {23\,d+4}{27 }} \,.
\ee
These three parameters increase with $d$ and satisfy $a_d>1/2$, $b_d>1$ and $M_d>1$, so that $\phi_{1,1}(\cdot)$ is maximum at $\widehat t=\sqrt{d/3+4/105}$. 

\begin{figure}[ht!]
\begin{center}
\includegraphics[width=.49\linewidth]{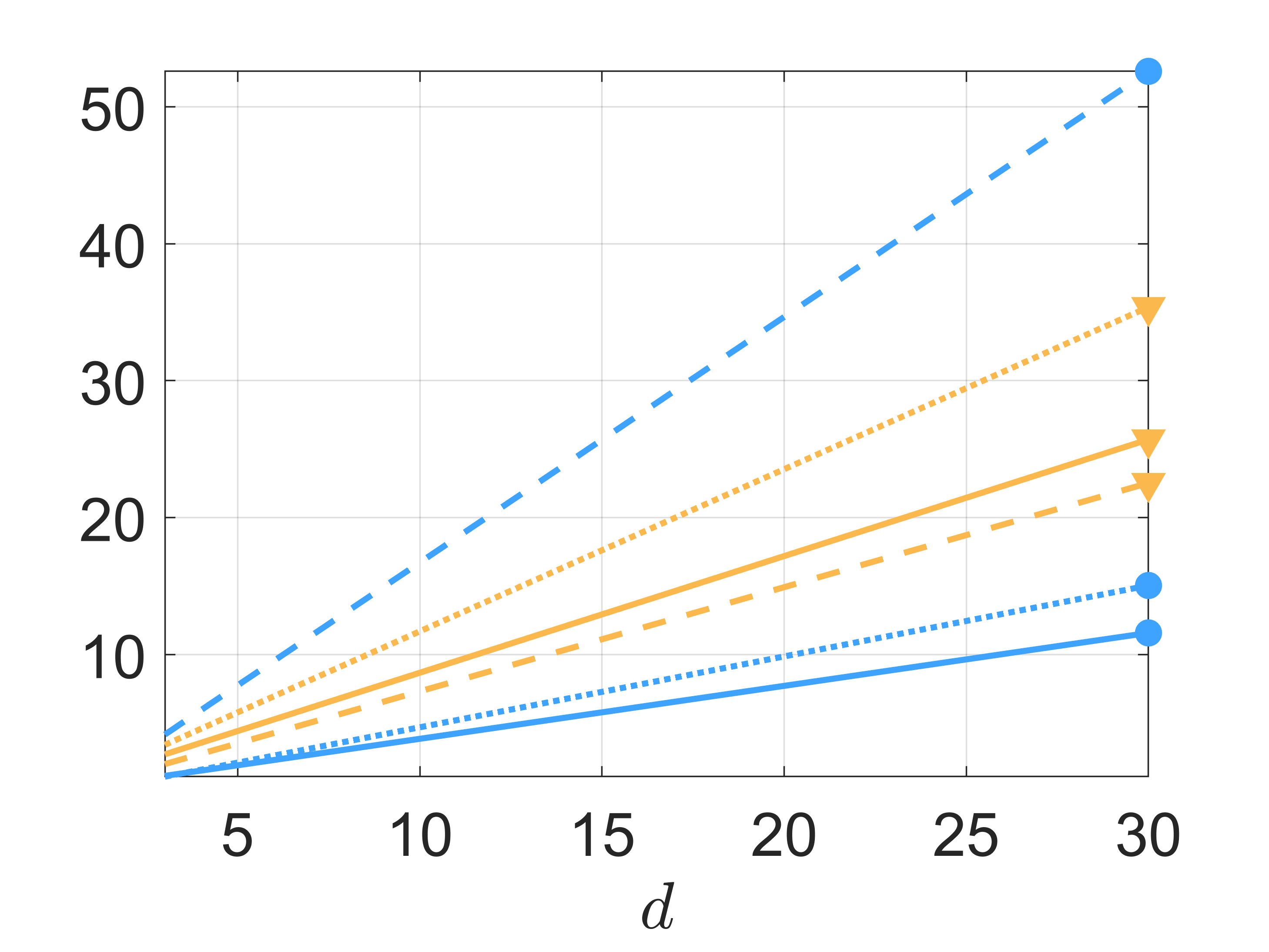} \includegraphics[width=.49\linewidth]{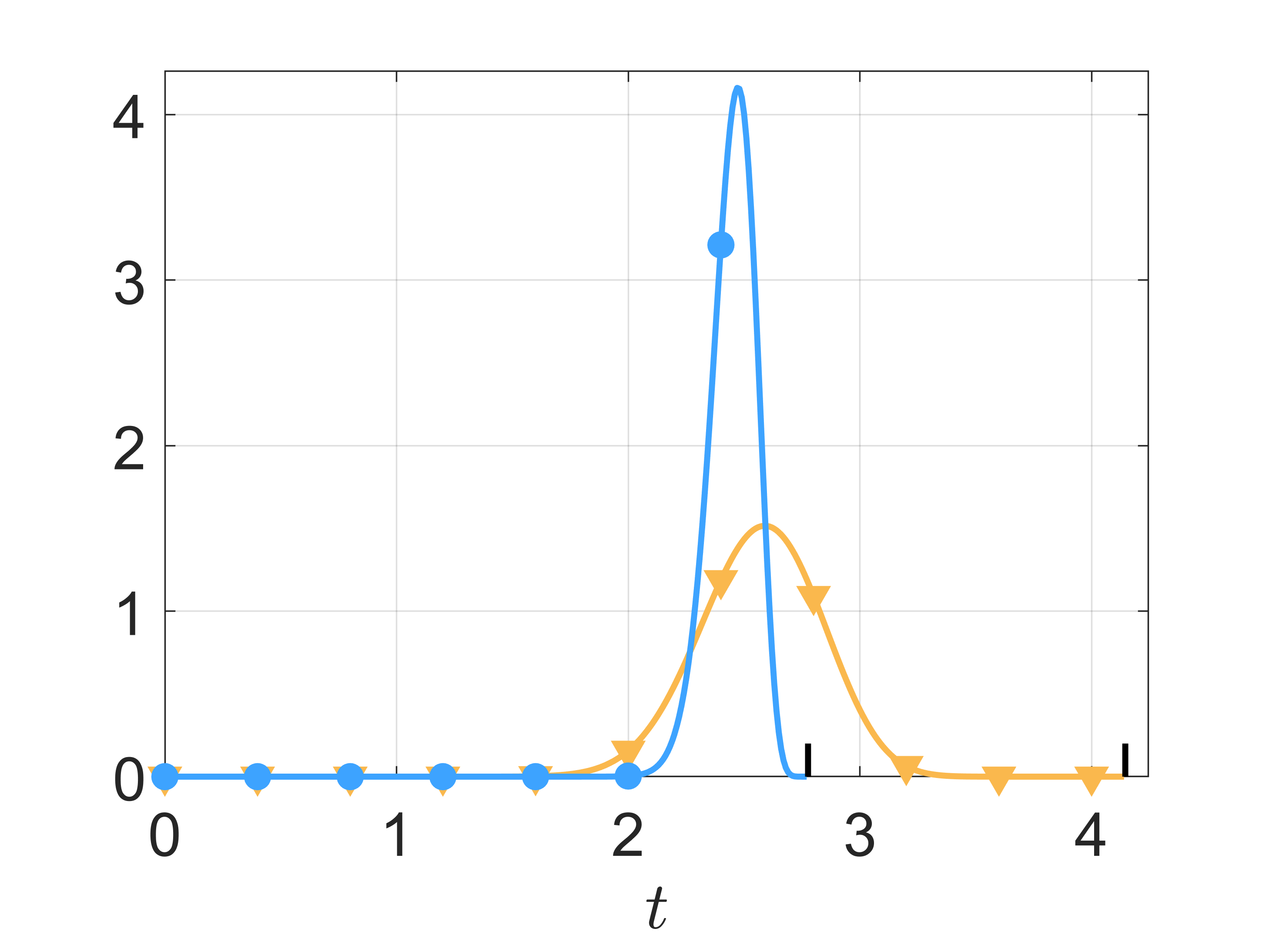}
\end{center}
\caption{\small Left: $a_*$ ({\bf - - -}), $b_*$ ({\bf $\cdots$}) and $M_*$ ({\bf ---}) as functions of $d$ for $\alpha=1$, $\delta=1$ ({\color{Dandelion} $\blacktriangledown$}) and $\alpha=0.1$, $\delta=0.6$  ({\color{Cerulean} $\bullet$}). Right: $\phi_{\ma,\delta}(\cdot)$ for $d=20$ and $\alpha=1$, $\delta=1$ (i.e.,  $\PP_{\ma,\delta}=\PP_{1,1}=\mu$, {\color{Dandelion} $\blacktriangledown${\bf ---}$\blacktriangledown$}) and $\alpha=0.1$, $\delta=0.6$  ({\color{Cerulean} $\bullet${\bf ---}$\bullet$}); the right limit $\sqrt{M_*}$ of the support is indicated.}
\label{F:d20_beta}
\end{figure}

Explicit expressions for the densities of $\| X\|^2$ when $X\simd\PP_{1,1}=\mu$ in $\SC_d$ are derived in \cite{weissman2017sum} for $d=2,3,4$, but for $d\geq 5$ such explicit expressions remain unknown. Nevertheless, the asymptotic behavior can be easily derived. By the Central Limit Theorem (CLT), we have
\be
\frac{1}{\sqrt{d}} \left(\|X\|^2-\frac{d}{3}\right) \rad \SN\left(0,\frac{4}{45}\right)\,, \quad d\to\infty\,, \label{CLT-normX2-cube} 
\ee
and from Hoeffding's inequality $\Pr\left\{\left|\|X\|^2-\frac{d}{3}\right|>t\right\} \leq 2\,\exp\left(-\frac{2\,t^2}{d}\right)$, showing that, for large $d$, the norm of a random point $X \simd \mu$ tends to be close to $\sqrt{d/3}$. The asymptotic distribution \eqref{CLT-normX2-cube} is already accurate for moderate $d$, the approximation $\varphi(\cdot;a_d,b_d|M_d)$, which matches exactly the first three moments of $\| X\|^2$, is even more precise and the accuracy improves when $d$ increases. Figure~\ref{F:densities-normX2} gives an illustration. 

\begin{figure}[ht!]
\begin{center}
\includegraphics[width=.49\linewidth]{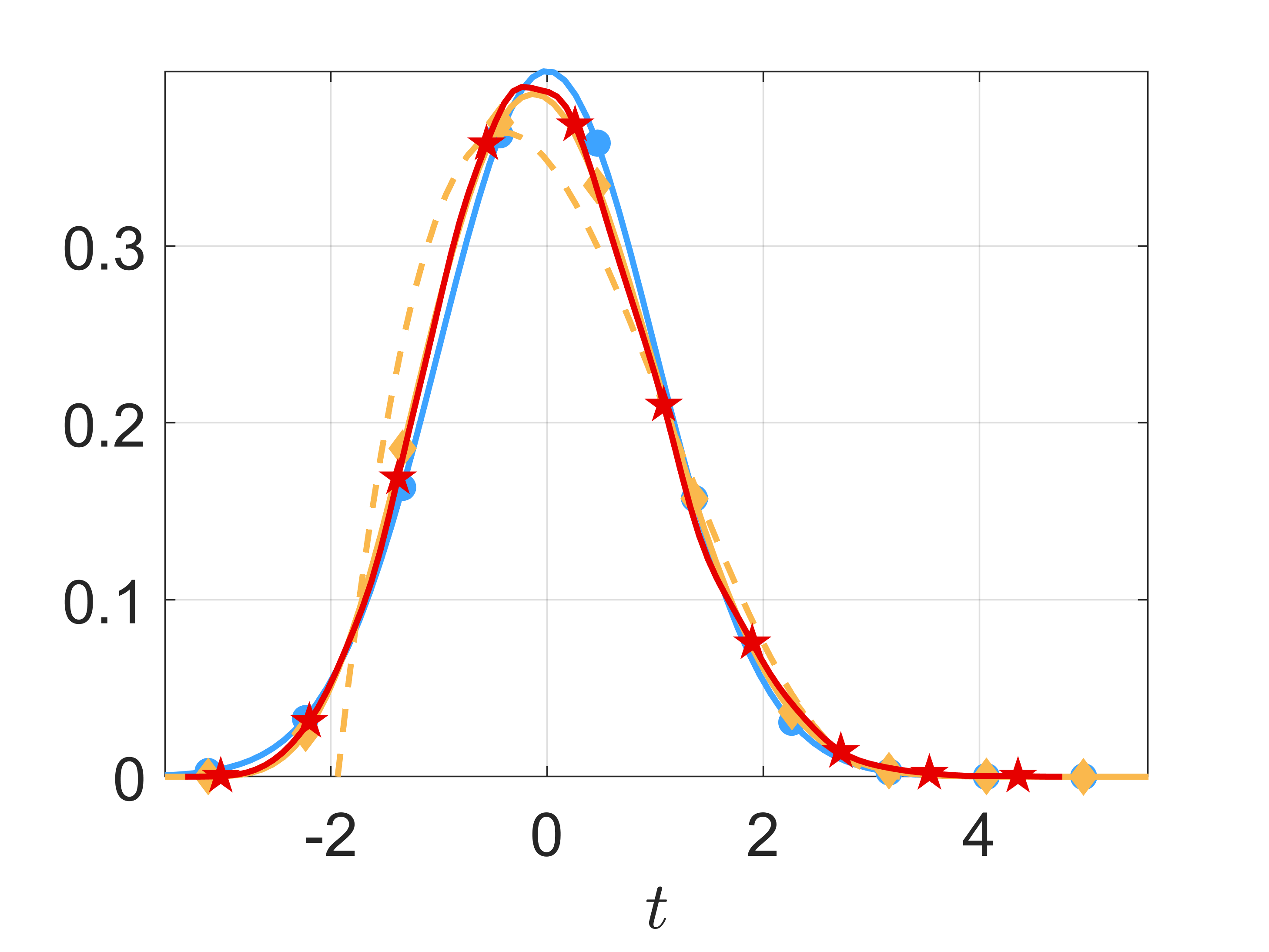} \includegraphics[width=.49\linewidth]{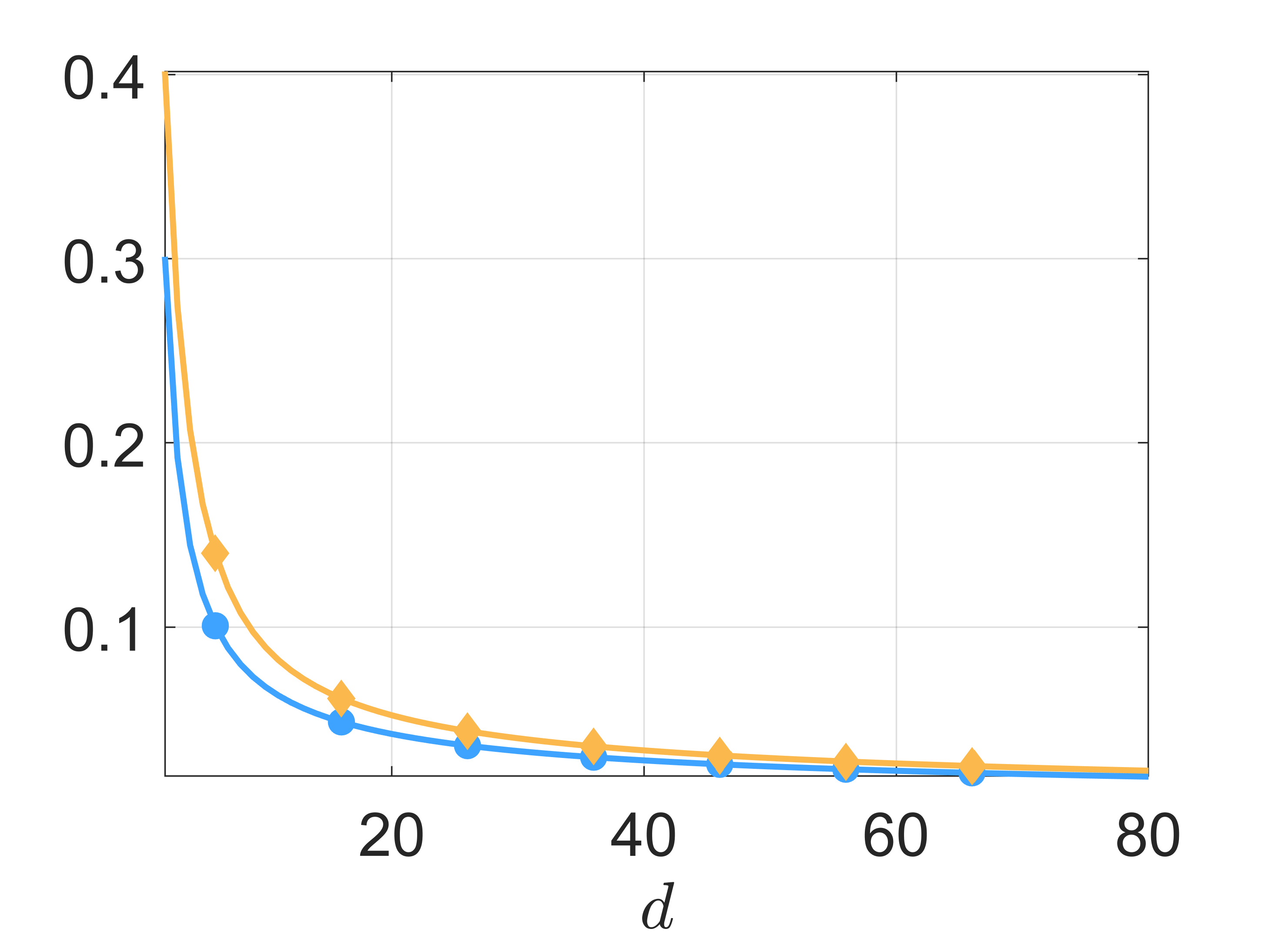}
\end{center}
\caption{\small Left: density of the standard normal distribution $\SN(0,1)$ ({\color{Cerulean} $\bullet$}); empirical (smoothed) density of $\sqrt{45}/(2\sqrt{d}) (\|X\|^2-d/3)$ for $d=10$, based on 10\,000 samples $X_i$ i.i.d.\ with $\mu$ ({\color{red} $\bigstar${\bf ---}$\bigstar$}); renormalized density $2\sqrt{d/45}\, \varphi(2\sqrt{d/45}\,t+d/3;a_d,b_d|M_d)$ for $d=10$ ({\color{Dandelion} $\blacklozenge${\bf ---}$\blacklozenge$}) and $d=3$ ({\color{Dandelion} {\bf - - -}}). Right: distance in variation ({\color{Cerulean} $\bullet$}) and Hellinger distance ({\color{Dandelion} $\blacklozenge$}) between the beta distribution with density $\varphi(\cdot;a_d,b_d|M_d)$ and the normal distribution $\SN(d/3,4d/45)$ as functions of $d$.}
\label{F:densities-normX2}
\end{figure}

\subsection{Approximation of the expected $(\mu,s)$-distortion $D_{\mu,s}(\PP_{\alpha,\delta}^{[n]})$}\label{S:approximate-Edistortion-cube}

The approximation is constructed in two steps. First, we replace $\PP_{\alpha,\delta}$ with its spherically symmetric approximation $\QQ_{\alpha,\delta}$ and compute the expected distortion $D_{\mu,s}(\QQ_{\alpha,\delta}^{[n]})$ as in \eqref{Ls-mean-final}. Second, we substitute $\QQ_{1,1}$ for $\mu$, i.e., we replace the c.d.f.\ $\Psi(\cdot)$ in \eqref{Ls-mean-final} with the c.d.f.\ corresponding to the density $\phi_{1,1}(\cdot)$ given by \eqref{phi11}. 
We denote the resulting approximation by
\be \label{Phi_alpha_delta}
\widehat D_{\mu,s}(\PP_{\alpha,\delta}^{[n]}) =  D_{\QQ_{1,1},s}(\QQ_{\alpha,\delta}^{[n]})\,,
\ee
which we can then minimize with respect to $\ma$ and $\delta$. 

In general, the calculation of 
$\widehat D_{\mu,s}(\PP_{\alpha,\delta}^{[n]})$ requires the evaluation of three nested integrals. The inner one corresponds to $\int_{\rho\geq 0} I_\upsilon(\dt,\dt) \,\phi_{\ma,\delta}(\rho )\,\dd\rho$, see \eqref{H}, 
for which $\rho$ varies in $[0,\sqrt{M_*}]$ with $M_*$ given by {\rm (\ref{eq:parameters_alpha_delta}b)}. The expectation with respect to $r$ in \eqref{Ls-mean-final} is for $r\in[0,\sqrt{M_d}]$, see \eqref{eq: matching three moments}. For the third integral
\eqref{Phi_alpha_delta}, we can restrict $t$ to $[0,t_*]$ with  $t_*$ such that 
$\int_0^{\sqrt{M_*}} I_\upsilon(\dt,\dt) \,\phi_{\ma,\delta}(\rho )\,\dd\rho=1$
for all $r\in[0,\sqrt{M_d}]$. From the convention \eqref{convention} used for $I_\upsilon(\dt,\dt)$, it is enough to have $\upsilon=\upsilon(t_*,\rho,r)\geq 1$ for all $\rho$ and $r$, and the expression \eqref{v} for $\upsilon(t,\rho,r)$ shows that $t_*=\sqrt{M_*}+\sqrt{M_d}$ is suitable. 

Let $\ma^*(d,n,s)$ and $\delta^*(d,n,s)$ denote the optimal values of $\ma$ and $\delta$ that minimize $\widehat D_{\mu,s}(\PP_{\alpha,\delta}^{[n]})$ for given values of $d,n$ and $s$. Their determination corresponds to a two-dimensional optimization problem, which can be solved numerically. 
At the same time, since $\QQ_{\alpha,\delta}$ is spherically symmetric, we can exploit results from \cite{PZ2026b-submitted} on optimal random quantizers for spherically symmetric distributions. This is the topic of the next section.

\subsection{Optimal spherically symmetric approximation}\label{S:optimal-spherically-symmetric approximation}

Consider the minimization of $D_{\QQ_{1,1},s}(\QQ^{[n]})$ with respect to a probability measure $\QQ$ on $\RR^d$. The expression \eqref{distortion-general} shows that $D_{\QQ_{1,1},s}(\QQ^{[n]})$ is convex in $\QQ$. As $\QQ_{1,1}$ is spherically symmetric and $D_{\QQ_{1,1},s}(\QQ^{[n]})$ is invariant under rotation of $\QQ$, by a standard symmetrization argument there exists an optimal $\QQ^*$ that is spherically symmetric. In \cite{PZ2026b-submitted}, a necessary and sufficient condition for the optimality of a spherically symmetric measure $\QQ^*$ is derived, and it is shown that the support of $\QQ^*$ is included in that of $\QQ_{1,1}$, here $\SB_d(\sqrt{M_d})$ with $M_d$ defined in \eqref{eq: matching three moments}. 
For any finite $n$, the optimal measure $\QQ^*$ is a mixture of uniform measures on concentric spheres with radii $R_i$. For fixed $d$ and $s$, the number of spheres increases (very) slowly with $n$; when $n\to \infty$, $\QQ^*$ increasingly resembles $\QQ_{1,1}$, in agreement with Zador's theorem. For $n$ small enough, $\QQ^*=\QQ_{R^*}$ uniform on a single sphere $\SS_{d-1}(R^*)$, with $R^*=R^*(d,n,s)$.
The necessary and sufficient condition for the optimality of $\QQ_{R^*}$ takes a simple form, which we formulate in the theorem below.

\begin{theorem}{\rm \cite[Corollary~3.1]{PZ2026b-submitted}}\label{Th:CNS} For any $d\geq 3$, $n\geq 1$ and $s>0$, the measure $\QQ_{R^*}$, supported in $\SS_{d-1}(R^*)$, minimizes $D_{\QQ_{1,1},s}(\QQ^{[n]})$ among all distributions on $\RR^d$ if and only if
\bea
S(\rho;R^* ,n,s)\leq
S(R^*;R^*,n,s)
\quad\hbox{for all }\rho\in[0,\sqrt{M_d}]\,,
\eea
with $S(\cdot\,;R,n,s)$ the sensitivity function defined by
\bea
S(\rho;R,n,s) = ns \int_{t\geq 0} t^{s-1}\int_{r\geq 0}
\left[1-I_{\upsilon(t,R,r)}(\dt,\dt)\right]^{n-1}\,I_{\upsilon(t,\rho,r)}(\dt,\dt)\,\phi_{1,1}(r)\,\dd r\,\dd t \,,
\eea
where $\upsilon(t,\rho,r)$ is defined by \eqref{v}, $\phi_{1,1}(\cdot)$ is defined by \eqref{phi11}, $I_\upsilon(\dt,\dt)$ is the regularized incomplete beta function with $\dt=(d-1)/2$ and $M_d$ is given by \eqref{eq: matching three moments}.
\end{theorem}

The left panel of Figure~\ref{F:sensitivity} plots $S(\rho;R^*,n,2)$ as a function of $\rho\in[0,\sqrt{M_d}]$ for $d=10$ and $n=1\,000$, where $R^* \simeq 1.5101$. 

\begin{figure}[ht!]
\begin{center}
\includegraphics[width=.49\linewidth]{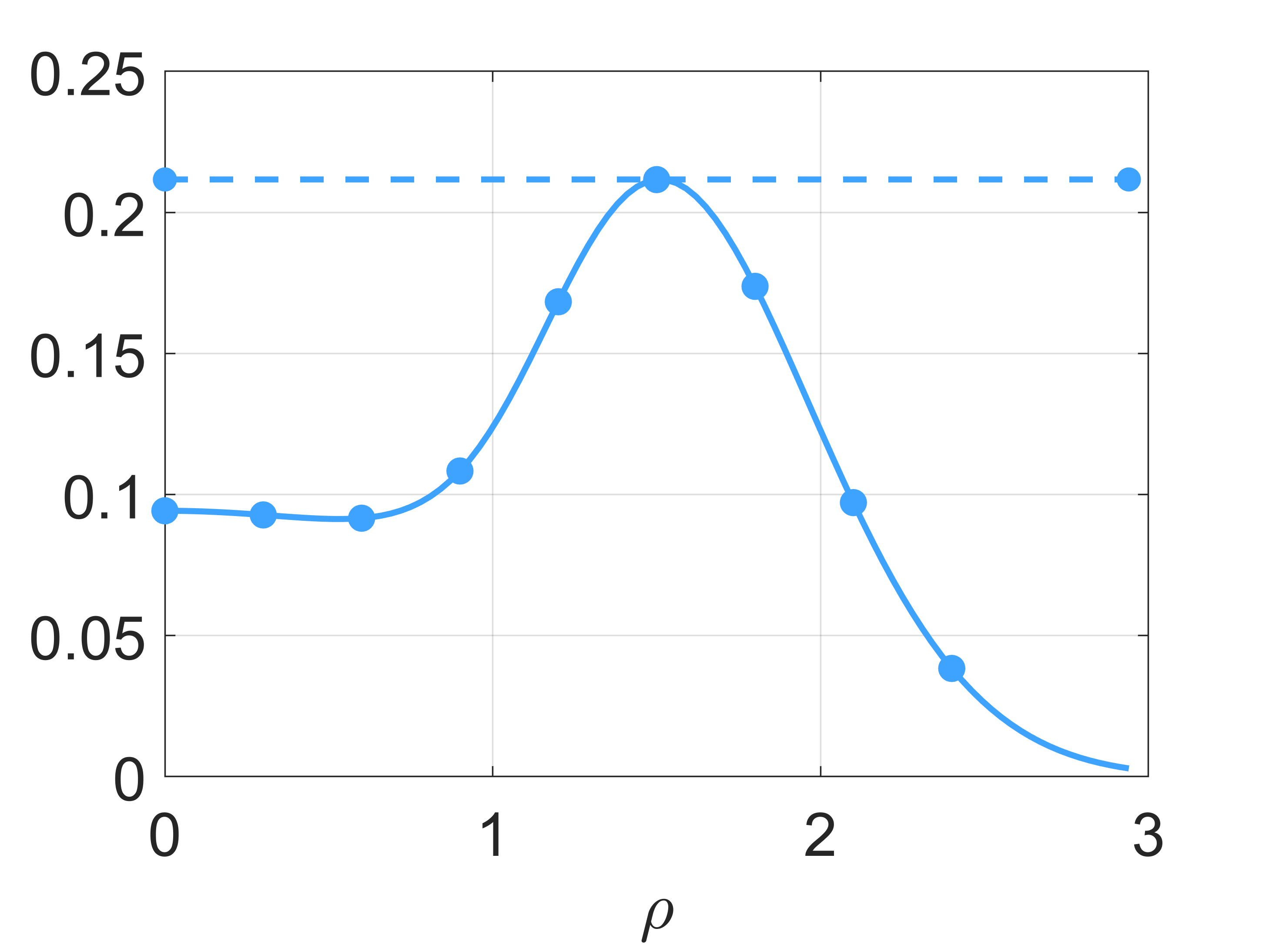}
\includegraphics[width=.49\linewidth]{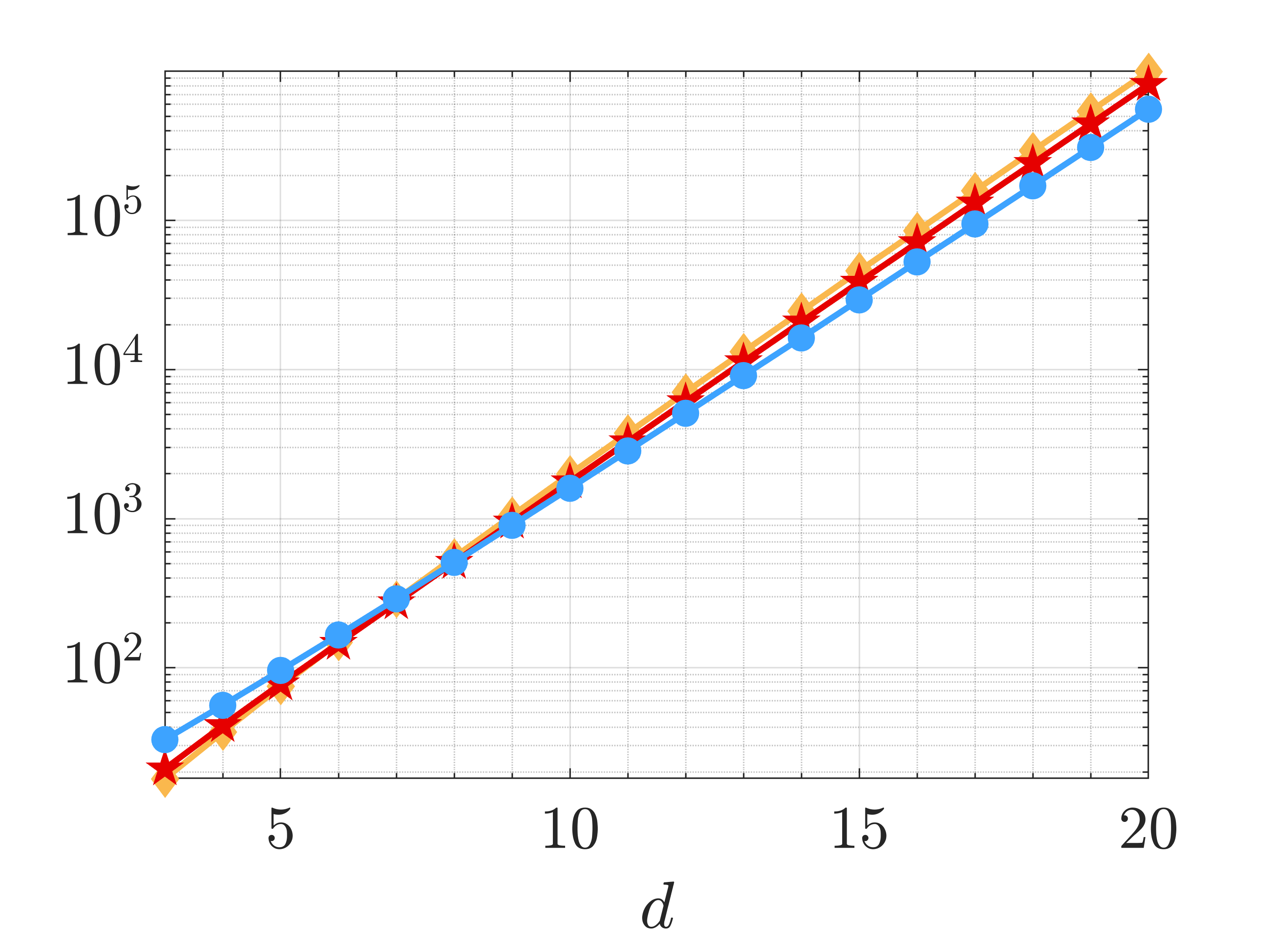}
\end{center}
\caption{\small Left: sensitivity function $S(\rho;R^*,n,s)$ for $d=10$, $s=2$ and $n=1\,000$; the value of $S(R^*;R^*,n,s)$ is indicated by the horizontal dashed line. Right: $n_1(d,s)$ as a function of $d$ for $s=2$ ({\color{Dandelion} $\blacklozenge$}), $s=4$ ({\color{red} $\bigstar$}) and $s=10$ ({\color{Cerulean} $\bullet$}).}
\label{F:sensitivity}
\end{figure}

From Zador's theorem, $\QQ_{R^*}$ ceases to be optimal for $n$ larger than some $n_1(d,s)$. The right panel presents $n_1(d,s)$, obtained by binary search based on Theorem~\ref{Th:CNS}, for $d=3,4,\ldots,20$ and $s=2,4$ and $10$ (on a logarithmic scale). It shows that for $n$ smaller than a value $n_1(d,s)$ growing exponentially fast with $d$, a measure uniform on a single sphere with radius $R^*=R^*(d,n,s)$ minimizes $D_{\QQ_{1,1},s}(\QQ^{[n]})$. This has the consequence that, for $n\leq n_1(d,s)$, the measure $\PP^*=\PP_{0,R^*/\sqrt{d}}$ minimizes $D_{\QQ_{1,1},s}(\overline \PP^{[n]})$ with respect to $\PP$ on $\RR^d$, with $\overline\PP$ the symmetrized version of $\PP$ defined by \eqref{symmetrization}, and the optimal $\PP_{\alpha,\delta}$ minimizing $\widehat D_{\mu,s}(\PP_{\alpha,\delta}^{[n]})$ is $\PP_{0,\delta_0^*}$ with $\delta_0^*=R^*/\sqrt{d}$.

For $n>n_1(d,s)$, $\QQ^*=\sum_{i=1}^q w_i^* \QQ_{R_i^*}$ is a mixture, and the product measure $\PP^*=\sum_{i=1}^q w_i^* \PP_{0,R_i^*/\sqrt{d}}$ minimizes $D_{\QQ_{1,1},s}(\overline \PP^{[n]})$. An algorithm for constructing $\QQ^*$ is provided in \cite{PZ2026b-submitted}.
However, for such large values of $n$ this approach is unnecessarily complex and  Section~\ref{S:n>2^d} proposes a simpler alternative construction.

For $i=0,1$, let $\delta_i^*$ denote the value of $\delta$ minimizing $\widehat D_{\mu,s}(\PP_{i,\delta}^{[n]})$.
The example in the next section shows that, in terms of $\widehat D_{\mu,s}(\cdot)$, $\PP_{1,\delta_1^*}$ performs only slightly worse than $\PP_{0,\delta_0^*}$. The former is uniform on $\delta_1^*\,\SC_d$, whereas the latter is concentrated on the vertices of $\delta_0^*\,\SC_d$. It also shows that, for fixed $d$, the values of $\delta_0^*$ and $\delta_1^*$ vary slowly with $n$ and $s$, with $\delta_1^*$ being significantly larger than $\delta_0^*$.

\subsection{Example: $\PP_{0,\delta_0^*}$ and $\PP_{1,\delta_1^*}$}\label{S:Example-1}

We take $s=2$, $d=10,20,\ldots,50$ and consider designs of sizes $n=\tau\,d$ with $\tau=10,20,\ldots,100$. For all cases considered, we have  $n<n_1(d,s)$ and the optimal $\ma$ equals zero. The optimal $\delta_0^*(d,n,s)$ is obtained by minimizing the double integral
\be\label{D0}
\widehat D_{\mu,s}(\PP_{0,\delta}^{[n]}) = s \int_{t\geq 0} t^{s-1} \int_{r\geq 0} H_0(r,t;\delta,n)\, \phi_{1,1}(r)\,\dd r\,\dd t \,, 
\ee
with respect to $\delta\in(0,1]$, where $\phi_{1,1}$ is given by \eqref{phi11} and
$H_0(r,t;\delta,n) =[1- I_\upsilon(\dt,\dt)]^n$
with $\upsilon=[t^2 -(\delta\sqrt{d}-r)^2]/(4 r\delta\sqrt{d})$, see (\ref{H}, \ref{v}). The values of $\delta_0^*(d,\tau d,2)$ are shown on the left panel of Figure~\ref{F:delta_s2_alpha0_d10-50_n10d-100d}, while the right panel shows $\widehat D_{\mu,2}(\PP_{0,\delta_0^*}^{[n]})$.
Note the slow increase of $\delta_0^*$ with $\tau$ (i.e., with the design size $n$).

\begin{figure}[ht!]
\begin{center}
\includegraphics[width=.49\linewidth]{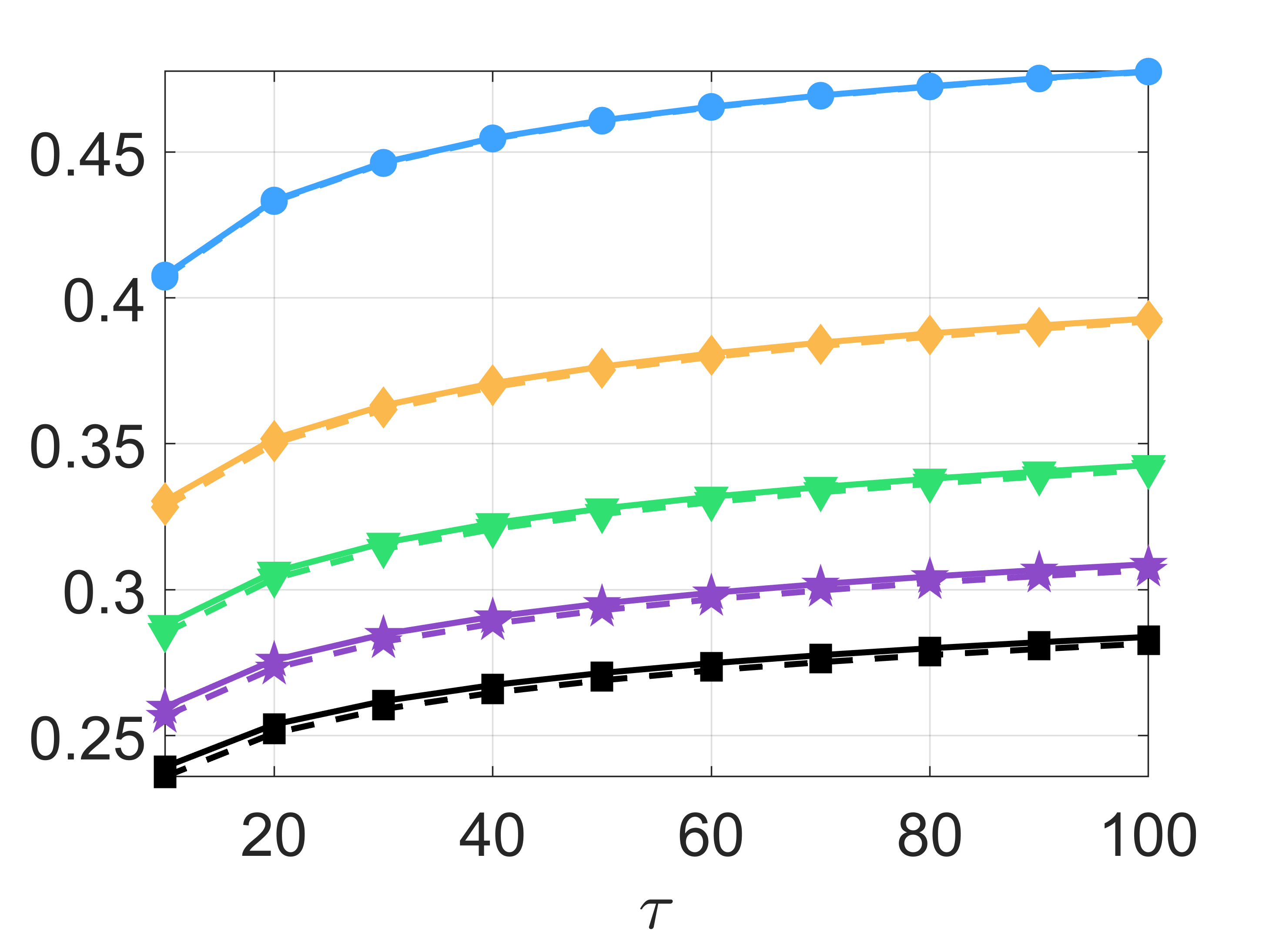}
\includegraphics[width=.49\linewidth]{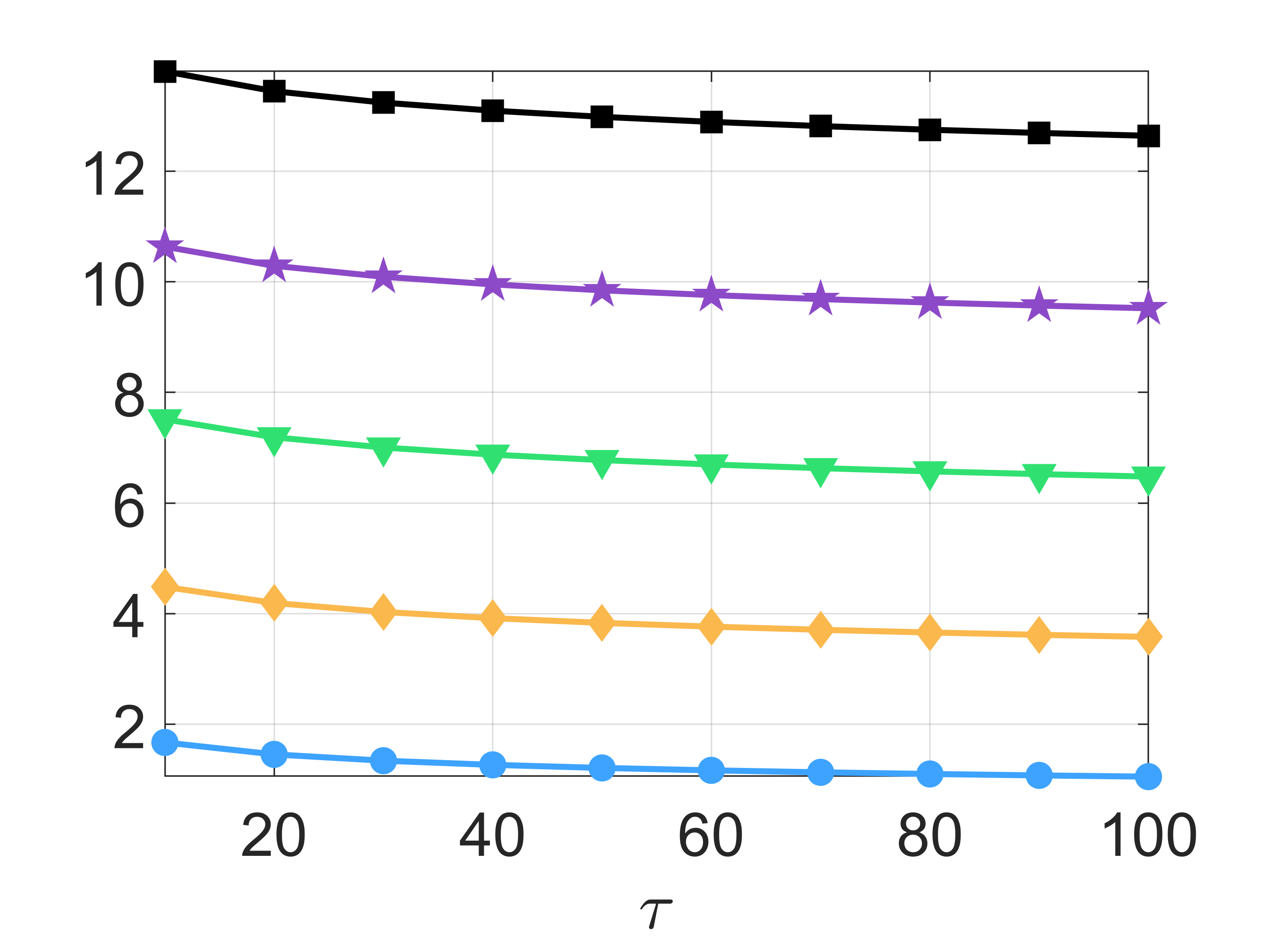}
\end{center}
\caption{\small Left: optimal $\delta_0^*=\delta_0^*(d,\tau d,2)$ minimizing $\widehat D_{\mu,2}(\PP_{0,\delta}^{[\tau d]})$
as a function of $\tau$ for $d=10$ ({\color{Cerulean} $\bullet$}), $d=20$ ({\color{Dandelion} $\blacklozenge$}), $d=30$ ({\color{green} $\blacktriangledown$}), $d=40$ ({\color{Orchid} $\bigstar$}), and $d=50$ ({\tiny $\blacksquare$}); the dashed lines correspond to the extreme-value approximation \eqref{EV-approx-cube}.
Right: $\widehat D_{\mu,2}(\PP_{0,\delta_0^*}^{[\tau d]})$ as a function of $\tau$, with the same color code as on the left panel.}
\label{F:delta_s2_alpha0_d10-50_n10d-100d}
\end{figure}

For $\ma=1$, $\delta_1^*(d,n,s)$ minimizes $\widehat D_{\mu,s}(\PP_{1,\delta}^{[n]})$ given by (\ref{Ls-mean-final}, \ref{H}), where $\Psi(\cdot)$ and $\Phi(\cdot)$ are the c.d.f.\ associated with $\phi_{1,1}(\cdot)$ and $\varphi(\cdot\,;a_d,b_d|\delta^2\,M_d)$, respectively; see \eqref{eq:beta_general} and \eqref{eq: matching three moments}. 
The left panel of Figure~\ref{F:delta_s2_alpha1_d10-50_n10d-100d} shows the optimal $\delta_1^*(d,\tau d,2)$ obtained when we set $\ma=1$: $\delta_1^*$ is much larger than $\delta_0^*$ on Figure~\ref{F:delta_s2_alpha0_d10-50_n10d-100d}-left and also increases with $\tau$, i.e.\ with $n$ (asymptotically, the optimal distribution is uniform, with $\ma=\delta=1$). The plots of 
$\widehat D_{\mu,2}(\PP_{1,\delta_1^*}^{[\tau d]})$
are visually indistinguishable from those of 
$\widehat D_{\mu,2}(\PP_{0,\delta_0^*}^{[\tau d]})$
on Figure~\ref{F:delta_s2_alpha0_d10-50_n10d-100d}-right, and  Figure~\ref{F:delta_s2_alpha1_d10-50_n10d-100d}-right presents the efficiencies  
$\widehat D_{\mu,2}(\PP_{0,\delta_0^*}^{[\tau d]})/\widehat D_{\mu,2}(\PP_{1,\delta_1^*}^{[\tau d]})$
obtained for the different $d$ and $n$ considered: the performance degradation due to the imposition of $\ma=1$ quickly appears negligible as $d$ increases.

\begin{figure}[ht!]
\begin{center}
\includegraphics[width=.49\linewidth]{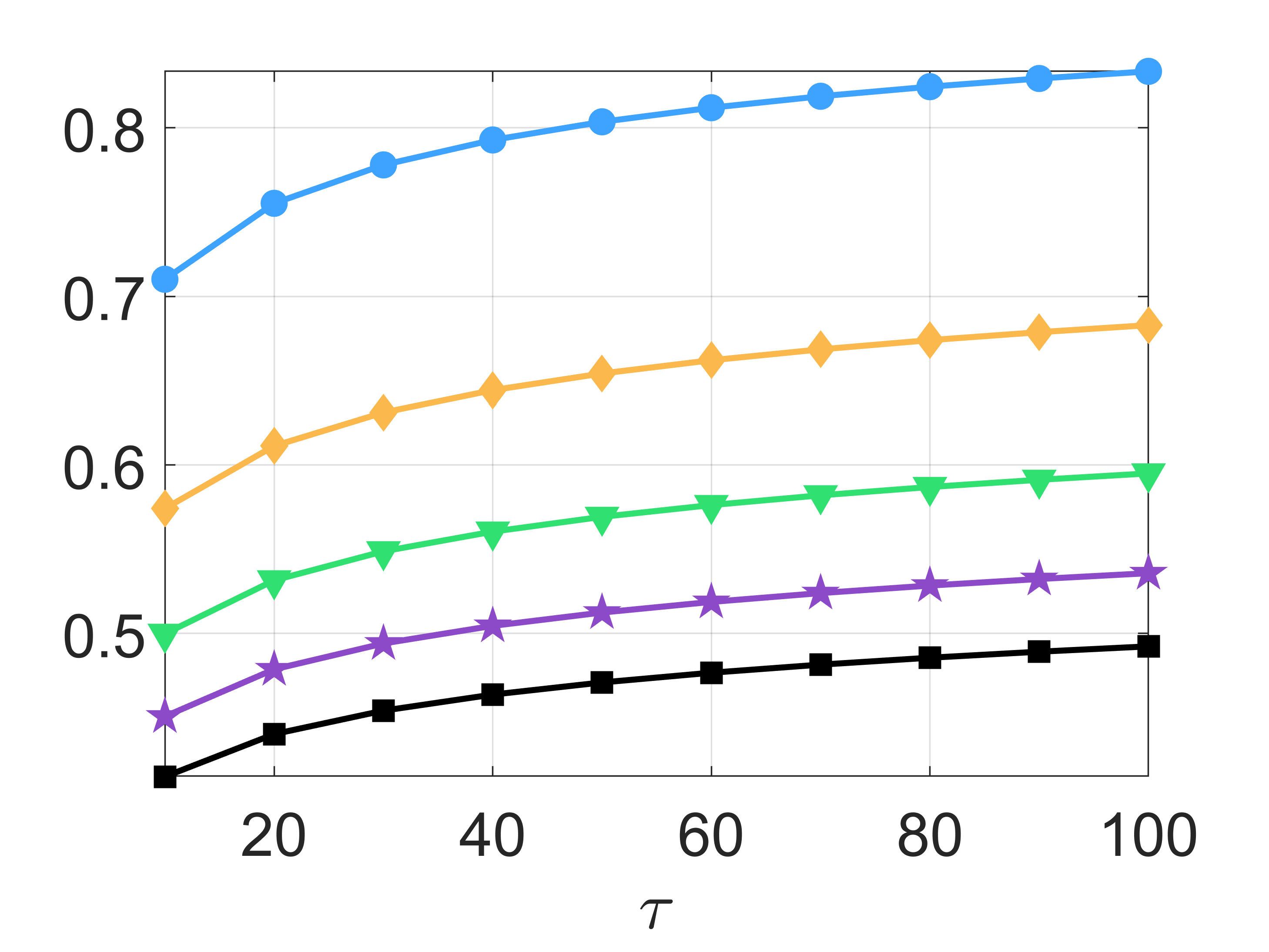}
\includegraphics[width=.49\linewidth]{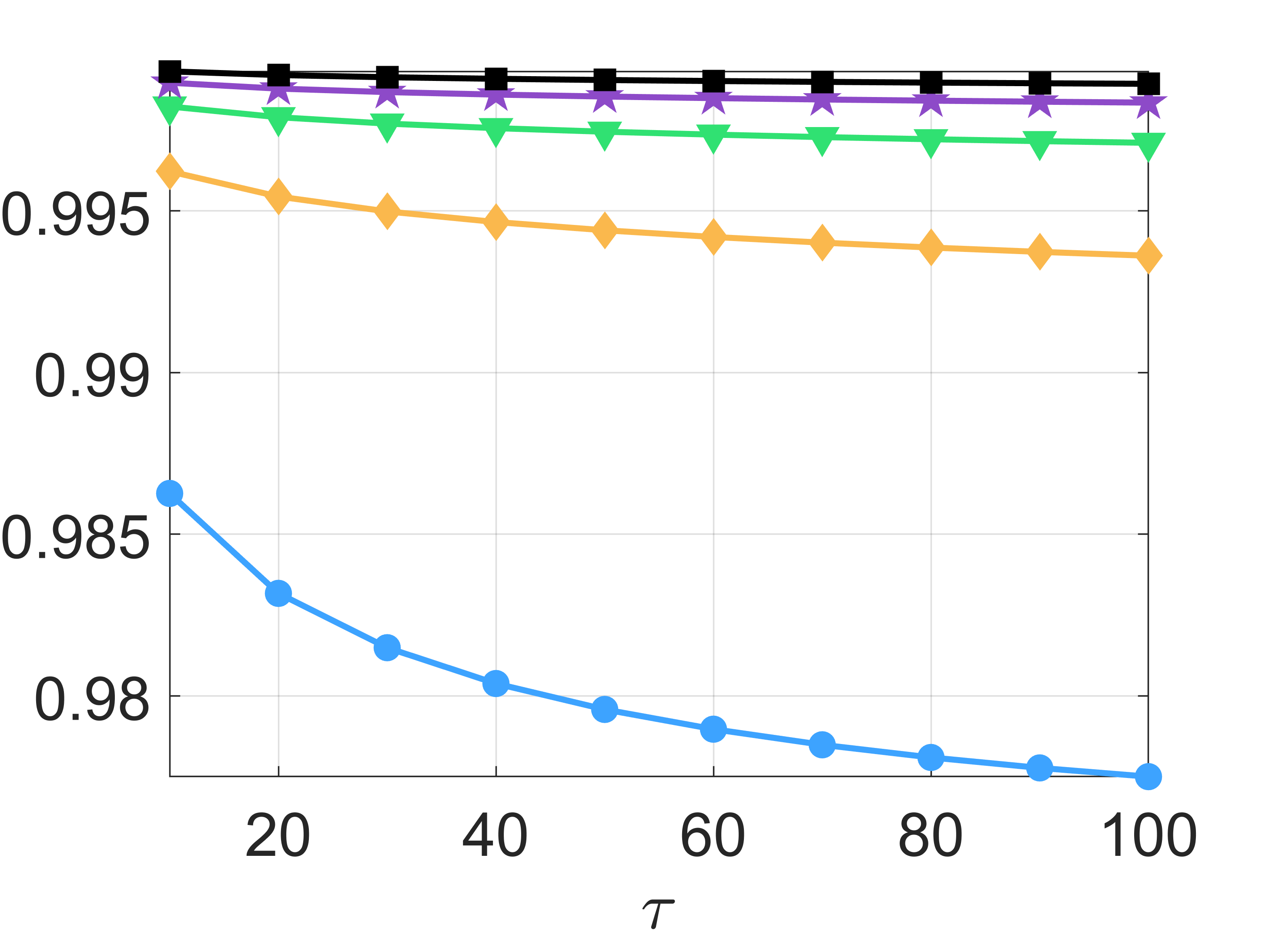}
\end{center}
\caption{\small Left: Optimal $\delta_1^*=\delta_1^*(d,\tau d,2)$ minimizing 
$\widehat D_{\mu,2}(\PP_{1,\delta}^{[\tau d]})$
as a function of $\tau$. Right: efficiency  
$\widehat D_{\mu,2}(\PP_{0,\delta_0^*}^{[\tau d]})/\widehat D_{\mu,2}(\PP_{1,\delta_1^*}^{[\tau d]})$.
The color code is as in Figure~\ref{F:delta_s2_alpha0_d10-50_n10d-100d}.}
\label{F:delta_s2_alpha1_d10-50_n10d-100d}
\end{figure}

We now set $n=50\,d$ and vary $d$ and $s$. The optimal $\ma$ equals zero, and the left panel of Figure~\ref{F:delta_s1-15_alpha0_d10-50_n50d} plots the optimal $\delta_0^*(d,50\,d,s)$ versus $s$ for different $d$. As $s$ increases, the boundary points play a more important role, so that $\delta_0^*$ (slowly) increases with $s$. The right panel shows 
$[\widehat D_{\mu,s}(\PP_{0,\delta_0^*}^{[50\, d]})]^{1/s}$
as a function of $s$. 
When we impose $\ma=1$, the evolution of the optimal values $\delta_1^*(d,50\,d,s)$ follows that shown in the left panel of Figure~\ref{F:delta_s1-15_alpha0_d10-50_n50d}, up to a shift toward higher values, similar to the shift between the left panels of Figures~\ref{F:delta_s2_alpha0_d10-50_n10d-100d} and \ref{F:delta_s2_alpha1_d10-50_n10d-100d}. The efficiency ratios $[\widehat D_{\mu,s}(\PP_{0,\delta_0^*}^{[50\, d]})]^{1/s}/[\widehat D_{\mu,s}(\PP_{1,\delta_1^*}^{[50\, d]})]^{1/s}$ are about 0.99 for $d=10$ and above 0.997 for $d\geq20$; i.e., the increase in quantization error when using $\ma=1$ instead of $\ma=0$ is negligible for large $d$. In contrast, the choice of $\delta$ (adapted to the chosen $\ma$) has a significant effect.

\begin{figure}[ht!]
\begin{center}
\includegraphics[width=.49\linewidth]{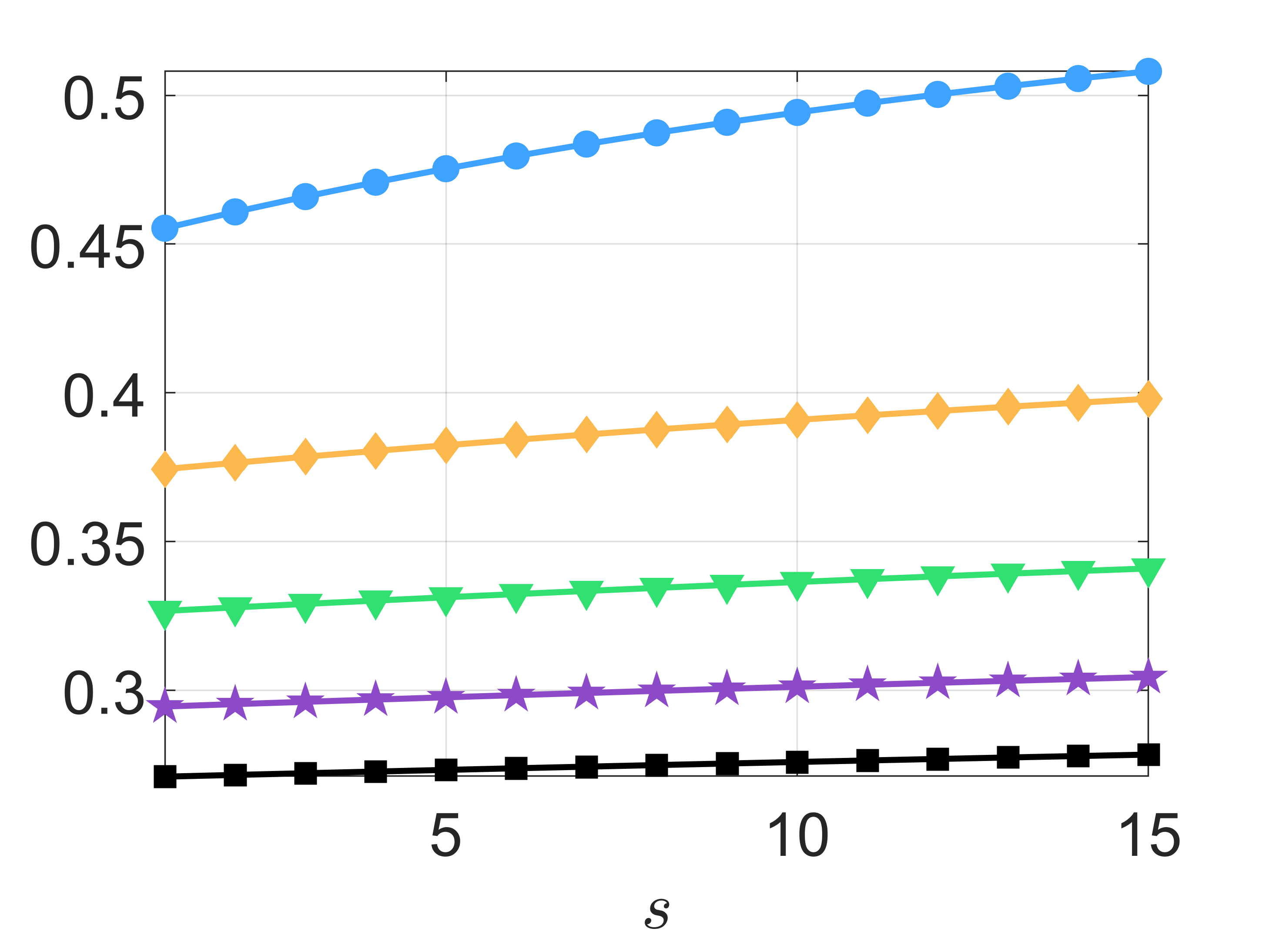}
\includegraphics[width=.49\linewidth]{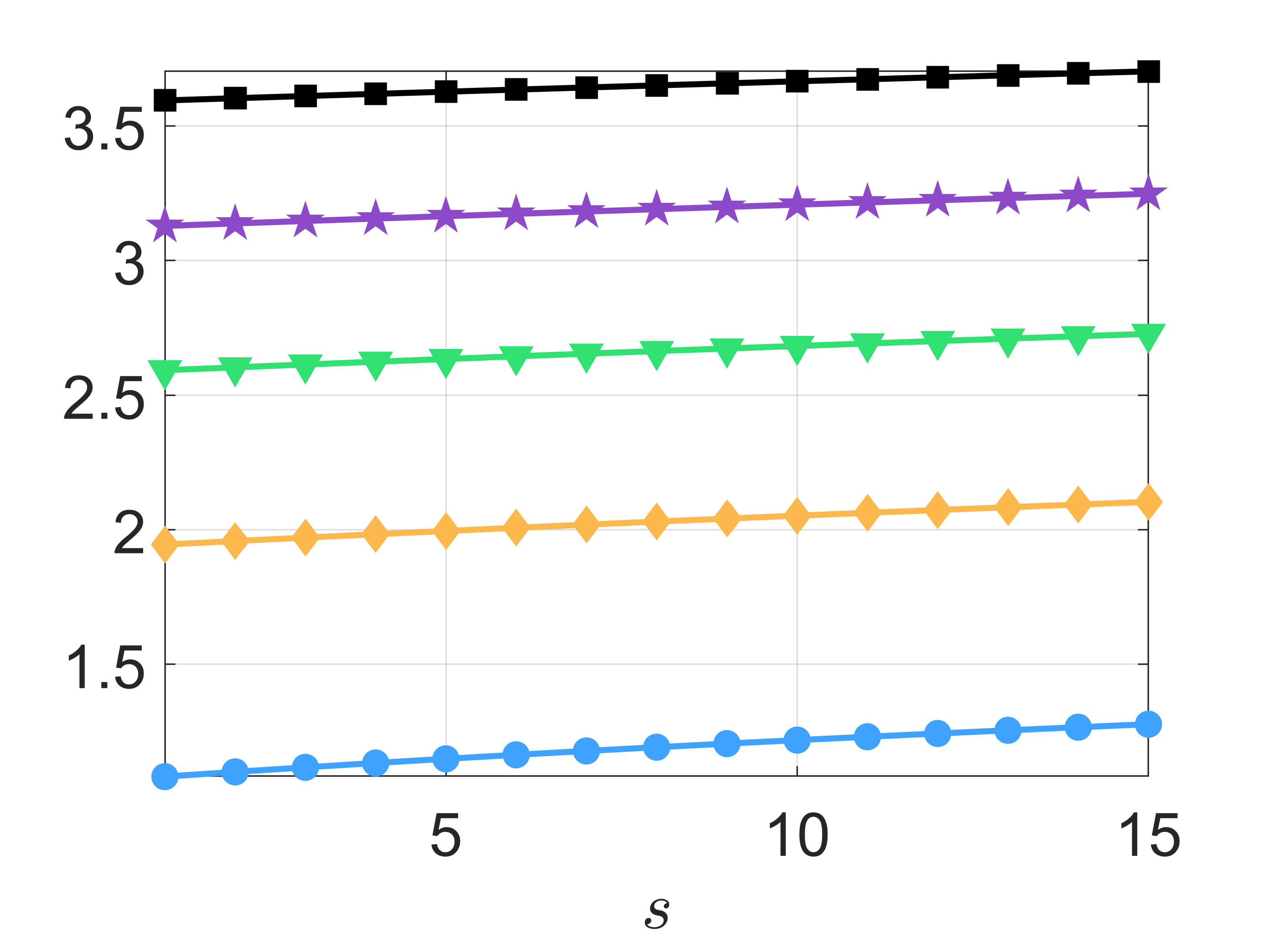}
\end{center}
\caption{\small Left: optimal $\delta_0^*=\delta_0^*(d,50\,d,s)$ minimizing $\widehat D_{\mu,s}(\PP_{0,\delta}^{[50\, d]})$
as a function of $s$ for $d=10$ ({\color{Cerulean} $\bullet$}), $d=20$ ({\color{Dandelion} $\blacklozenge$}), $d=30$ ({\color{green} $\blacktriangledown$}), $d=40$ ({\color{Orchid} $\bigstar$}), and $d=50$ ({\tiny $\blacksquare$}). Right: 
$[\widehat D_{\mu,s}(\PP_{0,\delta_0^*}^{[50\, d]})]^{1/s}$
as a function of $s$, with the same color code as on the left panel.}
\label{F:delta_s1-15_alpha0_d10-50_n50d}
\end{figure}


\subsubsection*{Extreme-value approximation for $\delta_0^*$}
An approximation of $D_{\mu,s}(\QQ_R^{[n]})$ is derived in \cite{PZ2026b-submitted} for $\QQ_R$ uniform on $\SS_{d-1}(R)$, $\mu$ spherically symmetric and $d\geq 3$, based on extreme-value theory:\index{extreme-value theory}
\bea
\widetilde D_{\mu,s}(\QQ_R^{[n]})
= \int_{r\geq 0} \int_{\zeta\geq 0} \left[(r-R)^2 + 4\,r\,R\,\kappa_{n,d}\,\zeta\right]^{s/2}\, \dd F_d(\zeta) \, \dd \Psi(r) \,.
\eea
Here, $\Psi(\cdot)$ is the c.d.f.\ of $\|U\|$ when $U\simd \mu$, $F_d(t)=1-\exp[-t(d-1)/2]$ is a Weibull c.d.f., and $\kappa_{n,d}$ is the $1/n$ quantile of the random variable $\beta_{\dt,\dt}$ in \eqref{eq:quantile2}, defined by $I_{\kappa_{n,d}}(\dt,\dt)=1/n$, with $I_\upsilon(\cdot,\cdot )$ the regularized incomplete beta function and $\dt=(d-1)/2$. 
Explicit expressions are obtained when $s$ is an even integer, yielding approximations $\widehat R^*=\widehat R^*(d,n,s)$ for the value of $R^*$ that minimizes $D_{\mu,s}(\QQ_R^{[n]})$ with respect to $R$. 

This directly provides an extreme-value approximation of $\delta_0^*$ that minimizes $\widehat D_{\mu,s}(\PP_{0,\delta}^{[n]})$ defined by \eqref{Phi_alpha_delta}. Indeed, the spherically symmetric approximation of $\PP_{0,\delta}$ is $\QQ_{0,\delta}=\QQ_{\delta\sqrt{d}}$, uniform on the sphere $\SS_{d-1}(\delta \sqrt{d})$. The approximation of $\delta_0^*$ is thus 
$\widehat \delta_0^*(d,n,s)= \widehat R^*(d,n,s)/\sqrt{d}$.  
In particular, for $s=2$ we get
\be
\widehat \delta_0^*(d,n,2) = \frac{\Ex_{\QQ_{1,1}}\{\|U\|\}}{\sqrt{d}}\, \left[1-2\,\kappa_{n,d}\,\Gamma\left(1+\frac{2}{d-1}\right)\right] \,, \label{EV-approx-cube}
\ee
with $U\simd \QQ_{1,1}$. The beta distribution defined by \eqref{eq:beta_general} yields
\bea
\Ex_{\QQ_{1,1}}\{\|U\|\} = \sqrt{M_d} \, \frac{\Gamma(a_d+1/2)\Gamma(a_d+b_d)}{\Gamma(a_d)\Gamma(a_d+b_d+1/2)}\,,
\eea
where $a_d,b_d$ and $M_d$ are given by \eqref{eq: matching three moments}. The values of $\widehat \delta_0^*(d,n,2)$ are plotted as dashed lines on the left panel of Figure~\ref{F:delta_s2_alpha0_d10-50_n10d-100d}. Notice the good agreement with the values $\delta_0^*$ obtained by numerical minimization of $\widehat D_{\mu,2}(\PP_{0,\delta}^{[n]})$. On the right panel, the values of $\widehat D_{\mu,2}(\PP_{0,\widehat \delta_0^*}^{[\tau d]})$ are indistinguishable from those of $\widehat D_{\mu,2}(\PP_{0,\delta_0^*}^{[\tau d]})$. 

 
\subsubsection*{Asymptotics for $\delta_0^*(d,n(d),s)$ and $\min_{\delta\geq0}\widehat D_{\mu,s}(\PP_{0,\delta}^{[n(d)]})$}
When $n=n(d)$ grows with $d$, the optimal $\delta_0^*(d,n(d),s)$ converges to a limit that depends on the growth regime of $n$ but not on $d$. Indeed, let $\QQ_{1,1;d}$ denote the spherically symmetric approximation of the uniform probability measure on the rescaled hypercube $\sqrt{3/d}\,\SC_d$. When $X\simd \QQ_{1,1;d}$, it follows from \eqref{CLT-normX2-cube} that $\|X\|\to 1$ in probability as $d\to\infty$, and Theorem~4.2 of \cite{PZ2026b-submitted}  implies
\be\label{asymptotics-delta0}
\delta_0^*(d,n(d),s)\to \left\{
\begin{array}{ll}
1/\sqrt{3} & \mbox{if } n^{1/d}\to\infty\,,\\[1mm]
\sqrt{1-1/\ml^2}/\sqrt{3}
& \mbox{if } n^{1/d}\to\ml\in(1,\infty)\,,\\[1mm]
0 & \mbox{if } (\log n)/d\to0\,.
\end{array}
\right.
\ee
Moreover, we also have
\be\label{asymptotics-ED0}
\frac{1}{d^{s/2}}\min_{\delta\geq0}\widehat D_{\mu,s}(\PP_{0,\delta}^{[n(d)]})
\to
\left\{
\begin{array}{ll}
0 & \!\!\! \mbox{if } n^{1/d}\to\infty\,,\\[1mm]
1/(3^{s/2}\, \ml^s)
& \!\!\!\mbox{if } n^{1/d}\to\ml\in(1,\infty)\,,\\[1mm]
1/3^{s/2} & \!\!\! \mbox{if } (\log n)/d\to0\,. 
\end{array}
\right.    
\ee
Figure~\ref{F:asymptotics-delta0} gives an illustration in the case $n(d)=\ml^d$ for $s=2$.

\begin{remark}[The $2^d$ product quantizer]\label{Rk:product quantizer}
Let $\Xb_{2^d}^*$ denote the $2^d$-point design $\{\pm \frac12,\ldots,\pm\frac12\}$. This product quantizer corresponds to the limiting case $\ml=2$; it is stationary in the sense that each point is the centroid of its Voronoi cell (the $2^d$ orthants of the cube), which is necessary (but not sufficient) for global optimality when $s\geq 1$ \cite[Th.~4.1]{GrafL2000}. It is optimal among product $2^d$-point quantizers for any $s\geq 1$, and globally optimal among all $2^d$-point quantizers for $d=s=2$ \cite{Roychowdhury2017} and for $s\to\infty$ when $d\leq 3$. Indeed, for $d\leq 3$, $\CR(\Xb_{2^d}^*)=\sqrt{d}/2<1$. Consider a design $\Xb_{2^d}'$ with minimum covering radius. Since the vertices of $\SC_d$ are pairwise at distance at least $2>2\,\CR(\Xb_{2^d}^*)$, none of the $2^d$ points of $\Xb_{2^d}'$ can cover more than one vertex; hence each covers exactly one. Moreover, at least one point must also cover the origin, which implies $\sqrt{d}\leq 2\,\CR(\Xb_{2^d}')$. Therefore $\CR(\Xb_{2^d}')=\sqrt{d}/2$, and $\Xb_{2^d}^*$ is covering optimal.
\fin    
\end{remark}

\begin{figure}[ht!]
\begin{center}
\includegraphics[width=.49\linewidth]{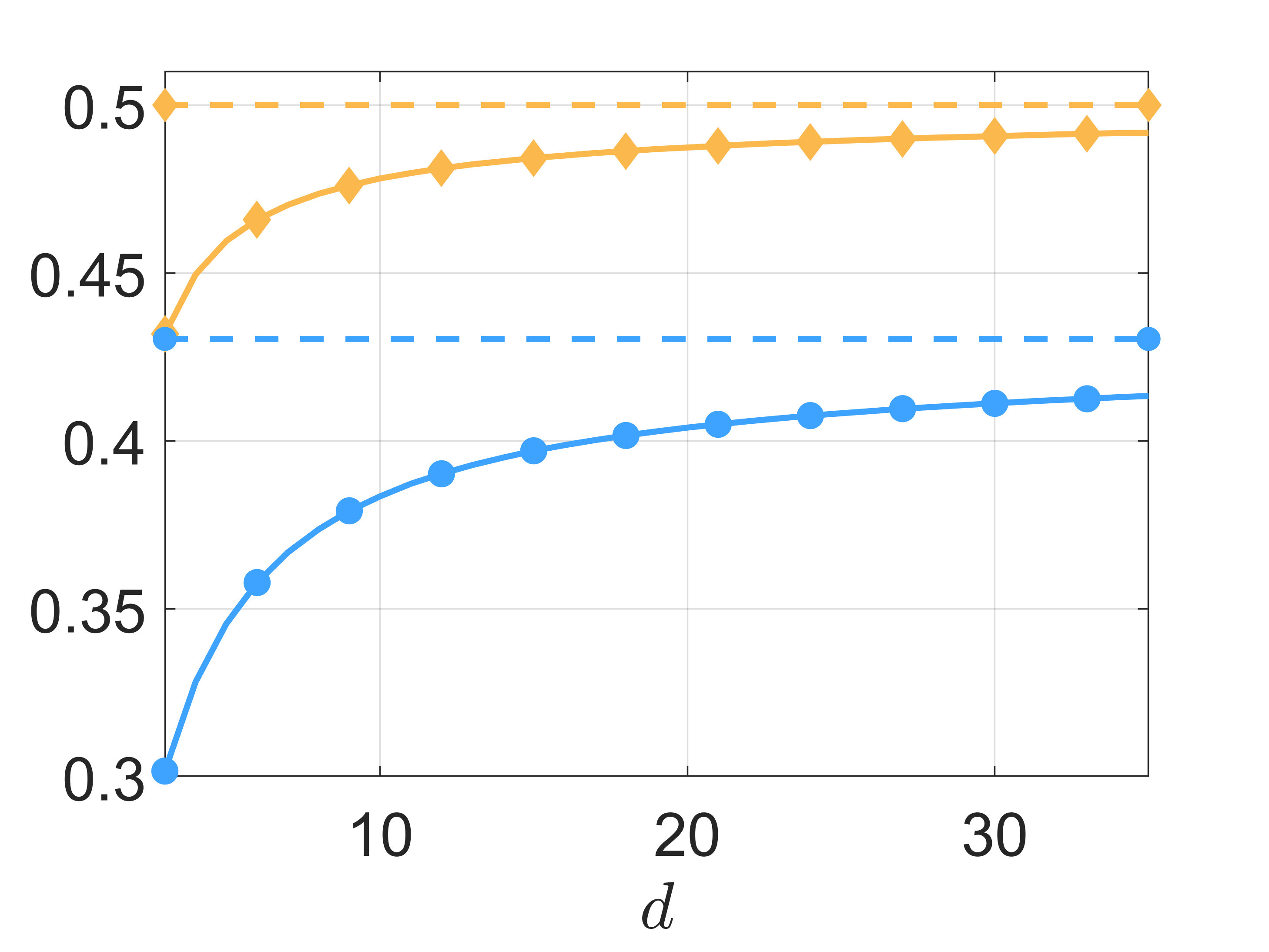}
\includegraphics[width=.49\linewidth]{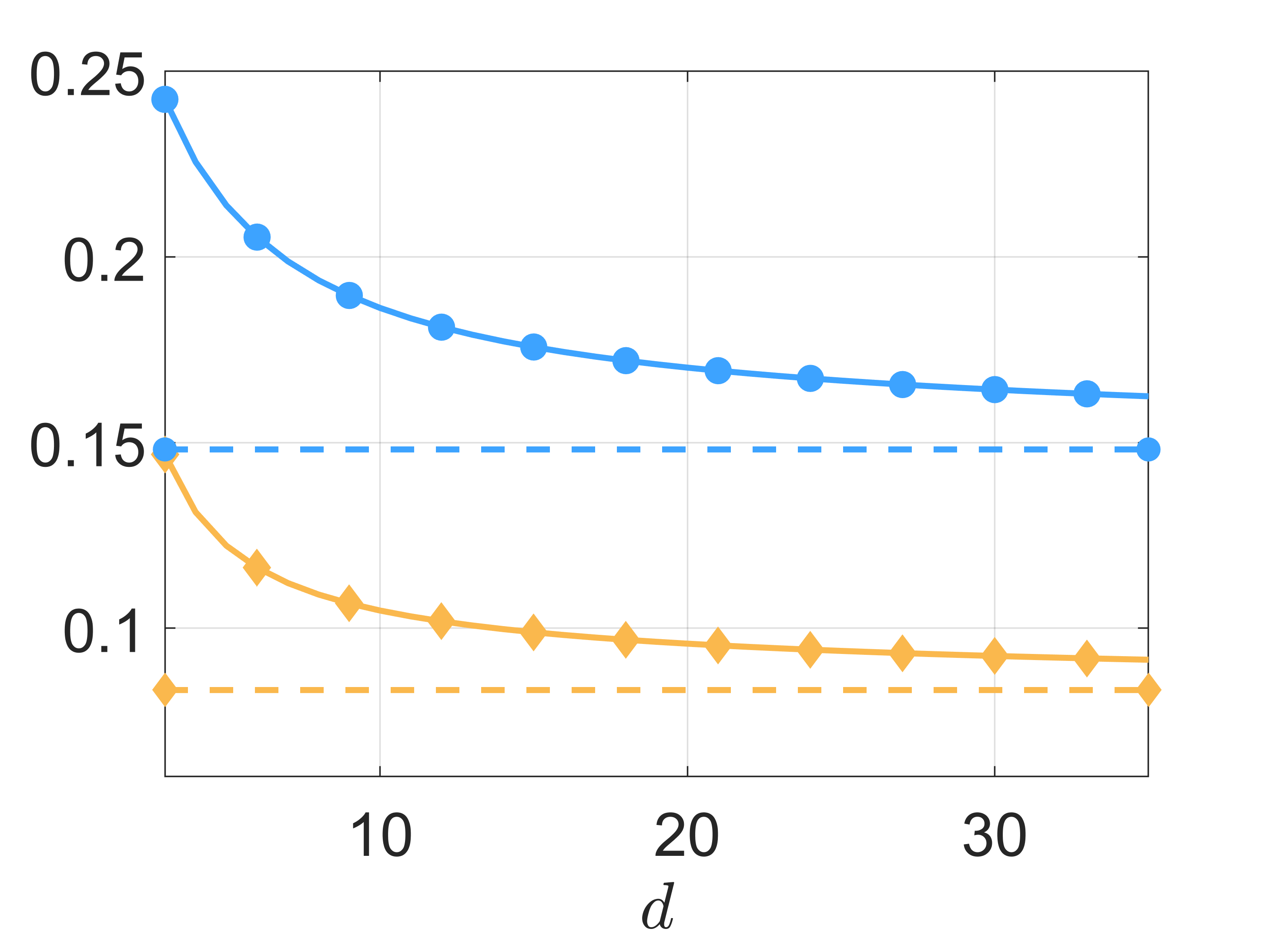}
\end{center}
\caption{\small $\delta_0^*(d,n(d),2)$ (left) and $(1/d)\,\min_{\delta\geq0}\widehat D_{\mu,2}(\PP_{0,\delta}^{[n(d)]})$ (right) as functions of $d$
for $n(d)=\ml^d$ with $\ml=1.5$ ({\color{Cerulean} $\bullet$}) and $\ml=2$ ({\color{Dandelion} $\blacklozenge$}). The limits \eqref{asymptotics-delta0} and \eqref{asymptotics-ED0} are indicated by horizontal dashed lines.}
\label{F:asymptotics-delta0}
\end{figure}

\subsection{Comparison between $D_{\mu,s}(\PP_{\alpha,\delta}^{[n]})$ and $\widehat D_{\mu,s}(\PP_{\alpha,\delta}^{[n]})$}\label{S:quality-approximate-Edistortion}

The calculation of $\widehat D_{\mu,s}(\PP_{\alpha,\delta}^{[n]}) =  D_{\QQ_{1,1},s}(\QQ_{\alpha,\delta}^{[n]})$ is exact, see \eqref{Ls-mean-final}---up to the numerical evaluation of integrals. In contrast with the case of spherically symmetric distributions, we do not have access to the ground truth $D_{\mu,s}(\PP_{\alpha,\delta}^{[n]})$ and need to rely on an estimator.

\subsubsection*{Estimation of $D_{\mu,s}(\PP^{[n]})$}
For any probability measure $\mu$ on a compact set $\SX$, the $(\mu,s)$-distortion of an arbitrary $n$-point design $\Xb_n$ can be approximated by the Monte Carlo estimator $D_{\mu_N,s}(\Xb_n)=(1/N)\sum_{i=1}^N d^s(U_i,\Xb_n)$, where the $U_i$ are i.i.d.\ with $\mu$. This estimator satisfies the classical central limit theorem
\bea
\sqrt{N} \left[D_{\mu_N,s}(\Xb_n)-D_{\mu,s}(\Xb_n)\right] \rad \SN(0,V(\Xb_n))\,, \quad N\to\infty\,,
\eea
where $V(\Xb_n)=D_{\mu,2s}(\Xb_n)-D_{\mu,s}^2(\Xb_n)$. Moreover, a direct application of Hoeffding's inequality gives:
\bea
\mbox{for all } \ma\in(0,1)\,, \ \Pr\left\{|D_{\mu_N,s}(\Xb_n)-D_{\mu,s}(\Xb_n)|>\ma\, \CR^s(\Xb_n) \right\} < 2\,\e1^{-2\,N\ma^2} \,,
\eea
showing the exponentially fast concentration of $D_{\mu_N,s}(\Xb_n)$ to its mean $D_{\mu,s}(\Xb_n)$ as $N$ increases. 
For large $N$, the expected $(\mu,s)$-distortion $D_{\mu,s}(\PP^{[n]})$ can then be estimated by the empirical mean of $D_{\mu_N,s}(\Xb_n)$ over a series of random designs $\Xb_n$ generated with $\PP^{[n]}$.

The left panel of Figure~\ref{F:accuracy} shows that when $N$ is large enough, the variability caused by the estimation of $E_{\mu,s}(\Xb_n)=[D_{\mu,s}(\Xb_n)]^{1/s}$ by $E_{\mu_N,s}(\Xb_n)$ is negligible compared to the variability 
of $E_{\mu,s}(\Xb_n)$ due to the randomness of $\Xb_n$. Here, $d=30$, $n=2^{10}$ and $N=2^{18}$. We generate 100 random designs $\Xb_n^{(i)}$, $i=1,\ldots,100$, each with $n$ points i.i.d.\ uniform in $\SC_d$. The gray envelope is for the values $k^{1/d}\,E_{\mu_N,s}(\Xb_{1:k}^{(i)})$ when $\mu_N$ is the empirical measure on the first $N$ points of a given scrambled Sobol' sequence\footnote{Throughout the paper, scrambled Sobol' points are generated using MATLAB's default scrambling scheme \cite{HongH2003}.} in $\SC_d$ (i.e., the $N$-point prefix of the sequence). The thinner cyan envelope is for $k^{1/d}\,E_{\mu_N^{(i)},s}(\Xb_{1:k}^{(1)})$, where the design $\Xb_n^{(1)}$ is fixed but we consider empirical measures $\mu_N^{(i)}$ supported on the first $N$ points of 100 independent scrambled Sobol' sequences; the magenta envelope is for $k^{1/d}\,E_{\mu_N^{(i)},s}(\Xb_{1:k}^{(2)})$ and empirical measures $\mu_N^{(i)}$ on 100 independent $N$-point designs i.i.d.\ uniform in $\SC_d$. The variability of $k^{1/d}\,E_{\mu_N^{(i)},s}(\Xb_k)$ induced by $\mu_N$ is roughly similar whether $\mu_N$ is supported on Sobol' points or on uniform designs, and is negligible compared to the variability arising from the random choice of $\Xb_n$. 
In the rest of the paper, all numerical evaluations of an $L_s$-mean quantization error in $\SX=\SC_d$ use $\mu_N$ supported on the first $N$ points of a scrambled Sobol' sequence in $\SC_d$. 

\begin{figure}[ht!]
\begin{center}
\includegraphics[width=.49\linewidth]{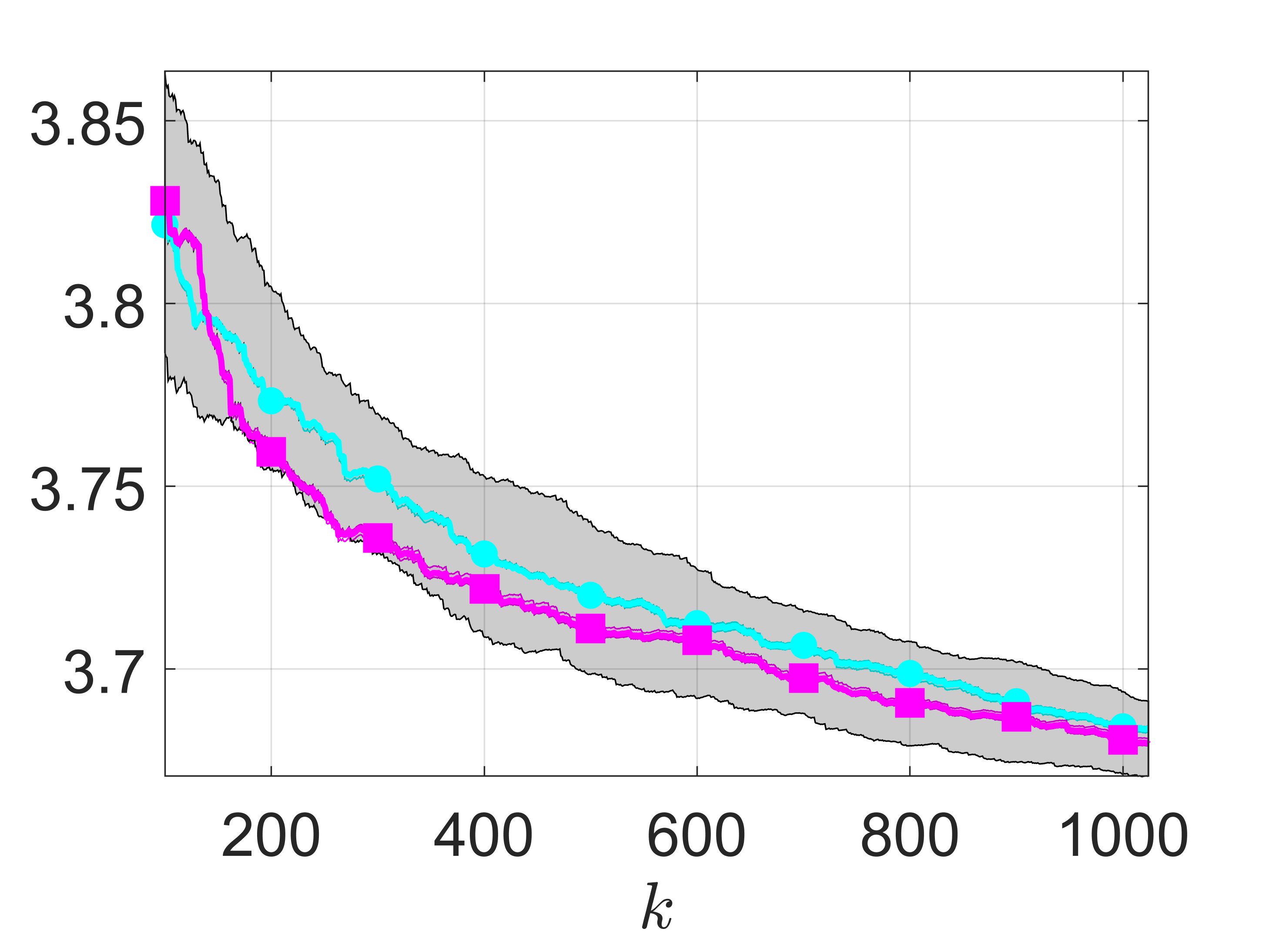}
\includegraphics[width=.49\linewidth]{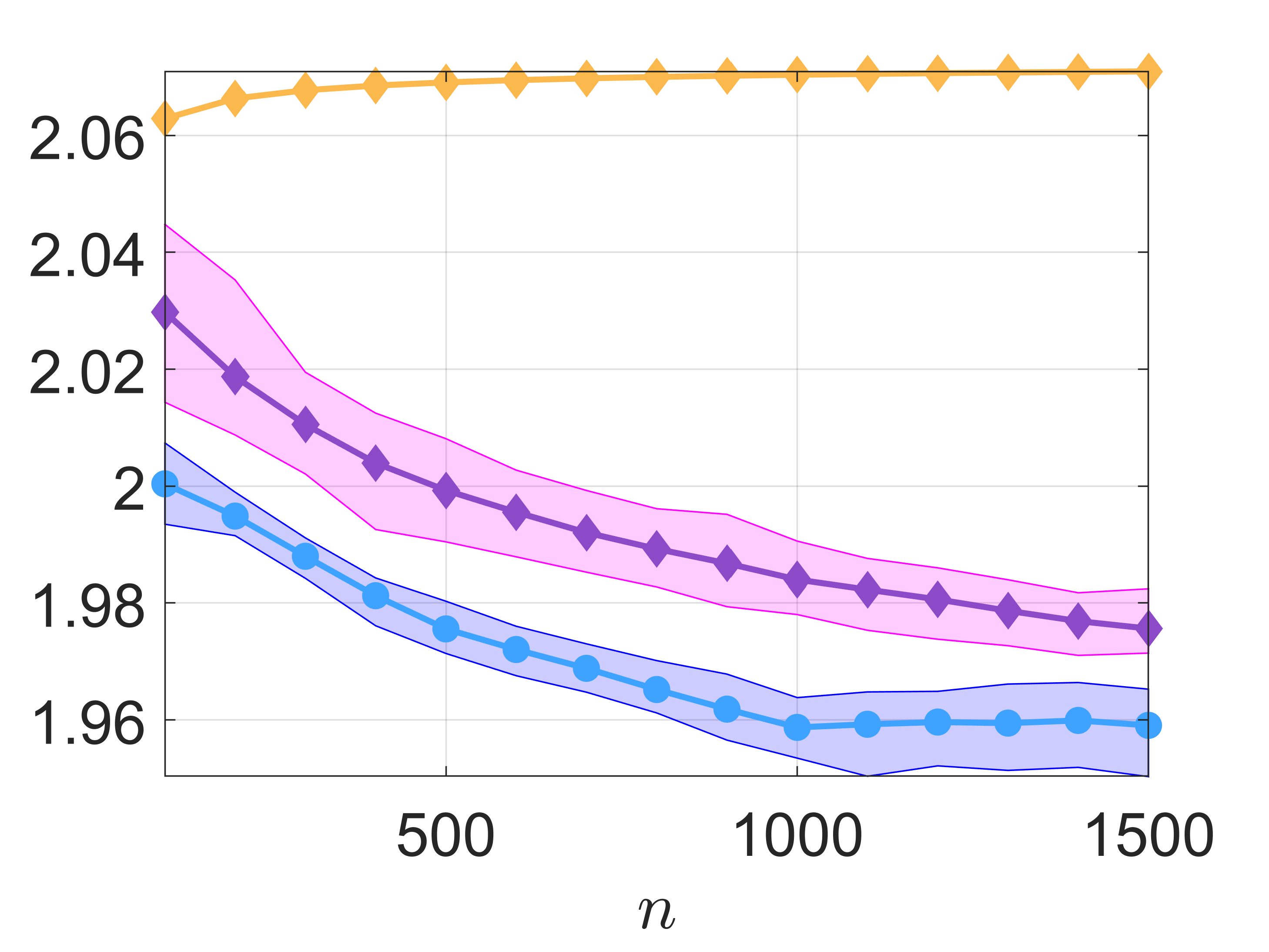}
\end{center}
\caption{\small Left: envelope of $k^{1/d}\,E_{\mu_N,s}(\Xb_{1:k}^{(i)})$ for 100 random designs $\Xb_n^{(i)}$ ($n=2^{10}$, $d=30$), uniform in $\SC_d$ (gray), and envelope of $k^{1/d}\,E_{\mu_N^{(i)},s}(\Xb_{1:k}^{(j)})$ for one random design $\Xb_n^{(j)}$ and 100 different empirical measures $\mu_N^{(i)}$ on $N$-point sets, $N=2^{18}$: $j=1$ and initial points of 100 scrambled Sobol' sequences ({\color{cyan} $\bullet$}); $j=2$ and 100 random $N$-point designs i.i.d.\ uniform in $\SC_d$ ({\color{magenta} {\tiny $\blacksquare$}}). 
Right: $n^{1/d}[\widehat D_{\mu,s}(\PP_{1,\delta_1^*}^{[n]})]^{1/s}$ ({\color{Dandelion} $\blacklozenge$}) and $n^{1/d}[\Ex\{D_{\mu_N,s}(\Xb_n)\}]^{1/s}$ for $\Xb_n\simd \PP_{1,\delta_1^*}^{[n]}$ ({\color{Orchid} $\blacklozenge$}) and $\Xb_n$ given by the $n$-point prefix of Sobol' sequence in $\delta_1\,\SC_d$ ({\color{Cerulean} $\bullet$}), as functions of $n$ (the envelopes for 100 independent Monte Carlo realizations and 100 independent scrambled Sobol' sequences are indicated).}
\label{F:accuracy}
\end{figure}

\subsubsection*{Comparison of $D_{\mu,s}(\PP_{\alpha,\delta}^{[n]})$ with $\widehat D_{\mu,s}(\PP_{\alpha,\delta}^{[n]})$ for $\ma=1$ and $\ma=0$}

We illustrate the approximation error introduced by considering spherically symmetric distributions through an example with $d=10$, $s=2$ and values of $n$ between $100$ and $1\,500$. Our numerical results indicate that $\widehat D_{\mu,s}(\PP_{\alpha,\delta}^{[n]})$ substantially overestimates $D_{\mu,s}(\PP_{\alpha,\delta}^{[n]})$. The behavior is similar for other values of $d$ and $s$.

We first consider the case $\ma=1$, so that $\PP_{1,\delta}$ is uniform on $\delta\,\SC_d$. We compare $D_{\mu_N,s}(\PP_{1,\delta}^{[n]})$, with $\mu_N$ uniform on the first $N=2^{15}$ points of a scrambled Sobol' sequence, with $\widehat D_{\mu,s}(\PP_{1,\delta}^{[n]})=D_{\QQ_{1,1},s}(\QQ_{1,\delta}^{[n]})$ computed with \eqref{Ls-mean-final}, where $\QQ_{1,\delta}$ is such that $\|X\|^2$ has the density $\varphi(t;a_d,b_d|\delta^2\,M_d)$ defined by \eqref{eq:beta_general} and $a_d$, $b_d$ and $M_d$ are given by \eqref{eq: matching three moments}.
For each $n$ considered, $n=100,200,\ldots,1\,500$, we calculate the optimal $\delta_1^*(n)$ minimizing $\widehat D_{\mu,s}(\PP_{1,\delta}^{[n]})$. We then estimate $D_{\mu_N,s}(\PP_{1,\delta_1^*}^{[n]})=\Ex_{\PP_{1,\delta_1^*}^{[n]}}\{D_{\mu_N,s}(\Xb_n)\}$ by Monte Carlo: for each $n$, we generate $100$ independent random designs $\Xb_n\simd \PP_{1,\delta_1^*}^{[n]}$ and compute $D_{\mu_N,s}(\Xb_n)=E_{\mu_N,s}^s(\Xb_n)$ for each of them. We also estimate $D_{\mu_N,s}(\PP_{1,\delta_1^*}^{[n]})$ by quasi-Monte Carlo, using the first $n$ points of 100 independent scrambled Sobol' sequences in $\delta_1^*\,\SC_d$. The right panel of Figure~\ref{F:accuracy} shows the results obtained. One may notice in particular that $\widehat D_{\mu,s}(\PP_{1,\delta_1^*}^{[n]})$ is already in its asymptotic regime for the values of $n$ considered (with a decrease rate of $n^{-s/d}$); that $\widehat D_{\mu,s}(\PP_{1,\delta_1^*}^{[n]})$ overestimates $D_{\mu,s}(\PP_{1,\delta_1^*}^{[n]})$; and that, unsurprisingly, uniform designs in $\delta_1^*\,\SC_d$ have smaller dispersion when using (scrambled) Sobol' points than when using i.i.d.\ random points. It is shown in \cite[Th.~4.1]{PZ2026b-submitted} that the variability of $D_{\mu,s}(\Xb_n)$ tends to zero as $n\to\infty$ when $\mu$ is spherically symmetric and $\Xb_n$ is a random design uniform on a sphere $\SS_{d-1}(R)$ for some $R>0$. This is not the case here, as $\mu$ and $\PP_{1,\delta}$ are uniform on hypercubes. Nevertheless, the variability of $D_{\mu,s}(\Xb_n)$ remains small\myfootnote{The difference between $[\Ex\{D_{\mu_N,s}(\Xb_n)\}]^{1/s}$ and $\Ex\{E_{\mu_N,s}(\Xb_n)\}$ is less than $10^{-5}$ for all values of $n$ considered.}, as illustrated by the envelopes shown in Figure~\ref{F:accuracy}-right.

We now take $\ma=0$, so that $\QQ_{0,\delta}$ is uniform on $\SS_{d-1}(\delta\sqrt{d})$. To investigate the approximation error introduced by using spherically symmetric distributions, we consider the following three distributions $\PP(\delta)$ with support in $\SS_{d-1}(\delta\sqrt{d})\cap\SC_d$: (\textit{i}) $\PP_U(\delta)$, the uniform distribution on the intersection $\SS_{d-1}(\delta\sqrt{d})\cap \SC_d$; (\textit{ii}) $\PP_R(\delta)=\PP_{0,\delta}$, independent sampling with replacement from the vertices of the hypercube $\delta\,\SC_d$; (\textit{iii}) $\PP_V(\delta)$, sampling without replacement from the vertices (as long as $n\leq 2^d$; for $n\geq 2^d$, $\Xb_n$ corresponds to the entire set of vertices). 

For each $n$ selected, we calculate the optimal $\delta_0^*=\delta_0^*(n)$. The left panel of Figure~\ref{F:EDistortion-d13-s2} presents $n^{1/d}[\widehat D_{\mu,s}(\PP_{0,\delta_0^*}^{[n]})]^{1/s}$ and $n^{1/d}[\Ex\{D_{\mu_N,s}(\Xb_n)\}]^{1/s}$  as functions of $n$, for $\Xb_n\simd \PP_U(\delta_0^*)^{[n]}$. As above, $\mu_N$ is uniform on the first $N=2^{15}$ points of a scrambled Sobol' sequence and the expectation is estimated through $m=100$ Monte Carlo evaluations of $D_{\mu_N,s}(\Xb_n)$---the envelope of the 100 realizations of $n^{1/d}E_{\mu_N,s}(\Xb_n)$ is also shown; again, the variability of $D_{\mu,s}(\Xb_n)$ is small\myref{},
although $\mu$ is not spherically symmetric and $\PP(\delta)$ is not uniform on a sphere (even if its support is contained in $\SS_{d-1}(\delta\sqrt{d})$). 
For $n=100$, $\delta_0^*\sqrt{d} \simeq 1.289$ is sufficiently small for $\PP_U(\delta_0^*)$ to resemble $\QQ_{0,\delta_0^*}$. Furthermore, as the distributions of $\|X\|^2$ are similar for $X\simd\mu$ and $X\simd\QQ_{1,1}$ (see Figure~\ref{F:densities-normX2}), we have $[\widehat D_{\mu,s}(\PP_{0,\delta_0^*}^{[n]})]^{1/s} \approx \Ex\{E_{\mu,s}(\Xb_n)\}$. As $n$ increases, $\delta_0^*(n)$ increases (see Figure~\ref{F:delta_s2_alpha0_d10-50_n10d-100d}-left) and the (normalized) spherical measure $V(n)$ of the intersection $\SS_{d-1}(\delta_0^*\sqrt{d})\cap \SC_d$ decreases (see the curve with red stars on Figure~\ref{F:EDistortion-d13-s2}-left), so that $\PP_U(\delta_0^*)$ resembles less and less $\QQ_{0,\delta_0^*}$: $E_{\mu,s}(\Xb_n)$ quickly differs significantly from $[\widehat D_{\mu,s}(\PP_{0,\delta_0^*}^{[n]})]^{1/s}$ as $n$ grows. The 
approximation $\widehat D_{\mu,s}(\PP_{0,\delta_0^*}^{[n]})$ is actually pessimistic: $\PP_U(\delta_0^*)$ accounts for the geometry of the cube, resulting in a smaller expected $(\mu,s)$-distortion than that predicted by spherically symmetric approximations.

The pessimism of $\widehat D_{\mu,s}(\PP_{0,\delta_0^*}^{[n]})$ is even stronger on the right panel of Figure~\ref{F:EDistortion-d13-s2} where we consider random designs distributed with $\PP_R(\delta_0^*)$ and $\PP_V(\delta_0^*)$. When $\Xb_n\simd\PP_V(\delta_0^*)^{[n]}$ (sampling without replacement), the design is fixed for $n\geq 2^d$ and $n^{1/d} E_{\mu_N,s}(\Xb_n)$ then grows like $n^{1/d}$. When $\Xb_n\simd\PP_R(\delta_0^*)^{[n]}$, the proportion of repeated vertices increases with $n$, an effect similar to slowing the growth of the design size. Ultimately, the behavior is similar to $\PP_V(\delta_0^*)^{[n]}$, as all vertices have been included in $\Xb_n$. Notice the marked superiority of $\PP_R(\delta_0^*)$, and especially $\PP_V(\delta_0^*)$, for moderate $n$ compared with $\PP_U(\delta_0^*)$ (and with $\PP_{1,\delta_1^*}$ on the right panel of Figure~\ref{F:accuracy}). This shows that, despite its strong pessimism, the approximation $\widehat D_{\mu,s}(\PP_{0,\delta}^{[n]})$ can be used to determine a distribution $\PP_{0,\delta_0^*}$ whose support provides a basis for designs with small quantization error, an idea further exploited in the next section. 

\begin{figure}[ht!]
\begin{center}
\includegraphics[width=.49\linewidth]{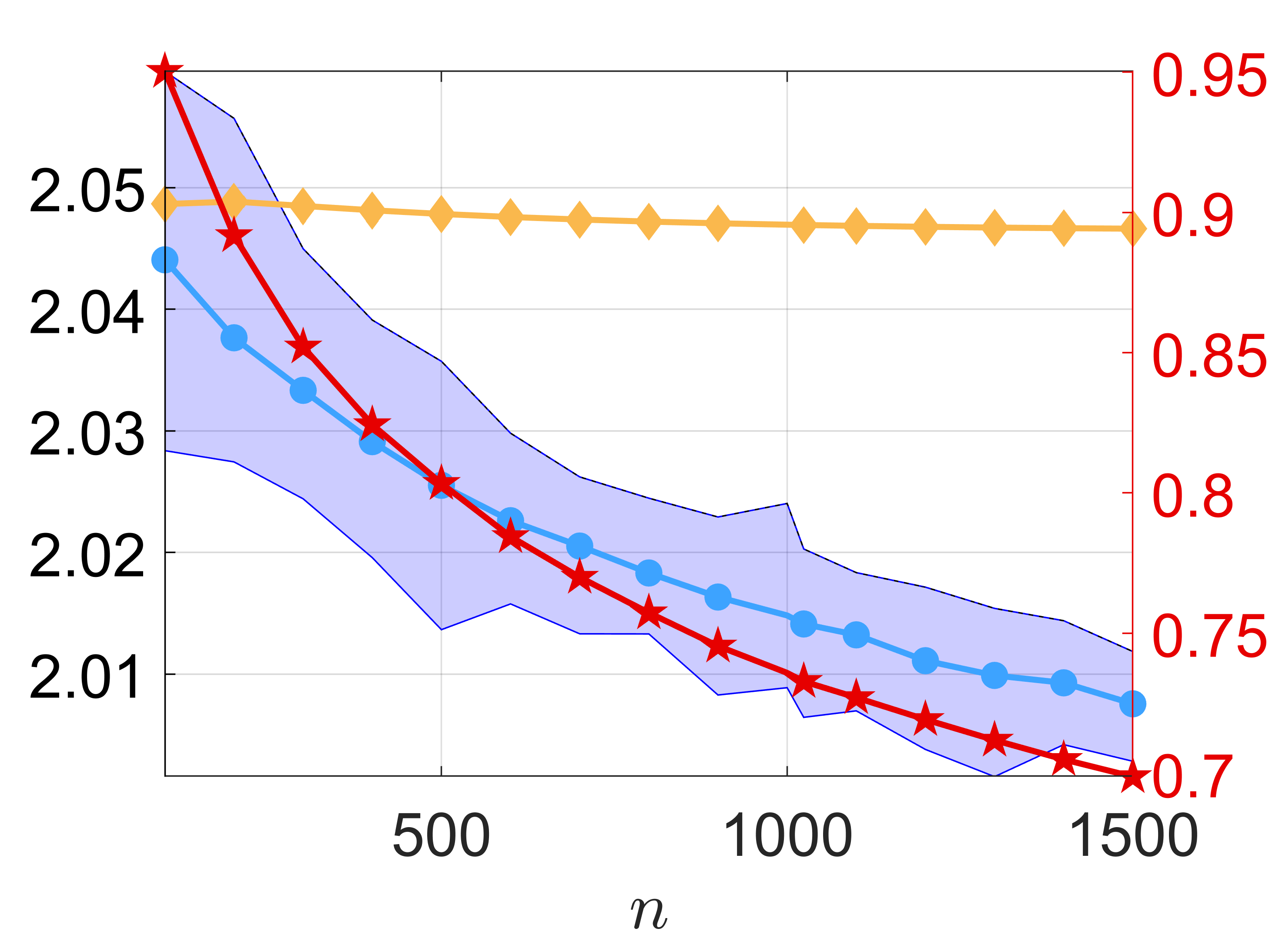}
\includegraphics[width=.49\linewidth]{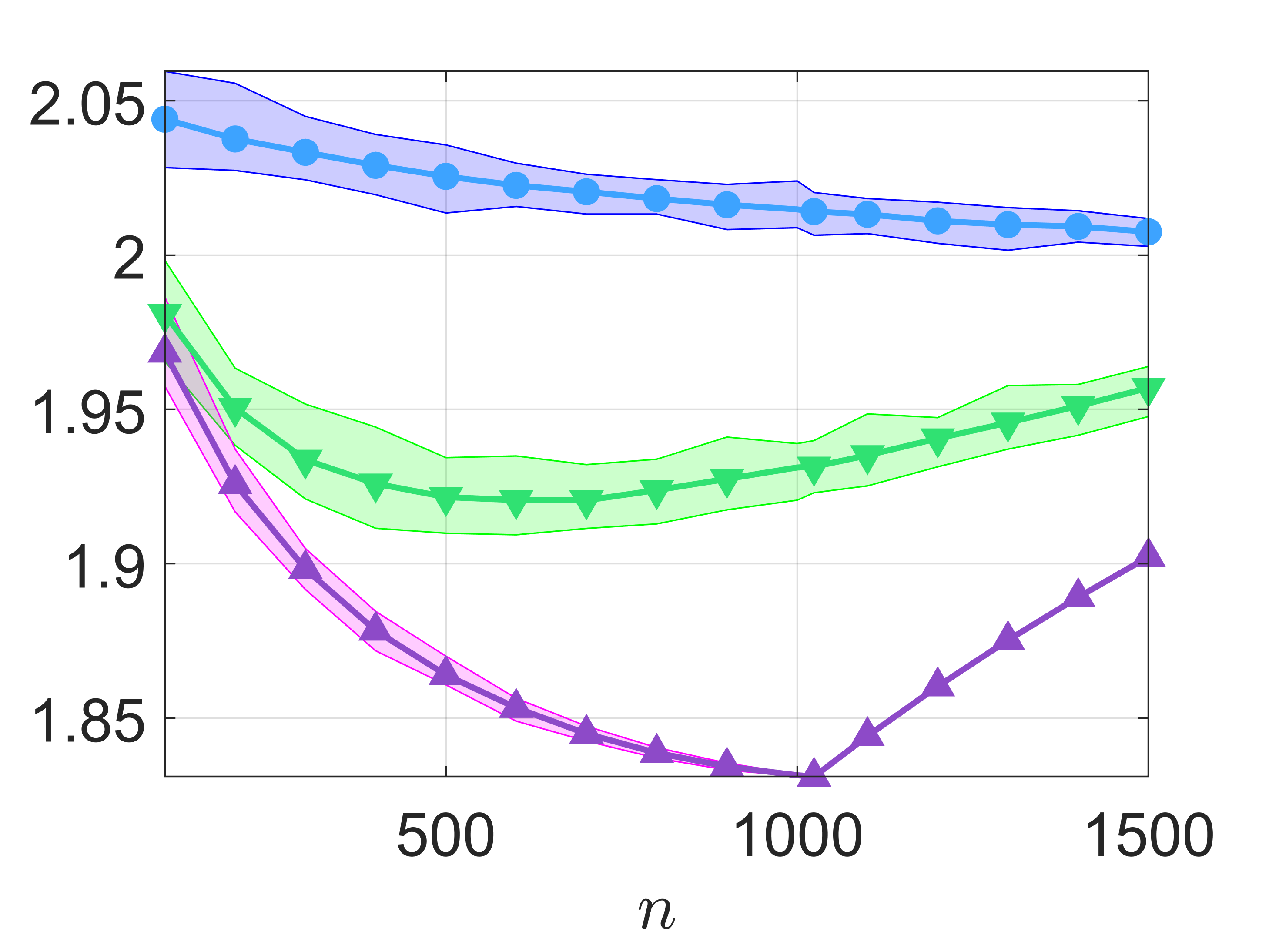}
\end{center}
\caption{\small Left: $n^{1/d}[\widehat D_{\mu,s}(\PP_{0,\delta_0^*}^{[n]})]^{1/s}$ ({\color{Dandelion} $\blacklozenge$}) and $n^{1/d}[\Ex\{D_{\mu_N,s}(\Xb_n)\}]^{1/s}$ for $\Xb_n\simd \PP_U(\delta_0^*)^{[n]}$ ({\color{Cerulean} $\bullet$}) as functions of $n$; Monte  Carlo estimate of the (normalized) spherical measure $V(n)$ of the intersection $\SS_{d-1}(\delta\sqrt{d})\cap \SC_d$ ({\color{red} $\bigstar$}, the scale is indicated by the red labels on the right vertical axis). Right: $n^{1/d}\Ex\{E_{\mu_N,s}(\Xb_n)\}$ for $\Xb_n\simd \PP_U(\delta_0^*)^{[n]}$ ({\color{Cerulean} $\bullet$}), $\PP_R(\delta_0^*)^{[n]}$ ({\color{green} $\blacktriangledown$}) and $\PP_V(\delta_0^*)^{[n]}$ ({\color{Orchid} $\blacktriangle$}) as functions of $n$. The envelopes for 100 independent realizations of $\Xb_n$ are indicated.}
\label{F:EDistortion-d13-s2}
\end{figure}

\section{Incremental designs in $d$-dimensional hypercubes}\label{S:designs-in-cube}

In this section, we compare the quantization-error performance of incremental designs based on $\PP_{0,\delta_0^*}$ and $\PP_{1,\delta_1^*}$ with that obtained with other incremental constructions. The comparison is through numerical examples, with $d=10$ and $d=30$. Determining $\PP_{0,\delta_0^*}$ and $\PP_{1,\delta_1^*}$ requires specifying $n$, which will denote here the maximum targeted design size. In many computer experiment frameworks, it is desirable to be able to interrupt data acquisition at $\Xb_{1:k}$ for any size $k\leq n$ once sufficient information has been gathered. Situations where the order of magnitude of $k$ is about $10\,d$ are of special interest (see \cite{LoeppkySW2009}), but we also consider designs of much larger size.
When $n\leq 2^d$ (see Sections~\ref{S:sampling-without-R}, \ref{S:sampling-fract-fact} and \ref{S:greedy-packing-fract-fact}), designs based on $\PP_{0,\delta_0^*}$ can be constructed by sampling, randomly or deterministically, from the vertices of $\delta_0^*\,\SC_d$. When $d$ is small enough that designs of size $n>2^d$ can be considered (Section~\ref{S:n>2^d}), we suggest constructions based on $\PP_{1,\delta_1^*}$. Since designs contained in the support of $\PP_{0,\delta_0^*}$ suffer from poor dispersion when projected onto lower-dimensional canonical subspaces, perturbing this support is necessary whenever such projections are of interest. Section~\ref{S:projections} explores an approach based on independent perturbations of the vertices of $\delta_0^*\,\SC_d$, trading off quantization in the full hypercube against that in lower-dimensional projections. A numerical comparison is also presented with design points independently distributed according to $\PP_U(\delta_0^*)$, the uniform distribution on $\SS_{d-1}(\delta_0^*)\cap\SC_d$.

For $d=10$, besides random and Sobol' points in $\SC_d$, we also consider two alternative constructions for performance comparison: greedy packing with boundary avoidance \cite{ShangA2020, PZ2022}, and greedy quantization \cite{NogalesPR2021}. However, both methods are too computationally expensive for $d=30$. They are briefly described below.

Let $\Yb_M$ denote a finite set of $M$ candidate points. Greedy packing with boundary avoidance selects the first design point $\xb_1$ at the center of $\SC_d$. Then, at each iteration $k$, the algorithm chooses $\xb_{k+1}$ in $\Arg\max_{\xb\in\Yb_M} \Delta_\beta(\xb;\Xb_{1:k},\SC_d)$, where $\Delta_\beta(\xb;\Xb_{1:k},\SC_d)=\min\{d(\xb,\Xb_{1:k}),\partial \SC_d\}$ with $\partial\SC_d$ the boundary of $\SC_d$; i.e., $\partial\SC_d=\{\xb\in\SC_d:\, |x_i|=1 \mbox{ for some } i\}$. 

The greedy minimization algorithm for $E_{\mu,s}(\Xb_n)$ also starts with $\xb_1$ at the center of $\SC_d$. Then, at each iteration, it selects $\xb_{k+1}$  in $\Arg\min_{\xb\in\SC_d} E_{\mu,s}(\Xb_{1:k}\cup\{\xb\})$; i.e., 
\be\label{greedy-Lsmean}
\xb_{k+1}\in \Arg\min_{\xb\in\SC_d} \int_{\SC_d} \min\left\{ d^s(\yb,\Xb_{1:k}),\|\yb-\xb\|^s\right\}\,\mu(\dd\yb) \,.
\ee
By \cite[Prop.~4.1]{LuschgyP2015}, this greedy algorithm is rate-optimal for any $s>0$ (i.e., it ensures $\limsup_{k\to\infty} k^{1/d} E_{\mu,s}(\Xb_{1:k})< \infty$). In practice, $\xb_{k+1}$ is selected among a finite set $\Yb_{M'}$ of candidates and the integral over $\yb\sim\mu$ in \eqref{greedy-Lsmean} is approximated by a finite sum over a set $\Yb_{M''}$ of $M''$ points uniformly distributed in $\SC_d$. The implementation of the algorithm requires computing all pairwise distances between points in $\Yb_{M'}$ and $\Yb_{M''}$, which strongly limits their feasible sizes. Since the set function $\Xb_n \mapsto E_{\mu,s}^s(\Xb_n)$ is supermodular, an accelerated lazy-greedy version of the algorithm can be implemented \cite{Minoux1977}, yielding significant computational speedup; see~\cite{NogalesPR2021}.

\subsection{Sampling vertices without replacement}\label{S:sampling-without-R}


We set $d=10$ and $n=2^8=256$. For greedy packing with boundary avoidance (GP), the candidate set $\Xb_M$ consists of the first $2^{19}$ points of Sobol' sequence in $\SC_d$. We set $\beta= 2\sqrt{2\,d}$, the value recommended in \cite{ShangA2020} and confirmed experimentally. For the greedy minimization of the quantization error (GQ) $E_{\mu,s}(\Xb_n)$, we take $\Xb_{M'}=\Xb_{M''}$ to be the set of the first $2^{13}$ points of Sobol' sequence in $\SC_d$, augmented with the $2^d$ vertices. Figure~\ref{F:permut-s2} shows $k^{1/d}\,E_{\mu_N,s}(\Xb_{1:k})$ for $k=50,\ldots,n$. Here and throughout Section~\ref{S:designs-in-cube}, $\mu_N$  denotes the uniform measure on the $2^{18}$-point prefix of a scrambled Sobol' sequence in $\SC_d$.

\begin{table}[t!]
\bgroup
\def\arraystretch{1} 
\setlength\tabcolsep{0.12cm} 
\caption{\small Color code for plots in Figure~\ref{F:permut-s2}: random points are i.i.d.\ uniform in $\SC_d$, GP corresponds to $\Xb_n$ generated by greedy packing with boundary avoidance, GQ to $\Xb_n$ obtained by greedy minimization of the $L_s$-mean quantization error for $s=2$ and $s=10$.
}
\label{Tb:color-code-10d_highd}
\begin{center}
\begin{tabular}{l|ll|llll} 
\hline
Design & random & Sobol' (scrambled) &
GP &
GQ &
GQ &
$\PP_{0,\delta_0^*}$ 
\\
& & & & $s=2$ & $s=10$ &  \\
\hline

Color & magenta env. & blue $\bullet$ (blue env.) &  green $\blacktriangle$ & red $\blacklozenge$ & red $\blacktriangledown$ & gray \\
\arrayrulecolor{black!100}
\end{tabular}
\end{center}
\egroup
\end{table}

\begin{figure}[ht!]
\begin{center}
\includegraphics[width=.49\linewidth]{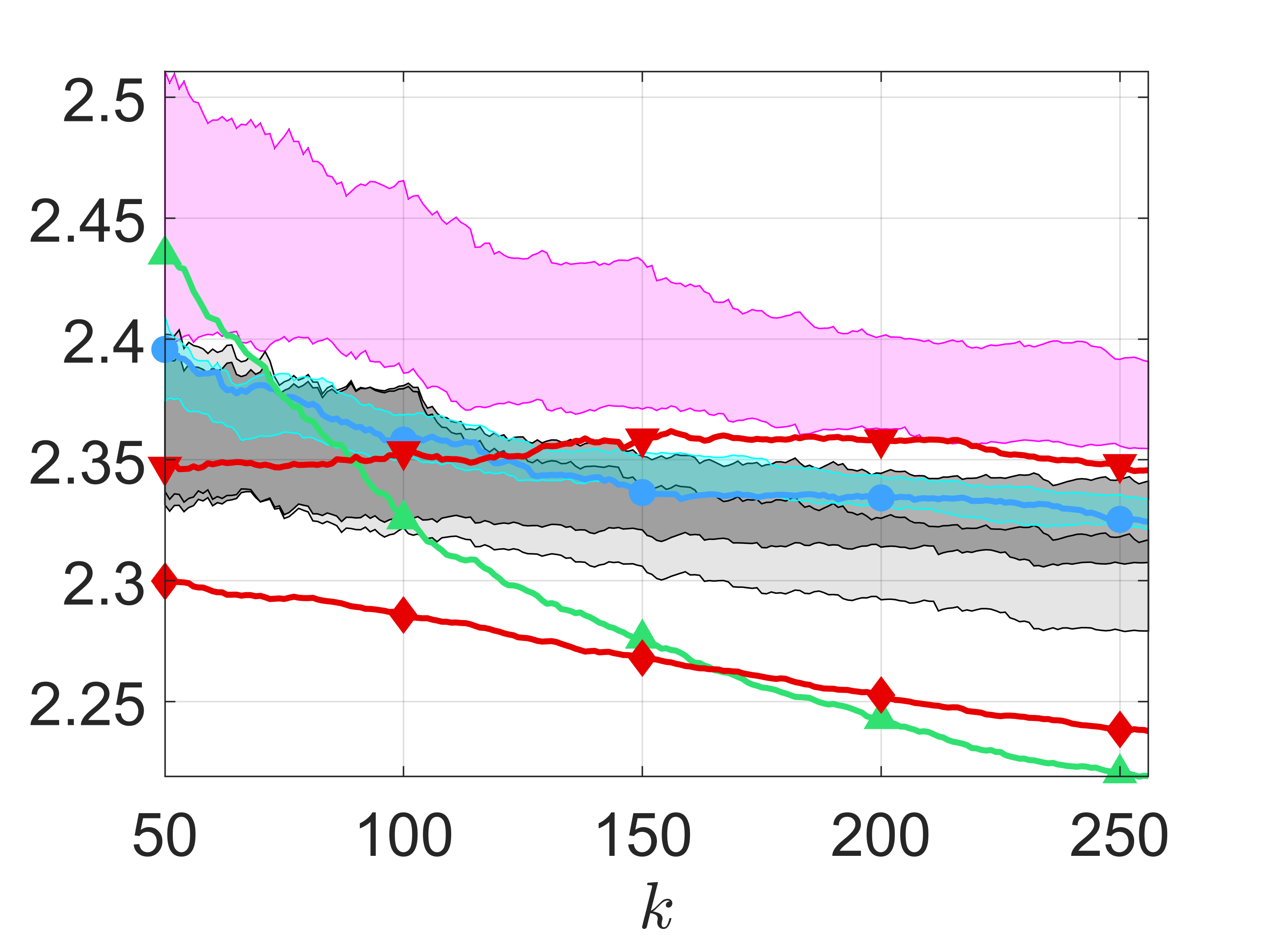}
\includegraphics[width=.49\linewidth]{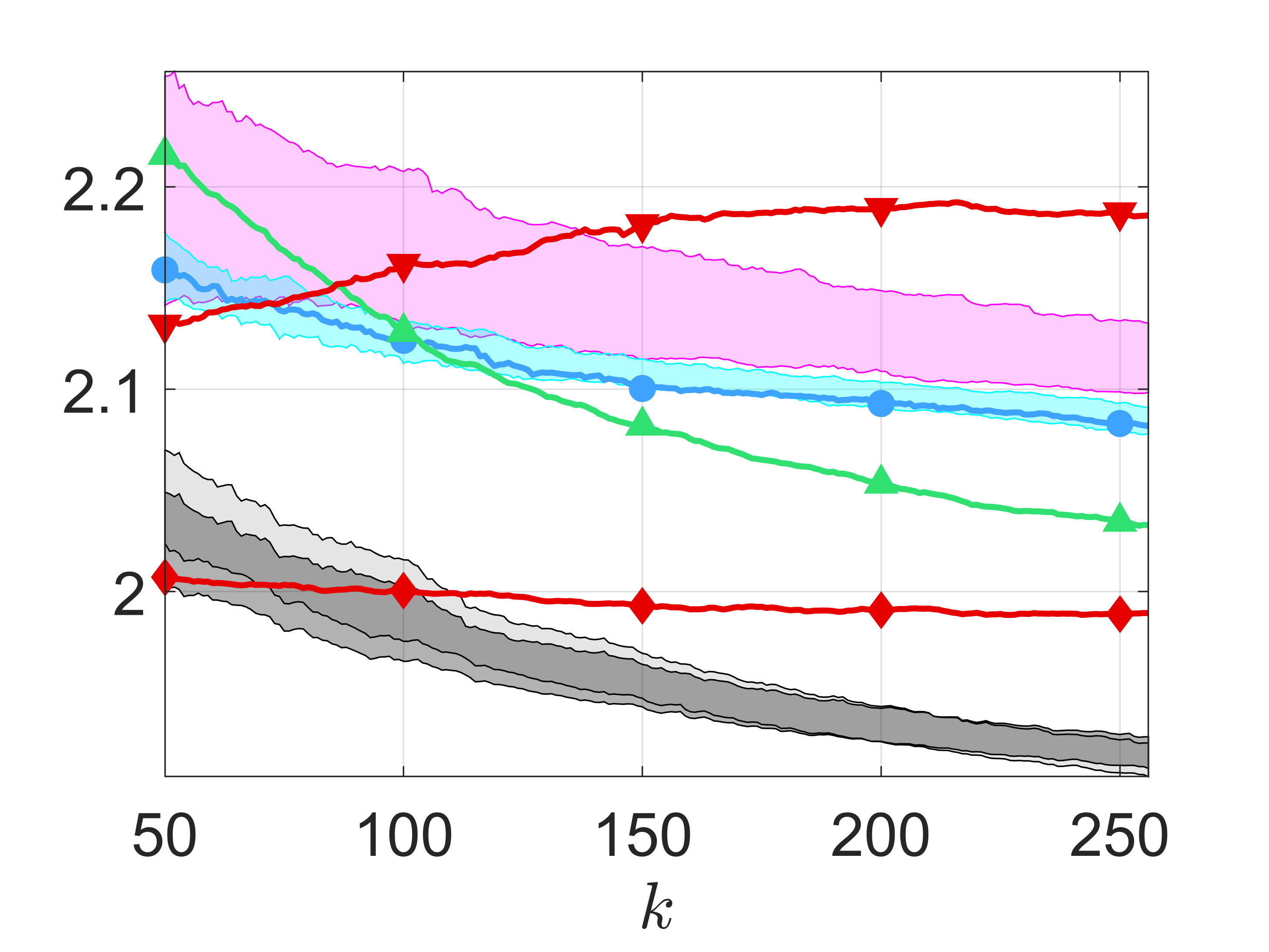}
\end{center}
\caption{\small 
$k^{1/d}\,E_{\mu_N,s}(\Xb_{1:k})$ versus $k$ for $s=10$ (left) and $s=2$ (right), for the designs in Table~\ref{Tb:color-code-10d_highd}. Here $\Xb_n$ based on $\PP_{0,\delta_0^*}$ is a random permutation of the vertices of $\delta_0^*\,\SC_d$: light gray for $\delta_0^*(d,n,10)$, dark gray for $\delta_0^*(d,n,2)$, with the darkest area showing the intersection of the envelopes (100 independent repetitions).}
\label{F:permut-s2}
\end{figure}

Unsurprisingly, scrambled Sobol' points outperform pure random designs, though both are outperformed by other constructions, in particular, for sufficiently large $k$, by greedy packing with boundary avoidance. This approach is grounded in geometrical considerations that make it well-suited for minimizing the covering radius, hence its strong performance in terms of $E_{\mu_N,s}(\Xb_{1:k})$ for large $s$  (left panel, $s=10$).
While it is rather common that greedy methods do not yield the overall best results, more surprising is the fact that greedy minimization of $E_{\mu,2}(\Xb_n)$ produces smaller $E_{\mu,10}(\Xb_{1:k})$ values (left panel) than greedy minimization of $E_{\mu,10}(\Xb_n)$. This may be explained by several factors: the candidate set $\Xb_{M'}$ is much smaller than for greedy packing ($M'=2^{13}$ versus $M=2^{19}$), and the algorithm is quite sensitive to the choice of $\Yb_{M''}$ (here equal to $\Yb_{M'}$) for large $s$ \cite{NogalesPR2021}. In contrast, designs given by permuting the vertices of $\delta_0^*\,\SC_d$ respect the expected ordering: those using $\delta_0^*(d,n,10)$ tend to outperform those using $\delta_0^*(d,n,2)$ in terms of $E_{\mu,10}(\Xb_{1:k})$ (left panel) and vice-versa (right panel). Their performance is, however, nearly identical, which is consistent with $\delta_0^*(d,n,s)$ varying little with $s$ (left panel of Figure~\ref{F:delta_s1-15_alpha0_d10-50_n50d}): $\delta_0^*(d,n,2)\simeq 0.4415$ and $\delta_0^*(d,n,10)\simeq 0.4690$. For $s=2$ (right panel), these random designs outperform the other methods considered.

\subsection{Sampling from a fractional-factorial subset of vertices}\label{S:sampling-fract-fact}

We continue the example from the previous section. As we consider designs of size $k\leq 2^8$, there is no need to consider permutations of all vertices of $\delta_0^*\,\SC_d$. For $\Zb_n=\{\zb_1,\ldots,\zb_n\}$ an $n$-point design with $\zb_i\in\{-1,1\}^d$ for all $i$, the Hamming distance $d_H(\zb_i,\zb_j)$ counts the number of components that differ between $\zb_i$ and $\zb_j$, and thus $d_H(\zb_i,\zb_j) = \frac14\, \|\zb_i-\zb_j\|^2$; the minimum Hamming distance $\rho_H(\Zb_n)$ of $\Zb_n$ is $\rho_H(\Zb_n) =\PR^2(\Zb_n)$ with $\PR(\Zb_n)=\frac12 \min_{i\neq j} \|\zb_i-\zb_j\|$ the packing radius of $\Zb_n$.

For a given $n\leq 2^d$, let $m=\lfloor d-\log_2 n \rfloor$, and let $\Zb_{2^{d-m}}$ be a $2^{d-m}$ fractional factorial design in $\{-1,1\}^d$ such that its minimum Hamming distance $\rho_H(\Zb_{2^{d-m}})$ is as large as possible. For instance, two-level fractional factorial designs $\Zb_{2^{d-m}}$ with large minimum Hamming distance can be obtained using the simulated-annealing algorithm of \cite{CabralFariasPR2023b}\footnote{An implementation and an example of use are provided by the MATLAB scripts {\tt SA-exchange-generators.m} and {\tt Example-fractional-factorial.m}, available at \url{https://github.com/lpronzato-prog/random-quantizers-high-d-cube}.}.
We can then generate random designs by randomly sampling without replacement from $\delta_0^*(d,n,s)\,\Zb_{2^{d-m}}$, which forms a fractional factorial subset of the vertices of $\delta_0^*(d,n,s)\,\SC_d$. In the example considered, $m=2$ and the $2^{d-2}$ fractional factorial design that we use has minimum Hamming distance 2 and resolution VI (with the notation of the pioneering paper \cite{BoxH61}, its generators  are 
$\Gb_1=\mathbf{1     4     5     6     7     8}$ and $\Gb_2=\mathbf{2     3     5     7     8}$; see also \cite[Chap.~5]{WuH2009}).

Figure~\ref{F:permut-frac-fact-s2} is analogous to Figure~\ref{F:permut-s2}, but for designs $\Xb_n$ based on $\PP_{0,\delta_0^*}$ obtained by  random permutations of $\delta_0^*(d,n,s)\,\Zb_{2^{d-m}}$. These random designs clearly outperform those in Figure~\ref{F:permut-s2}. Note that at $k=2^8$ there is no variability, since all points of $\delta_0^*(d,n,s)\,\Zb_{2^{d-m}}$ have already been selected.

\begin{figure}[ht!]
\begin{center}
\includegraphics[width=.49\linewidth]{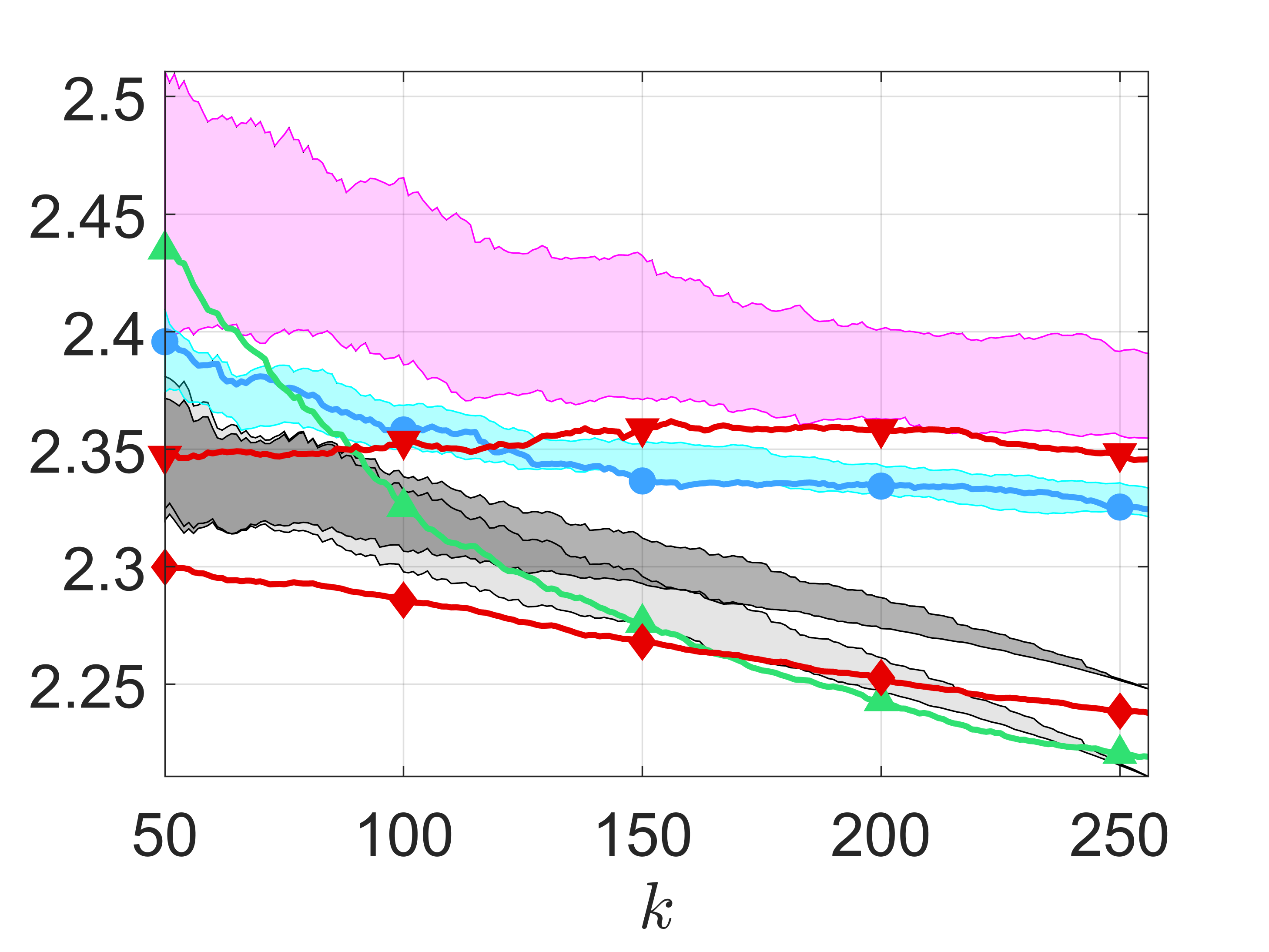}
\includegraphics[width=.49\linewidth]{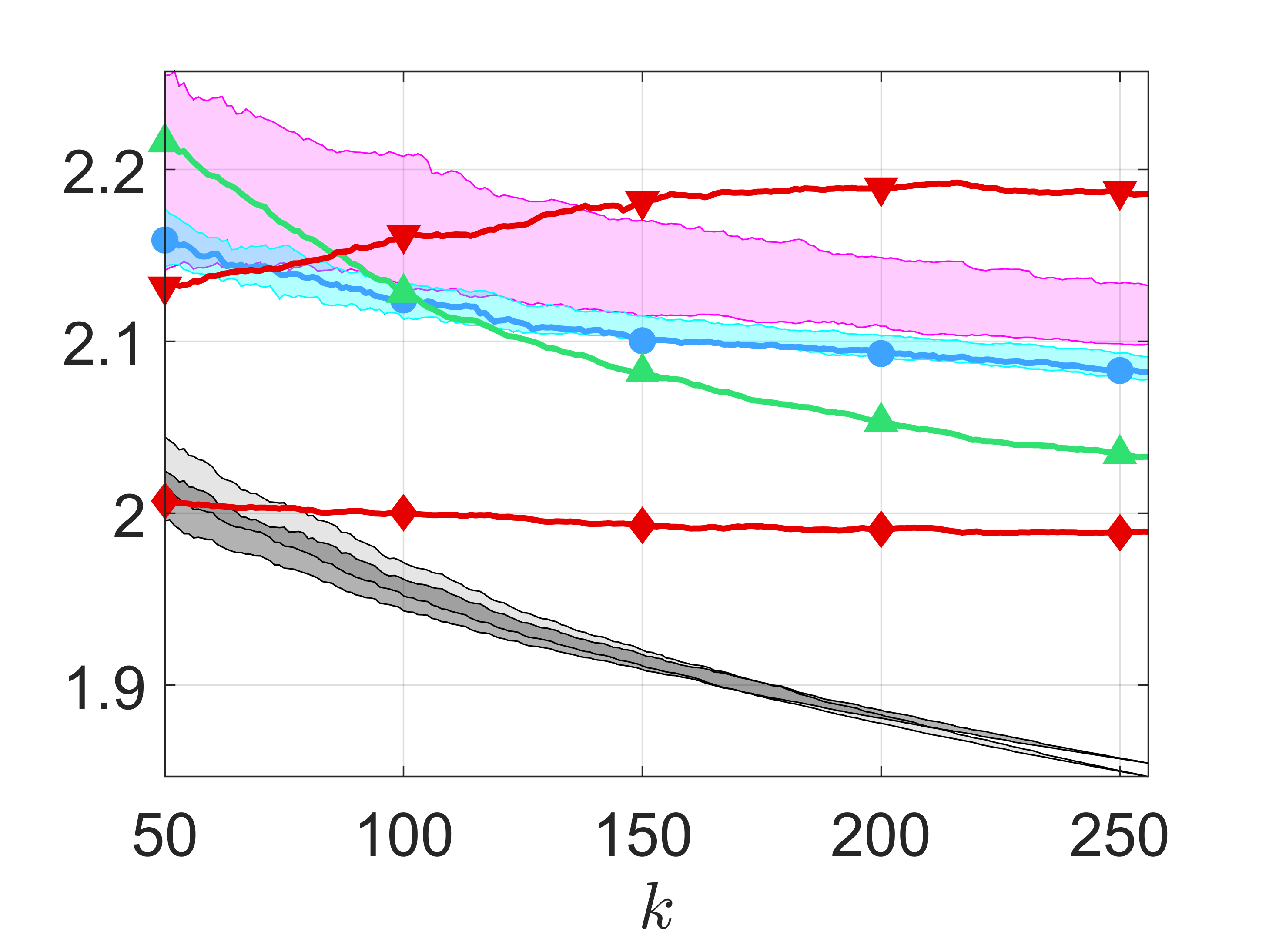}
\end{center}
\caption{\small Like Figure~\ref{F:permut-s2}, but now  $\Xb_n$ based on $\PP_{0,\delta_0^*}$ is a random permutation of $\delta_0^*(d,n,s)\,\Zb_{2^{d-m}}$.
}
\label{F:permut-frac-fact-s2}
\end{figure}

\subsection{Greedy packing applied to a subset of vertices}\label{S:greedy-packing-fract-fact}

Finally, designs based on $\PP_{0,\delta_0^*}$ can be refined by sampling greedily from its support. We use greedy packing: $\xb_1$ is drawn uniformly at random from $\delta_0^*(d,n,s)\Zb_{2^{d-m}}$, and at each subsequent step $k$, $\xb_{k+1}$ is the point in $\delta_0^*(d,n,s)\,\Zb_{2^{d-m}}$ furthest from $\Xb_{1:k}$, with ties broken randomly.

\subsubsection*{$d=10$, $n=2^8=256$}
We continue with the example discussed in the previous two sections. Figure~\ref{F:CH-frac-fact-s2} is analogous to Figure~\ref{F:permut-frac-fact-s2}, but with $\Xb_{1:k}$ generated by greedy packing from the candidate set $\delta_0^*(d,n,s)\,\Zb_{2^{d-m}}$: the designs $\Xb_n$ are the same, but the intermediate designs $\Xb_{1:k}$, $k<n$, tend to perform better. 

\begin{figure}[ht!]
\begin{center}
\includegraphics[width=.49\linewidth]{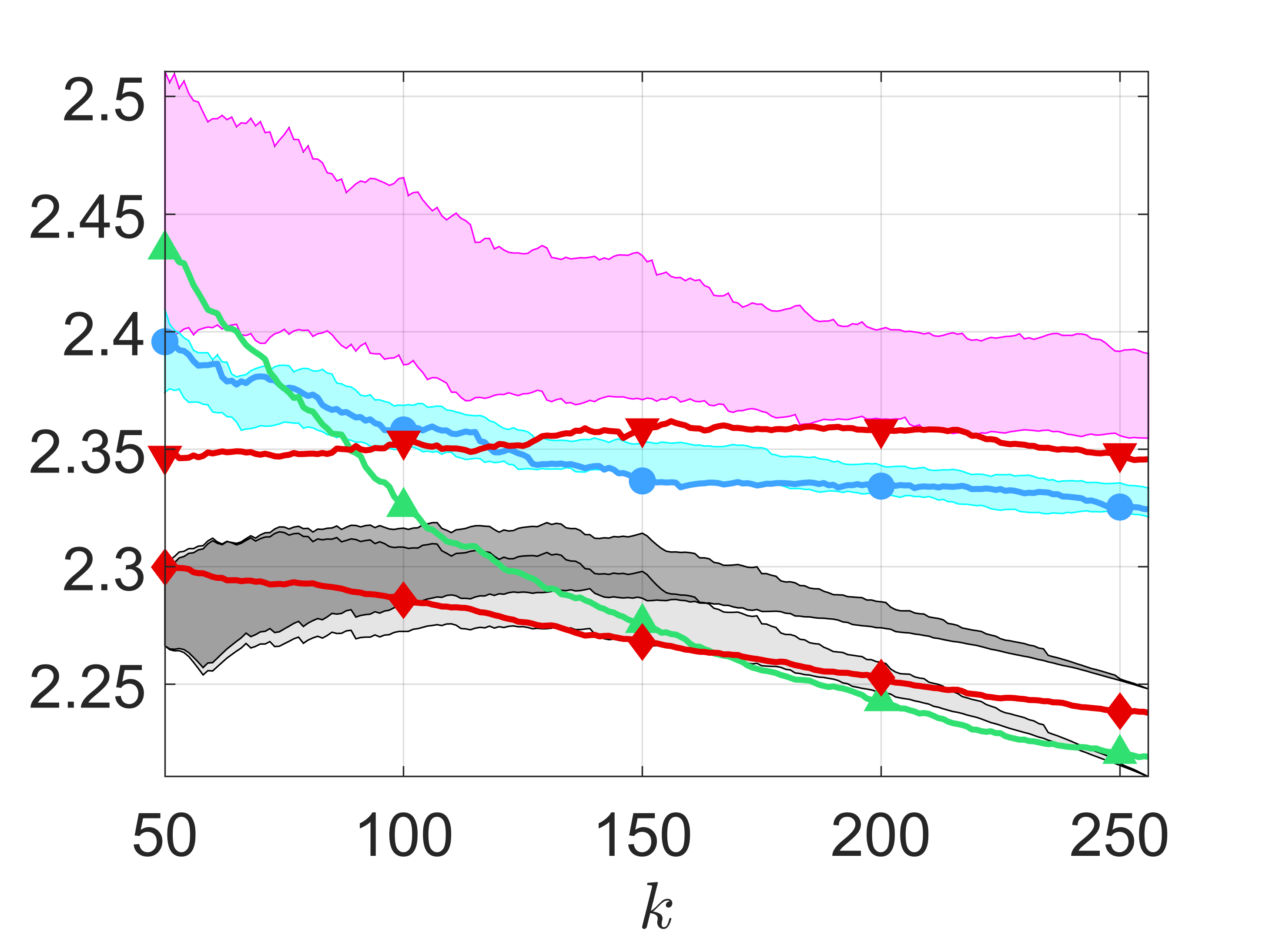}
\includegraphics[width=.49\linewidth]{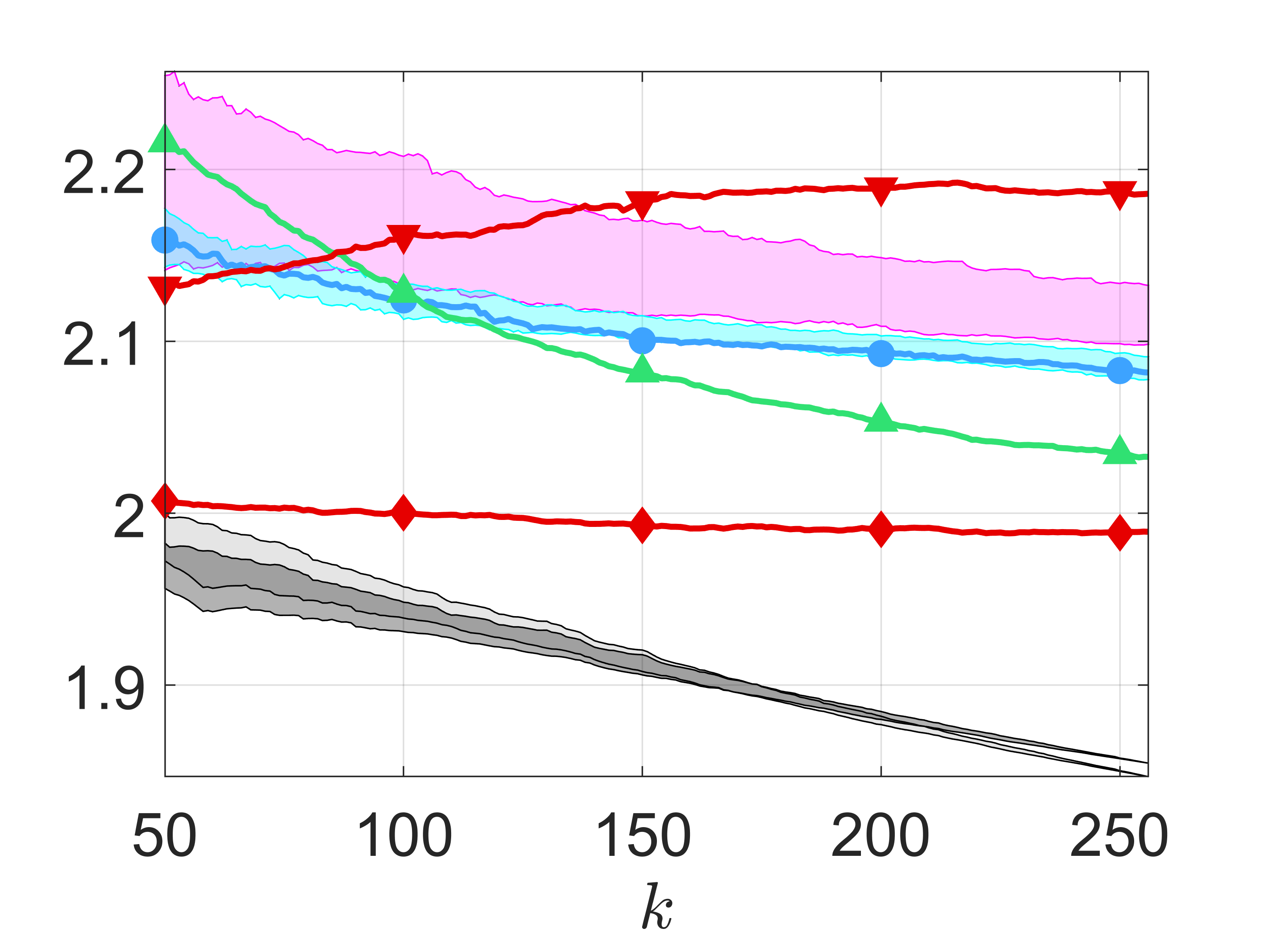}
\end{center}
\caption{\small Like Figure~\ref{F:permut-frac-fact-s2}, but now  $\Xb_{1:k}$ based on $\PP_{0,\delta_0^*}$ is generated by the greedy packing algorithm applied to $\delta_0^*(d,n,s)\,\Zb_{2^{d-m}}$.}
\label{F:CH-frac-fact-s2}
\end{figure}

\subsubsection*{$d=30$, $n=2^{10}=1\,024$}

Owing to the high dimension $d$, greedy packing with boundary avoidance and greedy minimization of the quantization error are no longer computationally feasible. The left panel of Figure~\ref{F:E2_d30_n1024-1} presents $k^{1/d}\,E_{\mu_N,2}(\Xb_{1:k})$ versus $k$ for random designs i.i.d.\ uniform in $\SC_d$, $n$-point prefixes of Sobol' and independent Sobol' sequences, in $\SC_d$ and in $\delta_1^*\,\SC_d$ (with $\delta_1^*=\delta_1^*(d,n,2) \simeq 0.5540$). We also generate designs in the intersection $\SS_{d-1}(\delta_0^*\,\sqrt{d})\cap\SC_d$ (with $\delta_0^*=\delta_0^*(d,n,2) \simeq 0.3191$)---the spherical measure of this intersection is large enough to permit simple rejection sampling. Specifically, points on $\SS_{d-1}(1)$ are obtained by applying the inverse standard normal c.d.f.\ to the first $n'$ points of Sobol' and scrambled Sobol' sequences in $[0,1]^d$, with $n'>n$; each point is then rescaled to have norm $\delta_0^*\,\sqrt{d}$, and we retain the first $n$ points falling in $\SC_d$. Finally, we
construct a $2^{d-m}$ fractional factorial design $\Zb_{2^{d-m}}$, with $m=\lfloor d-\log_2 n \rfloor=20$, having a large minimum Hamming distance ($\rho_H(\Zb_{2^{d-m}})=10$), and construct designs $\Xb_{1:k}$ by greedy packing using the candidate set $\delta_0^*(d,n,s)\,\Zb_{2^{d-m}}$. The variability due to the random choice of $\xb_1$ and to random tie-breaking is negligible and not visible in the figure. 

\begin{figure}[ht!]
\begin{center}
\includegraphics[width=.49\linewidth]{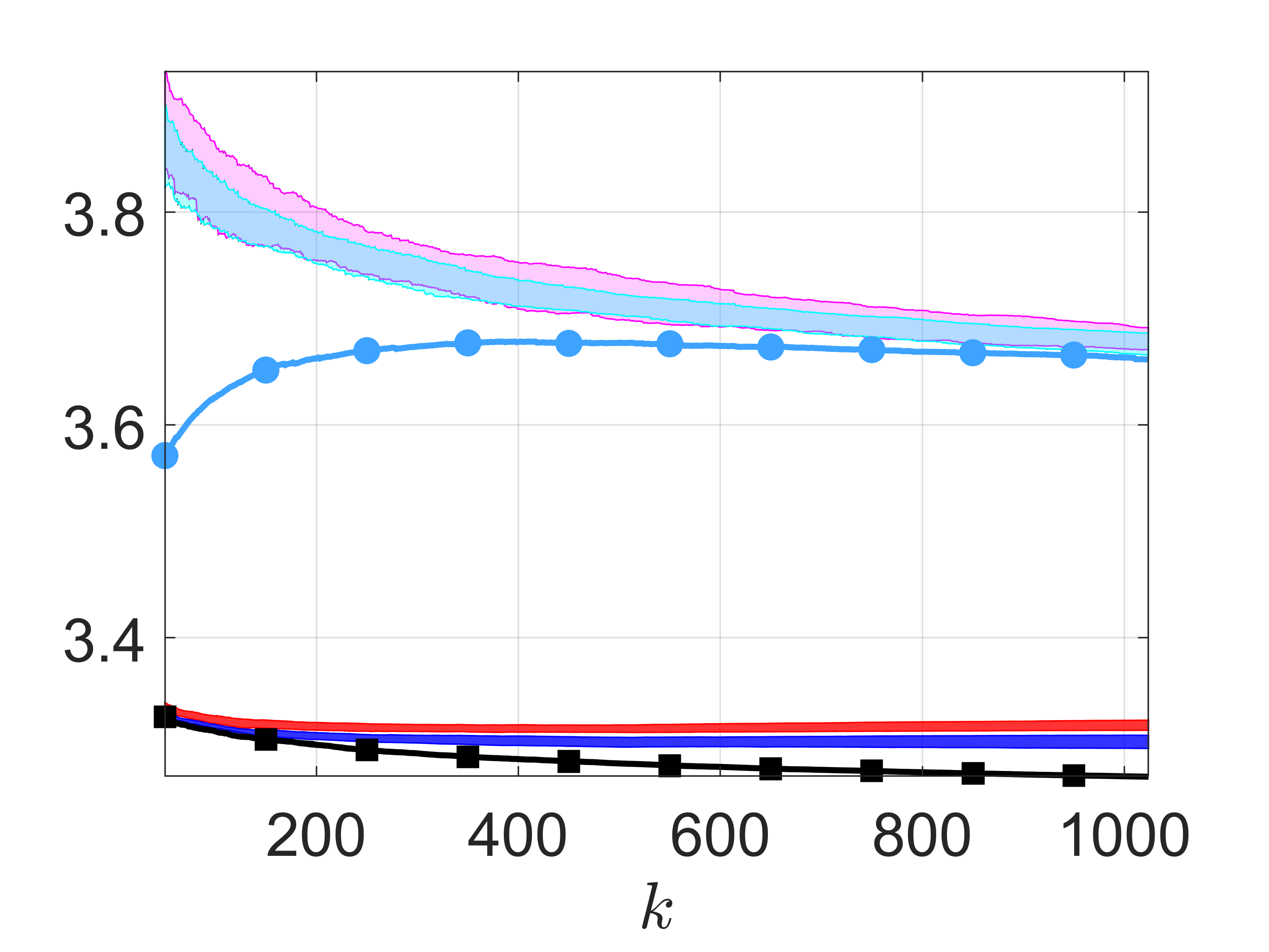}
\includegraphics[width=.49\linewidth]{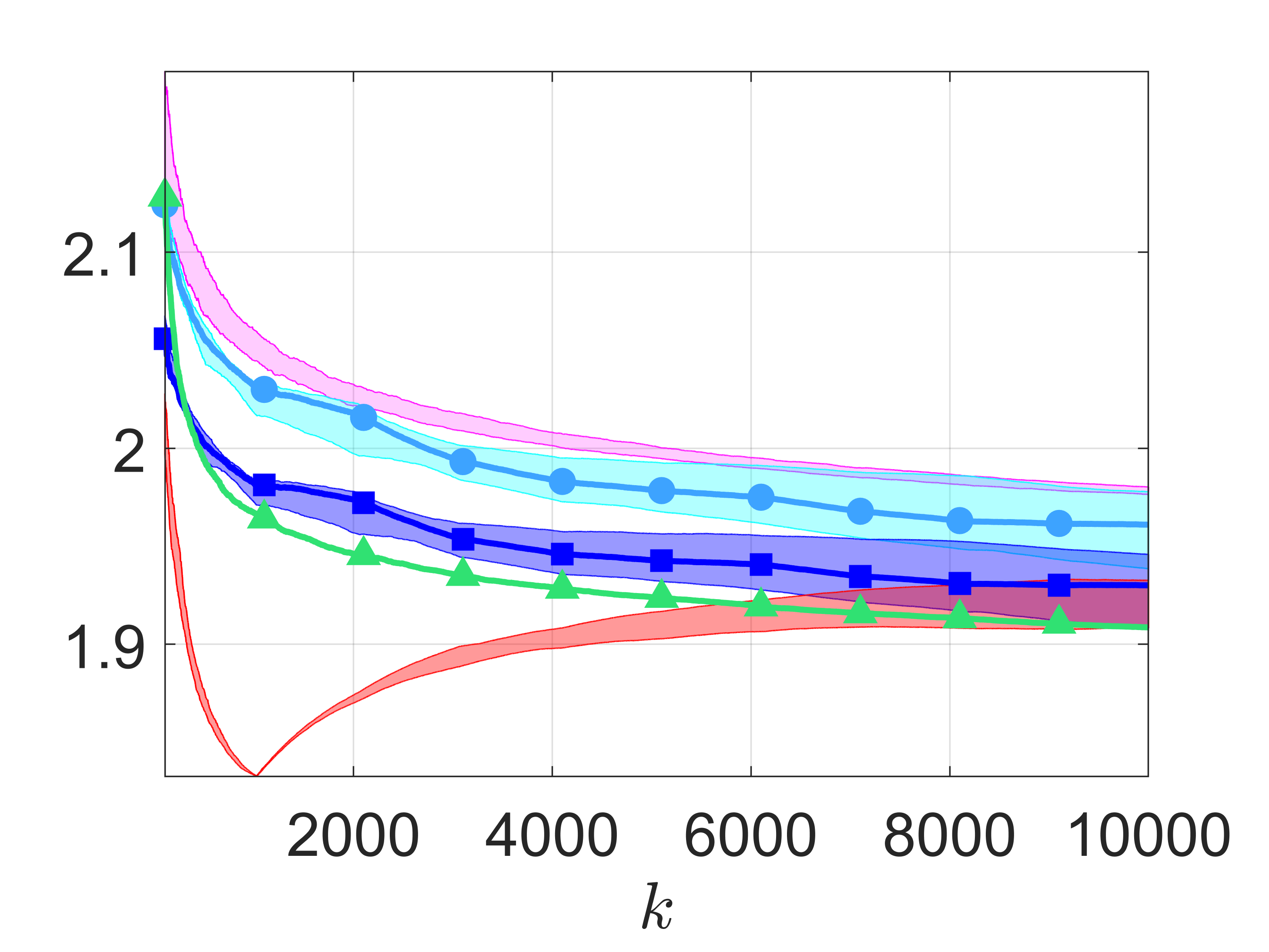}
\end{center}
\caption{\small Left ($d=30$, $n=1\,024$; $\delta_0^*=\delta_0^*(d,n,2)$, $\delta_1^*=\delta_1^*(d,n,2)$): $k^{1/d}\,E_{\mu_N,2}(\Xb_{1:k})$ versus $k$, for 100 random designs i.i.d.\ uniform in $\SC_d$ (magenta envelope); $k$-point prefixes of Sobol' sequence ({\color{Cerulean} $\bullet${\bf ---}$\bullet$}) and 100 independent scrambled Sobol' sequences, in $\SC_d$ (light blue envelope) and in $\delta_1^*\,\SC_d$ (dark blue envelope); $n$-point designs uniform in $\SS_{d-1}(\delta_0^*\,\sqrt{d})\cap\SC_d$ (red envelope); $\Xb_{1:k}$ generated by greedy packing from the candidate set $\delta_0^*\,\Zb_{2^{d-m}}$ ({\tiny $\blacksquare$}{\bf ---}{\tiny $\blacksquare$}).
Right ($d=10$, $n=10\,000$): $k^{1/d}\,E_{\mu_N,2}(\Xb_{1:k})$ versus $k$ for the first three designs in Table~\ref{Tb:color-code-10d_highd} and for $k$-point prefixes of Sobol' sequence ({\color{blue} {\tiny $\blacksquare$}{\bf ---}{\tiny $\blacksquare$}}) and 100 independent scrambled Sobol' sequences (dark blue envelope) in $\delta_1^*(d,n,2)\,\SC_d$. The red envelope corresponds to random permutations of the product quantizer $\{\pm\frac12,\ldots,\pm\frac12\}$ augmented with prefixes of independent scrambled Sobol' sequences in $\delta_1^*(d,n,2)\,\SC_d$.}
\label{F:E2_d30_n1024-1}
\end{figure}

Note the striking difference in quantization performance for small $k$ between the $k$-point prefix of Sobol' sequence and those of scrambled Sobol' sequences (other scrambling methods, e.g., Cranley–Patterson rotation \cite{CranleyP76}, yield similar results): although their filling properties become similar asymptotically, the initial points of the Sobol' sequence exhibit a specific structure that results in lower quantization error for design sizes $k \ll 2^d$. The designs based on $\PP_{1,\delta_1^*}$ (dark blue) and $\PP_{0,\delta_0^*}$ (red and black) perform significantly better than random and Sobol' designs in $\SC_d$. Greedy packing applied to the fractional factorial design $\Zb_{2^{d-m}}$ yields the best overall performance, and designs uniform in $\delta_1^*\,\SC_d$ are preferable to those in $\SS_{d-1}(\delta_0^*\,\sqrt{d})\cap\SC_d$. The plots of $k^{1/d}\,E_{\mu_N,10}(\Xb_{1:k})$ versus $k$ are similar, up to a shift towards larger values.
Using $\delta_0^*(d,n,10)$ and $\delta_1^*(d,n,10)$ instead of $\delta_0^*(d,n,2)$ and $\delta_1^*(d,n,2)$ has little effect, since $\delta_0^*$ and $\delta_1^*$ vary only slightly with $s$ 
(see the left panel of Figure~\ref{F:delta_s1-15_alpha0_d10-50_n50d} for $\delta_0^*$ and the comments after the figure for the evolution of $\delta_1^*$):
we have $\delta_0^*(d,n,10) \simeq 0.3269$ and $\delta_1^*(d,n,10) \simeq 0.5676$. 
\subsection{Designs with size above $2^d$}\label{S:n>2^d}

When the design size can reach $n>2^d$ we cannot restrict our attention to the vertices of $\delta_0^*\,\SC_d$. A first option is to take $\Xb_k$ as the $k$-point prefix of a random design, or a (scrambled) Sobol' sequence, in $\delta_1^*\,\SC_d$. The right panel of Figure~\ref{F:E2_d30_n1024-1} provides an illustration for $d=10$ and $n=10\,000$: designs uniform in $\delta_1^*\,\SC_d$ 
perform significantly better than designs uniform in $\SC_d$. 
Here, $\delta_1^*=\delta_1^*(d,n,2)\simeq 0.9046$. 
Although greedy packing with boundary avoidance performs well, prefixes of the Sobol' (or a scrambled Sobol') sequence on $\delta_1^*\,\SC_d$ constitute a valuable alternative, owing to the simplicity of their construction. 
These prefixes can also be used to augment the full factorial design $\Xb_{2^d}^*$ of Remark~\ref{Rk:product quantizer}, yielding even better performance.

\subsection{Accounting for low-dimensional projections}\label{S:projections}

By projection, we mean projection onto a canonical subspace of dimension $p<d$.
For any $n$-point design $\Xb_n$, let $\{\Xb_n\}_\SJ$ denote the projection of $\Xb_n$ onto the coordinates in $\SJ\subset\{1,\ldots,d\}$, with $|\SJ|=p$.  

The designs from Sections~\ref{S:sampling-without-R}-\ref{S:n>2^d} are concentrated in a cube $\delta\,\SC_d$ with $\delta<1$, and even $\delta<1/2$ when $d$ is large and $n \ll 2^d$, 
see the left panel of Figure~\ref{F:delta_s2_alpha0_d10-50_n10d-100d}.
Consequently, their projections onto low-dimensional subspaces are poor: $\{\Xb_n\}_\SJ\in\delta\SC_p$ for every $p\in\{1,\ldots,d\}$. 
In particular,
when $\Xb_n$ is based on $\PP_{0,\delta_0^*}$ (Sections~\ref{S:sampling-without-R} to \ref{S:greedy-packing-fract-fact}), its projections on a $p$-dimensional subspace contain at most $n_p=\min\{n,2^p\}$ distinct points.

The method below builds on the greedy-packing approach of Section~\ref{S:greedy-packing-fract-fact}, which uses a $2^{d-m}$ fractional factorial design $\Zb_{2^{d-m}}$ as the candidate set, with $m=\lfloor d-\log_2 n \rfloor$ and $n$ the maximum targeted design size. We first generate points $\xb_k'$, $k=1,\ldots,n$ in $\{-1,1\}^d$ via greedy-packing: $\xb_1'$ is a random element of $\Zb_{2^{d-m}}$ and $\xb_{k+1}$ is the point in $\Zb_{2^{d-m}}$ furthest from $\Xb_{1:k}'=\{\xb_1',\ldots,\xb_k'\}$, with ties broken randomly. The actual design points are  then obtained by componentwise scaling: $\xb_k = \xb_k'\circ\yb_k$ for some  $\yb_k\in[0,1]^d$, where $\circ$ denotes the Hadamard product (i.e.\ $(\xb\circ\yb)_i=x_i y_i$, $i=1,\ldots,d$).

Let $\varphi_y(\cdot)$ be a p.d.f.\ supported on $[0,1]$ and let $F_y(\cdot)$ denote its c.d.f. A first option is to use i.i.d.\ random vectors $Y_i \simd \varphi_y$. Alternatively, one may set $\yb_k=F_y^{-1}(\bs_k)$, $k=1,2,\ldots$, where $\bs_k$ is the $k$-point prefix of Sobol' sequence in $[0,1]^d$ and $F_y^{-1}$ (the inverse c.d.f.) is applied componentwise.
To keep $E_{\mu,s}(\Xb_{1:k})$ small (i.e., to obtain space-filling designs), $\varphi_y$ should be peaked at $\delta_0^*$. We propose 
\be
\label{phiz-beta}
\varphi_y(t)=\varphi(t;a,b|1)
\ee
given by \eqref{eq:beta_general} with
\be
\label{a-phiz-beta}
a=1+\frac{\delta_0^*(b-1)}{1-\delta_0^*}\,,
\ee
so that $\dd\varphi_y/\dd t=0$ at $t=\delta_0^*$. Hence, $\varphi_y$ attains its maximum at $\delta_0^*$ for $b>1$. We take $b\in[1,\infty)$: when $b=1$, $\yb_k$ is uniform in $[0,1]^d$, while $\yb_k \to \delta_0^* (1,\ldots,1)$ as $b\to\infty$.

In the following example, we compare quantization performance, in the full space and in projections, across different values of $b$ in (\ref{phiz-beta}–\ref{a-phiz-beta}), highlighting the inherent trade-off between the two.


Consider again the case $d=30$, $n=2^{10}=1\,024$. The fractional factorial design $\Zb_{2^{d-m}}$ of Section~\ref{S:greedy-packing-fract-fact}, with minimum Hamming distance $\rho_H(\Zb_{2^{d-m}})=10$, has resolution IV, with 16 defining relations of length 4 (see, e.g., \cite{BoxH61} and \cite[Chap.~5]{WuH2009} for definitions). Since $m=10$, every $p$-dimensional projection of $\Zb_{2^{d-m}}$ involving only the first 10 variables yields a $2^p$ full factorial design; moreover, any projection with $p\leq 3$ is also a $2^p$ full factorial design. For $p=4$, out of the $27\,405=\binom{30}{4}$ possible 4-dimensional projections, all but 16 ($\binom{30}{4} -16=27\,389$) are $2^p$ full factorial designs. This illustrates the benefit of choosing a design with high resolution $R$ and few defining relations of length $R$. However, a trade-off with the minimum Hamming distance is necessary: we could find designs with less than 16 defining relations of length 4, but their minimum Hamming distance drops to only~9. 

Figure~\ref{F:Xn_d30_n1024_proj_29-30} shows the projections $\{\Xb_n\}_\SJ$ of different designs $\Xb_n$ onto the subspace defined by $\SJ=\{29, 30\}$  (the reason for this choice is that, for the Sobol' sequence (top row, left), the projection onto the last two coordinates is less even than the projections onto pairs of variables with smaller indices). The center panel shows quasi-random points with distribution $\PP_{1,\delta_1^*}$, and the right panel those with $\PP_{0,\delta_0^*}$.
On the bottom row, the designs are generated as $\xb_k = \xb_k'\circ F_y^{-1}(\bs_k)$, where $\xb_k'$ is obtained by greedy packing applied to $\Zb_{2^{d-m}}$, $\bs_k$ is the $k$-th Sobol' point in $[0,1]^d$, and $F_y$ is the c.d.f.\ of the beta density \eqref{phiz-beta} with $a$ from \eqref{a-phiz-beta}, for three values of $b$. When $b=1$ (left), $\yb_k=\bs_k$, and the projection $\{\Xb_n\}_{\{29,30\}}$ is more evenly spread over $\SC_2$ than for the standard Sobol' sequence (top left). As $b$ increases (center and right), $\{\Xb_n\}_{\{29,30\}}$ becomes increasingly concentrated around the four points $(\pm \delta_0^*,\pm \delta_0^*)$.

\begin{figure}[ht!]
\begin{center}
\includegraphics[width=.32\linewidth]{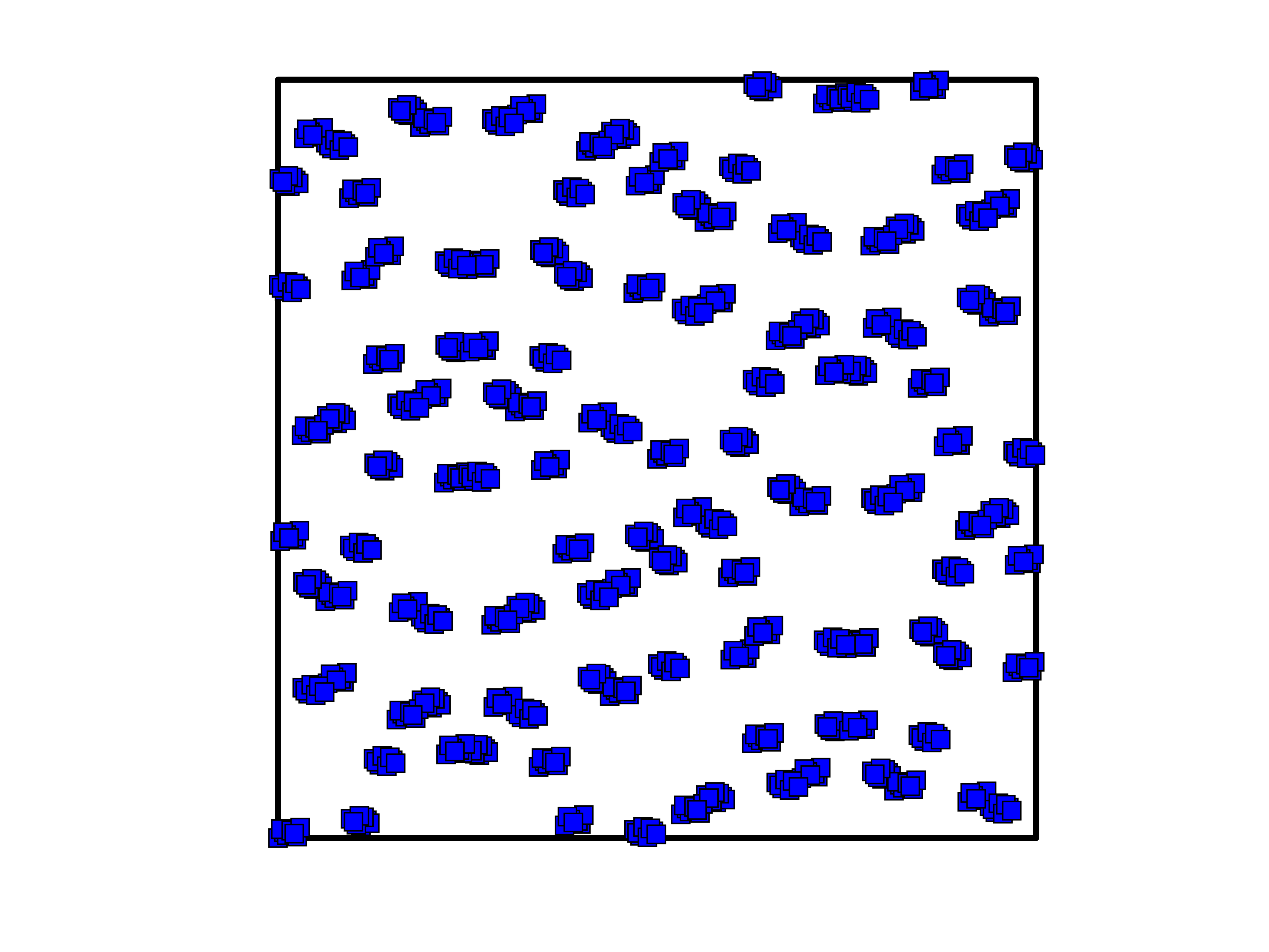} 
\includegraphics[width=.32\linewidth]{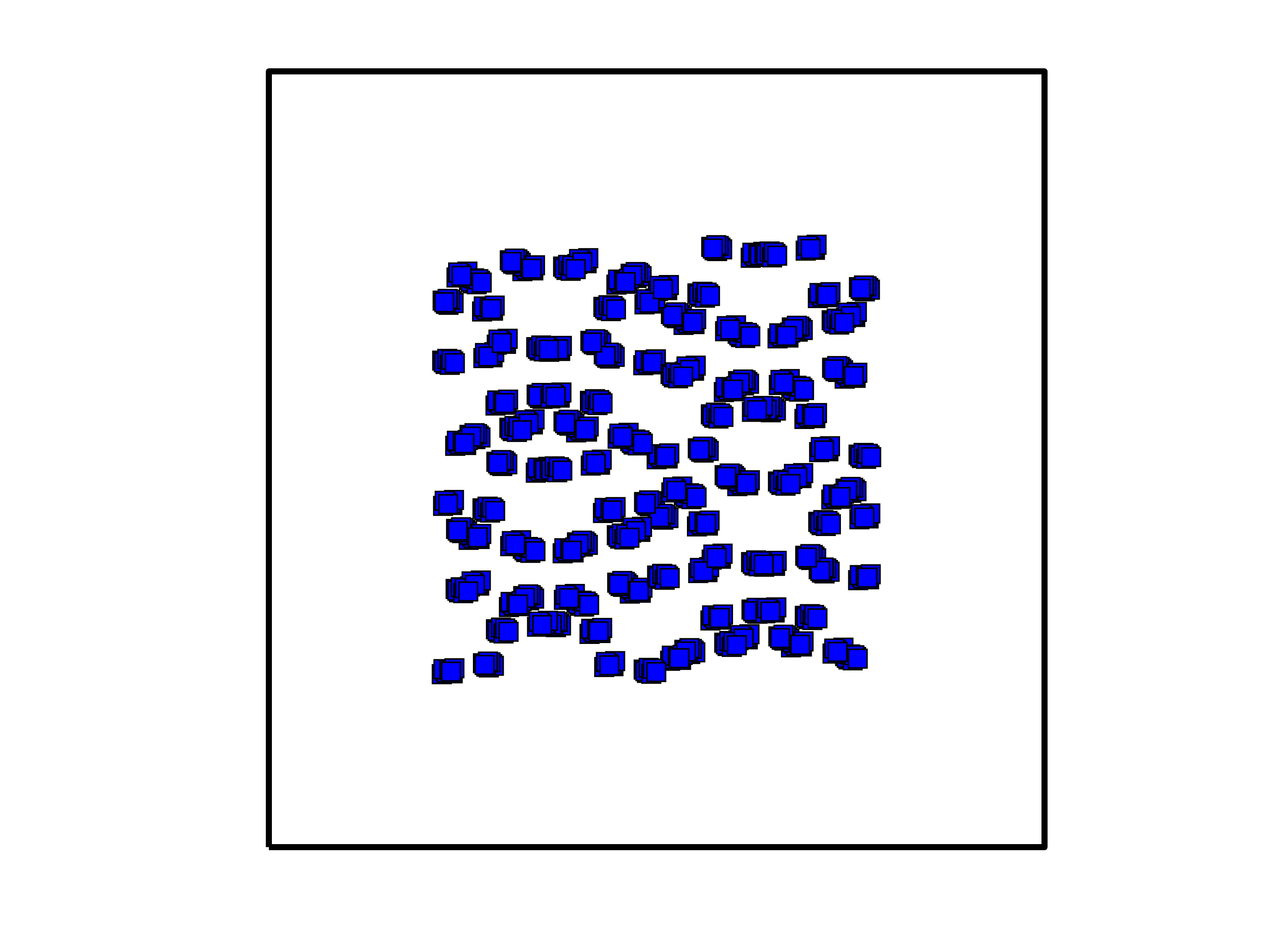}\includegraphics[width=.32\linewidth]{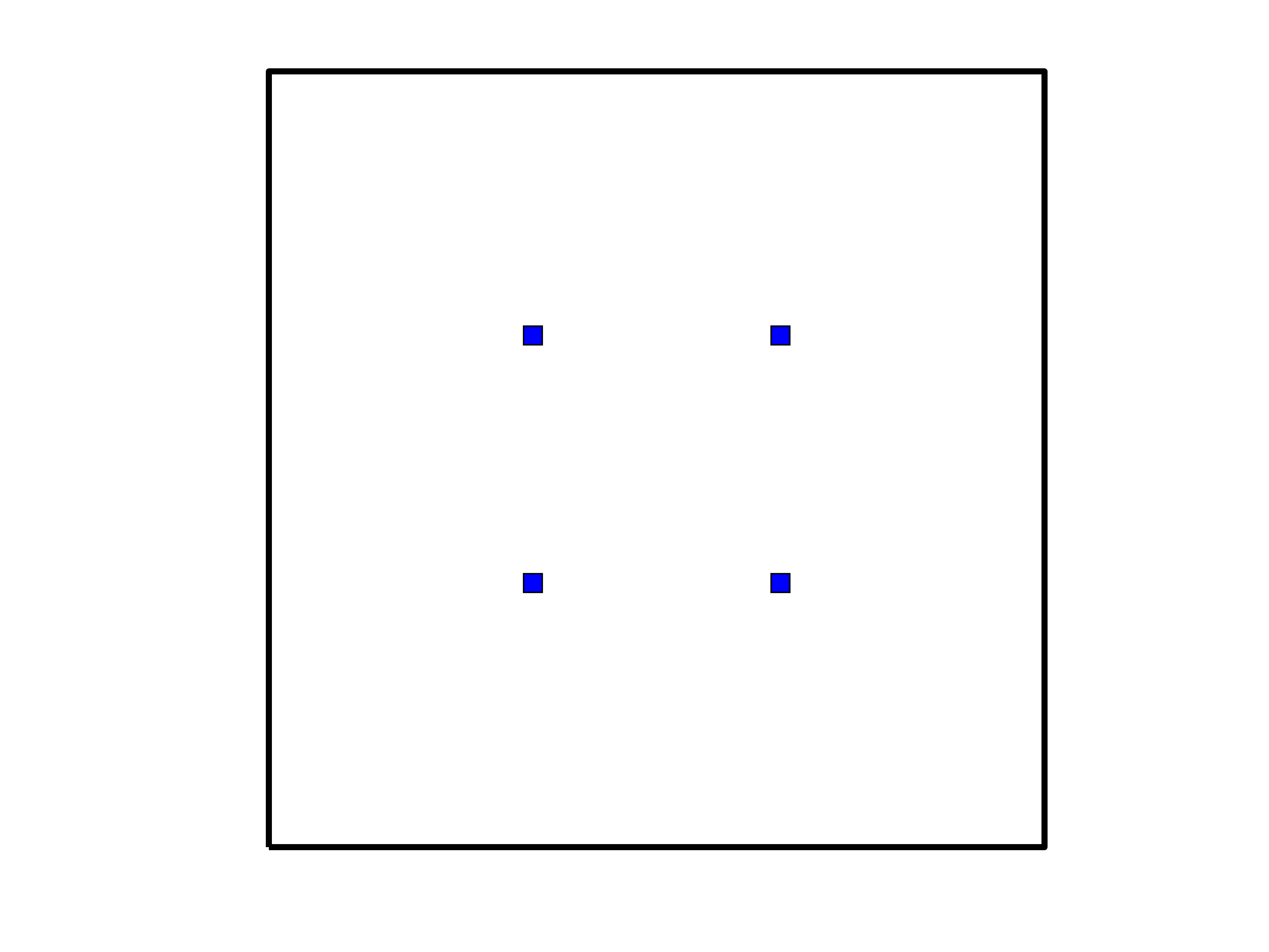}
\\
\includegraphics[width=.32\linewidth]{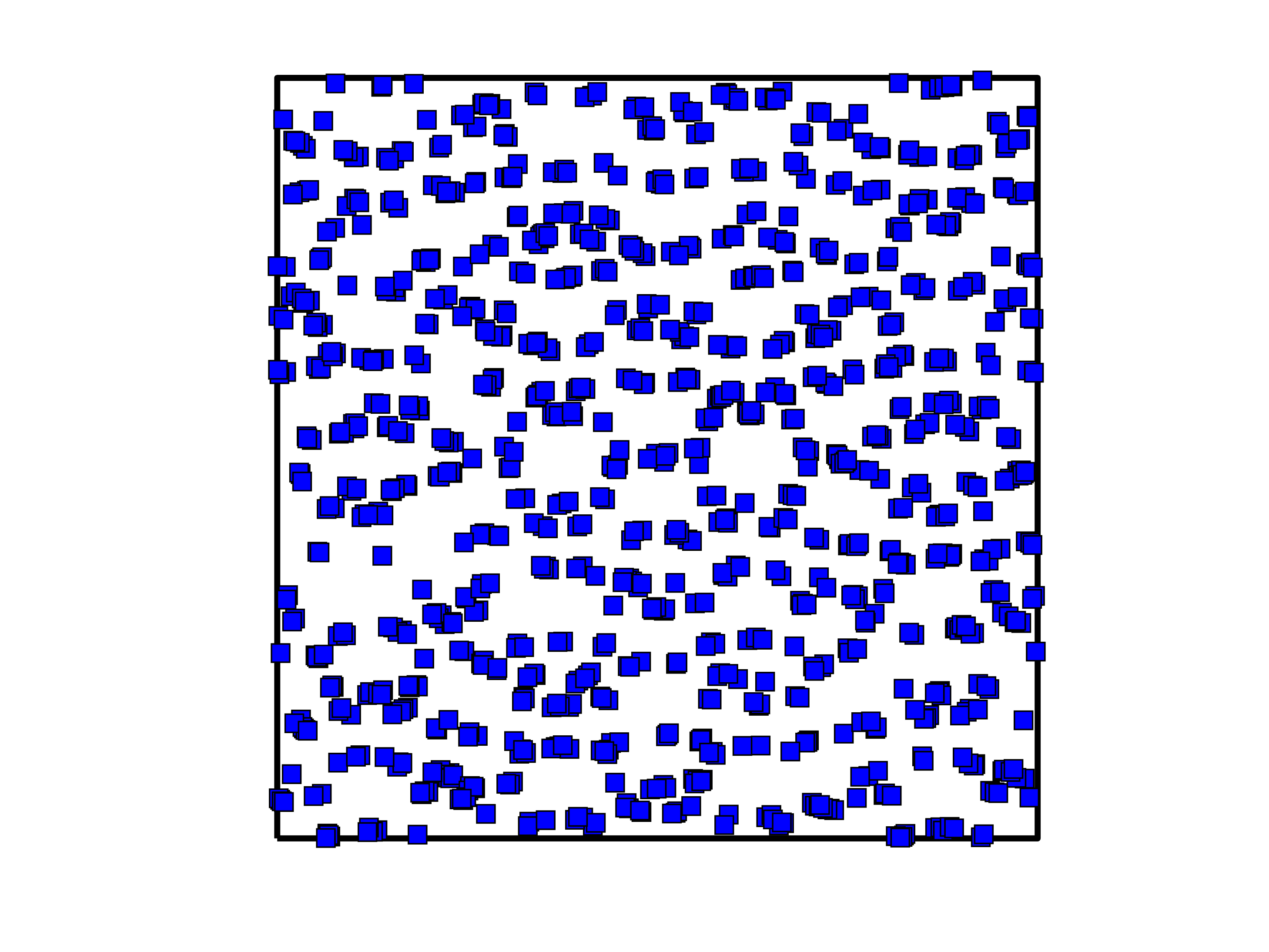}
\includegraphics[width=.32\linewidth]{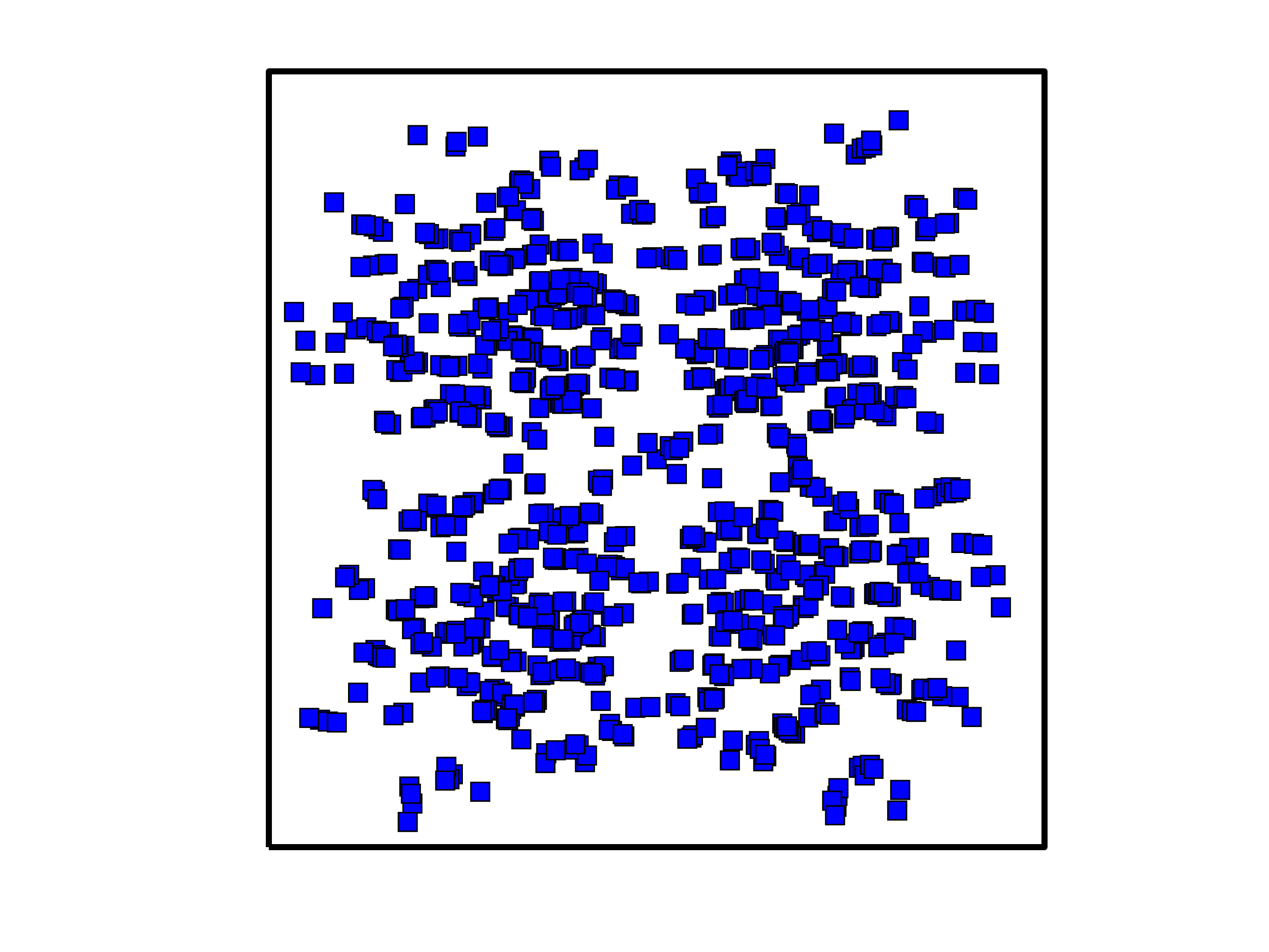}
\includegraphics[width=.32\linewidth]{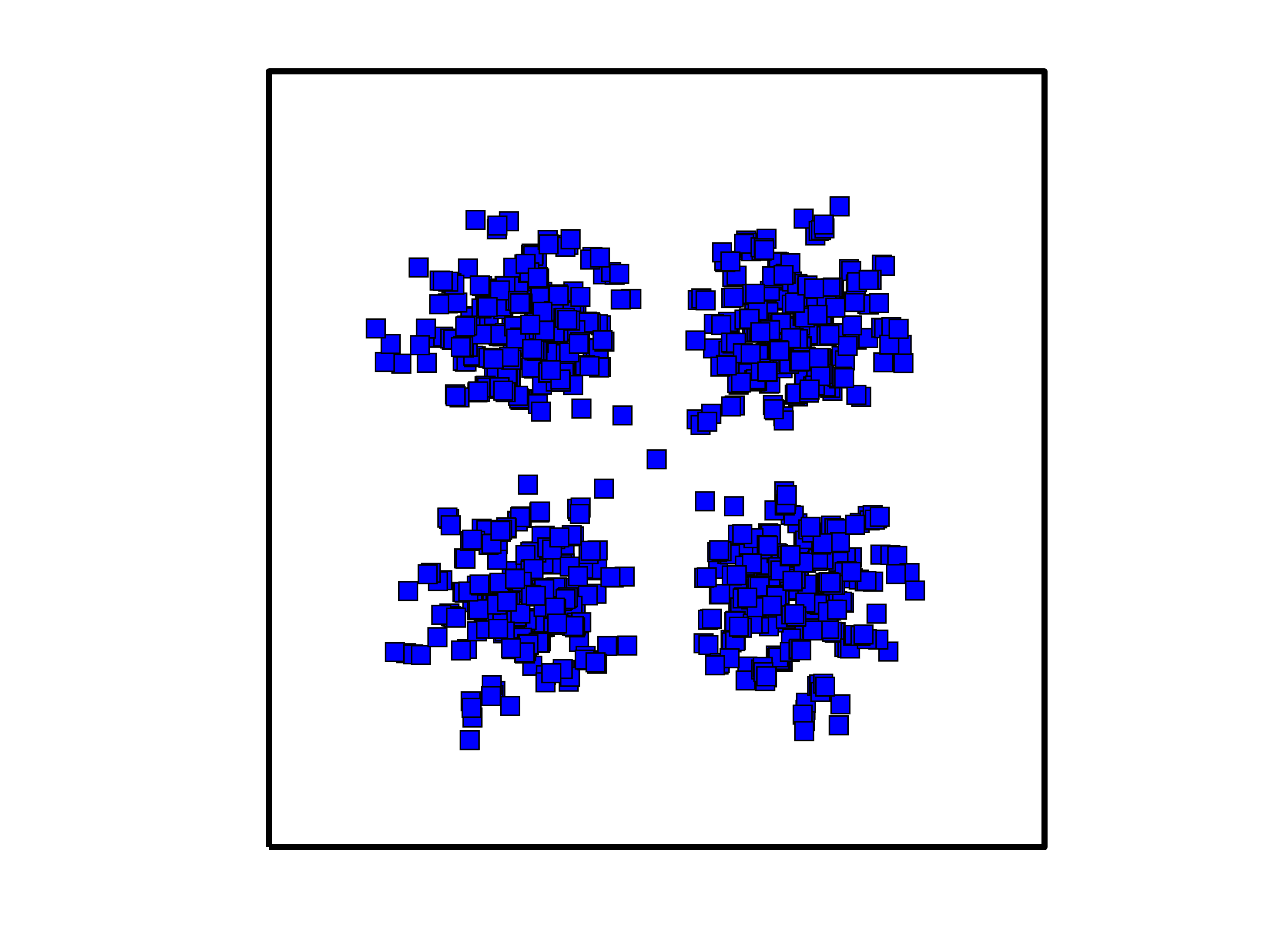}
\end{center}
\caption{\small Design projections $\{\Xb_n\}_{\{29,30\}}$. First row: prefixes of Sobol' sequence in $\SC_d$ (left) and in $\delta_1^*\,\SC_d$ (center), $\delta_0^*\,\Zb_{2^{d-m}}$ (right).
Second row: $\xb_k = \xb_k'\circ F_y^{-1}(\bs_k)$, $k=1,\ldots,n$, with $\bs_k$ the $k$-th element of Sobol' sequence in $[0,1]^d$ and $\varphi_y$ given by (\ref{phiz-beta}–\ref{a-phiz-beta}) with $b=1$ (left), $b=3$ (center) and $b=10$ (right).}
\label{F:Xn_d30_n1024_proj_29-30}
\end{figure}

Figure~\ref{F:E2_d30_n1024_proj_d4} plots $k^{1/p} E_{\mu_{N|p},2}(\{\Xb_k\}_\SJ)$ versus $k$ for different designs, with $p=|\SJ|=4$ and $\mu_{N|p}$ uniform on the first $p$ components of the points supporting $\mu_N$. On the left panel, $\SJ=\{27, 28, 29, 30\}$, corresponding to a 4-dimensional subspace for which the projection $\{\Zb_{2^{d-m}}\}_\SJ$ is a 16-point full factorial design repeated 64 times. On the right panel, $\SJ=\{2, 4, 14, 22\}$, corresponding to one of the 16 poor 4-dimensional projections out of $\binom{30}{4}=27\,405$: $\{\Zb_{2^{d-m}}\}_\SJ$ is a half fraction (see \cite{BoxH61}) repeated 128 times. Therefore, for any $b$, the designs $\{\Xb_k\}_\SJ$ only occupy half of the 4-dimensional cube; hence the larger $L_s$-mean quantization error observed on the right panel compared to the left one. 

\begin{figure}[ht!]
\begin{center}
\includegraphics[width=.49\linewidth]{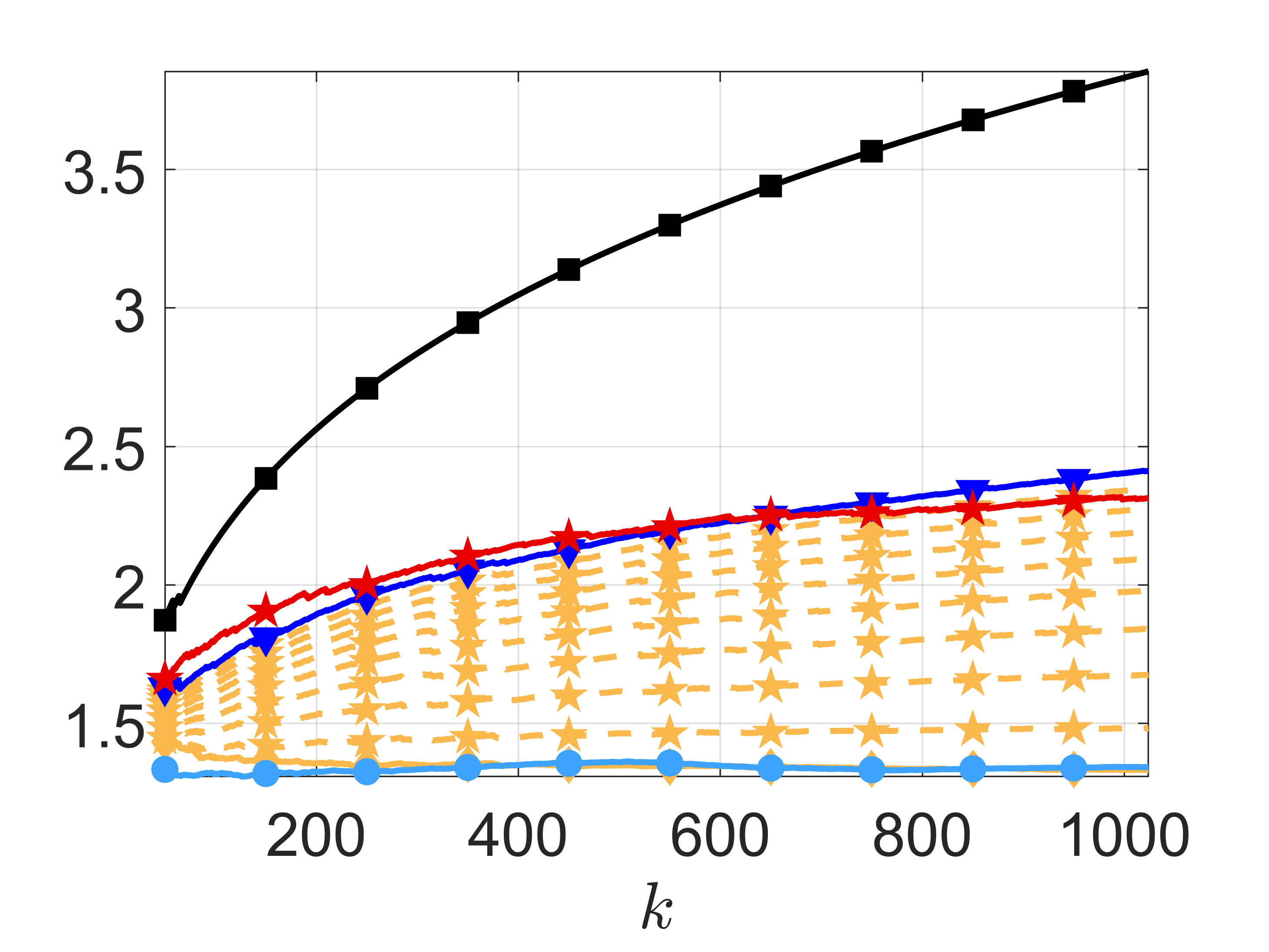}
\includegraphics[width=.49\linewidth]{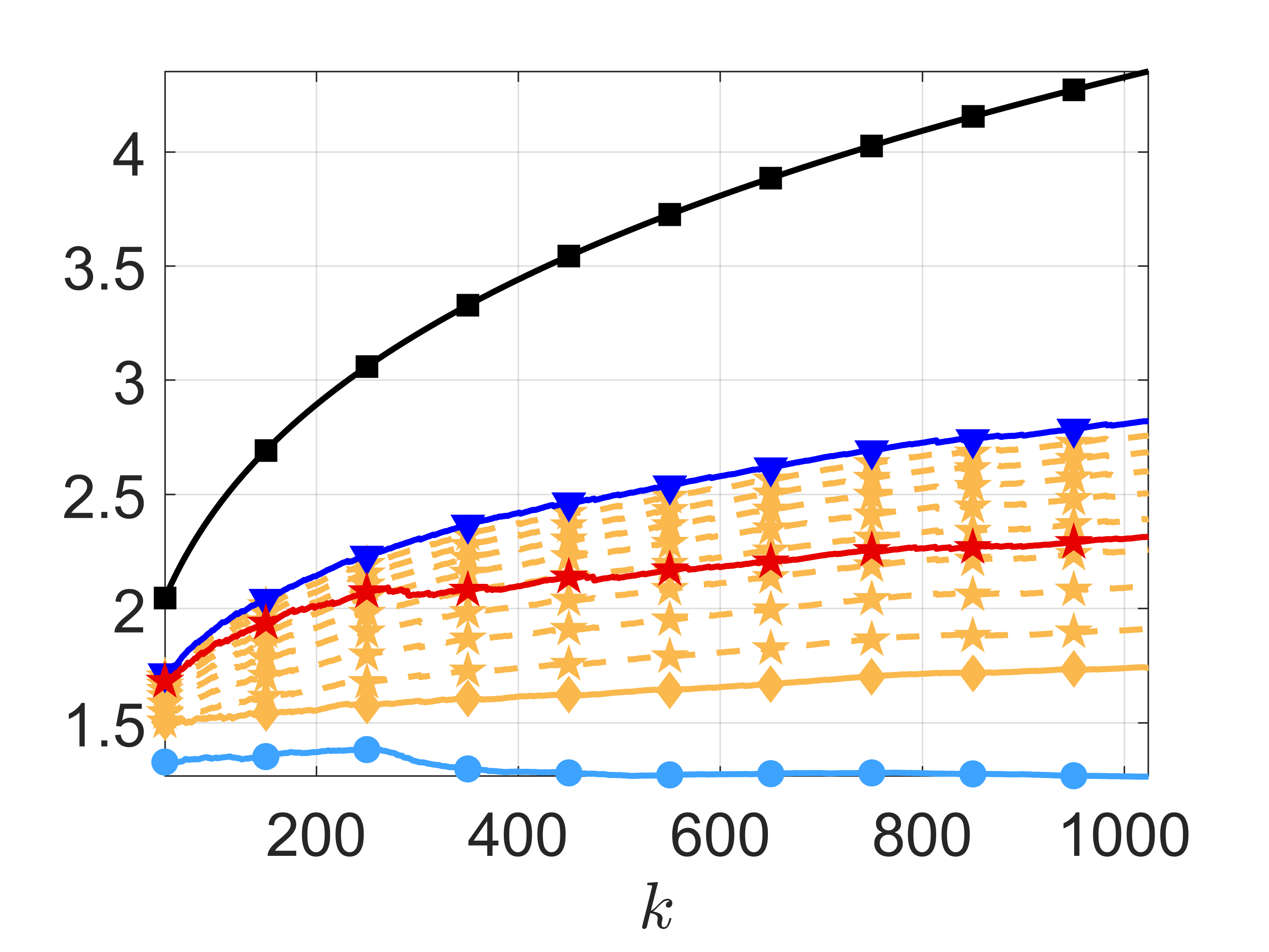}
\end{center}
\caption{\small $k^{1/4} E_{\mu_{N,4},2}(\{\Xb_k\}_\SJ)$ versus $k$ for $\SJ=\{27, 28, 29, 30\}$ (left) and $\SJ=\{2, 4, 14, 22\}$ (right):  $k$-point prefixes of Sobol' sequence ({\color{Cerulean} $\bullet${\bf ---}$\bullet$}),
$\Xb_{1:k}$ generated by greedy packing from the candidate set $\delta_0^*(d,n,2)\,\Zb_{2^{d-m}}$ ({\tiny $\blacksquare$}{\bf ---}{\tiny $\blacksquare$}),
quasi random (Sobol') design uniform in $\SS_{d-1}(\delta_0^*\,\sqrt{d})\cap\SC_d$ 
({\color{red} $\bigstar${\bf ---}$\bigstar$}),
and $\Xb_{1:k}=\{\xb_1,\ldots,\xb_k\}$ where $\xb_k = \xb_k'\circ F_y^{-1}(\bs_k)$, $\bs_k$ is the $k$-th element of Sobol' sequence in $[0,1]^d$, $\varphi_y$ is given by \eqref{phiz-beta} with $a$ as in \eqref{a-phiz-beta}, and $b=1$ ({\color{Dandelion} $\blacklozenge${\bf ---}$\blacklozenge$}), $b=2,\ldots,9$ ({\color{Dandelion} $\bigstar${\bf - - -}$\bigstar$}), $b=10$ ({\color{blue} $\blacktriangledown${\bf ---}$\blacktriangledown$}).  }
\label{F:E2_d30_n1024_proj_d4}
\end{figure}

Figure~\ref{F:E2_d30_n1024_proj_b1-10} shows the full-space quantization performance (for $s=10$ and $s=2$) of the same designs as in Figure~\ref{F:E2_d30_n1024_proj_d4}. The quasi-random design uniform on $\SS_{d-1}(\delta_0^*\,\sqrt{d})\cap\SC_d$ (see Section~\ref{S:greedy-packing-fract-fact} for $d=30$ and $n=2^{10}$) achieves a trade-off between quantization in $\SC_d$ and in its four-dimensional projections, but offers no tuning parameter to control it. In contrast, designs $\Xb_{1:k}$ generated by $\xb_k = \xb_k'\circ F_y^{-1}(\bs_k)$ allow such control through the parameter $b$ in \eqref{a-phiz-beta}: for $b=1$, performance is similar to that of Sobol' prefixes in $\SC_d$ (good in projections, weak in the full space), whereas for large $b$ it approaches that of greedy-packing designs based on $\delta_0^*\,\Zb_{2^{d-m}}$ (excellent in the full space, weak in projections).

\begin{figure}[ht!]
\begin{center}
\includegraphics[width=.49\linewidth]{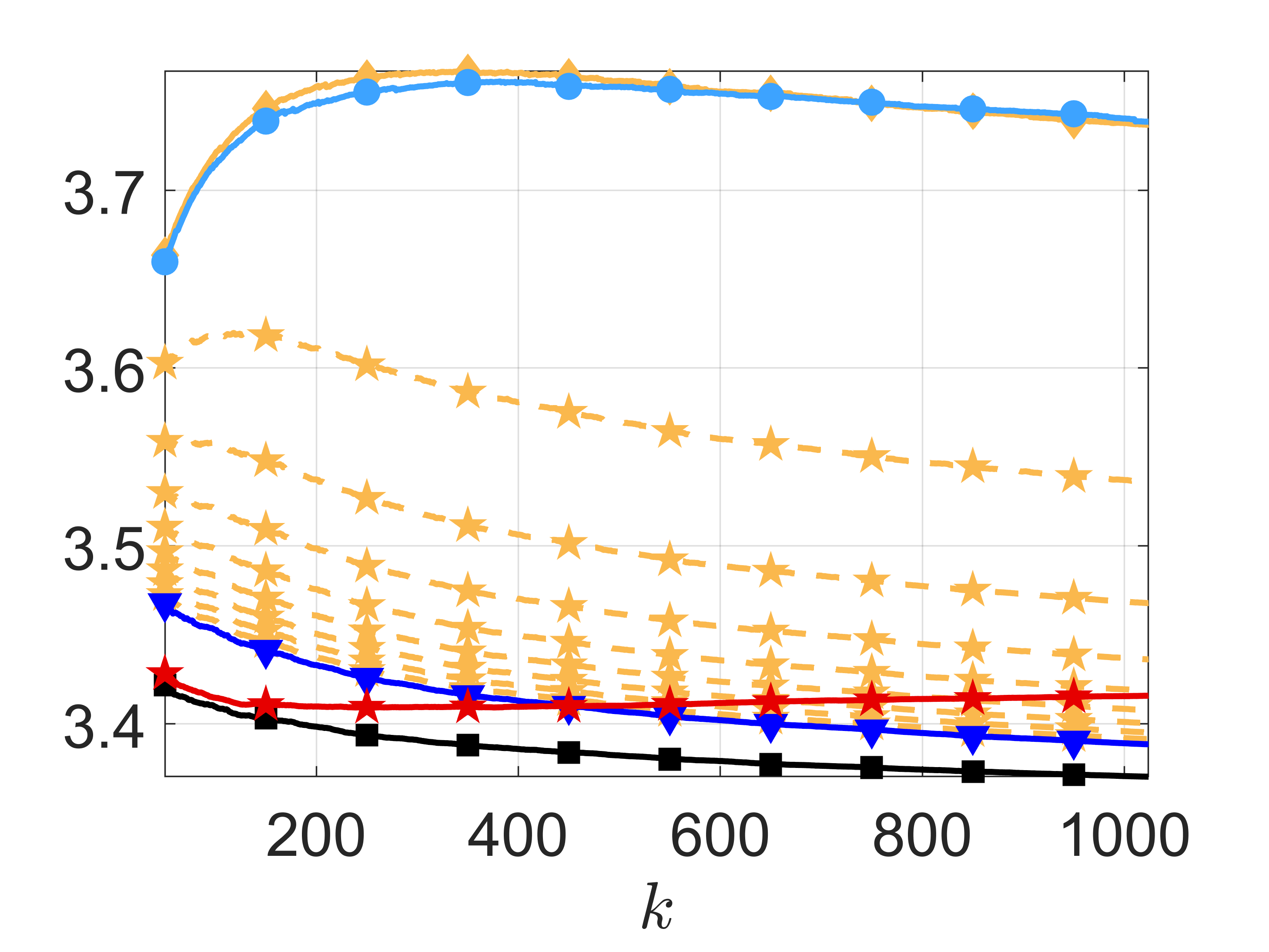}
\includegraphics[width=.49\linewidth]{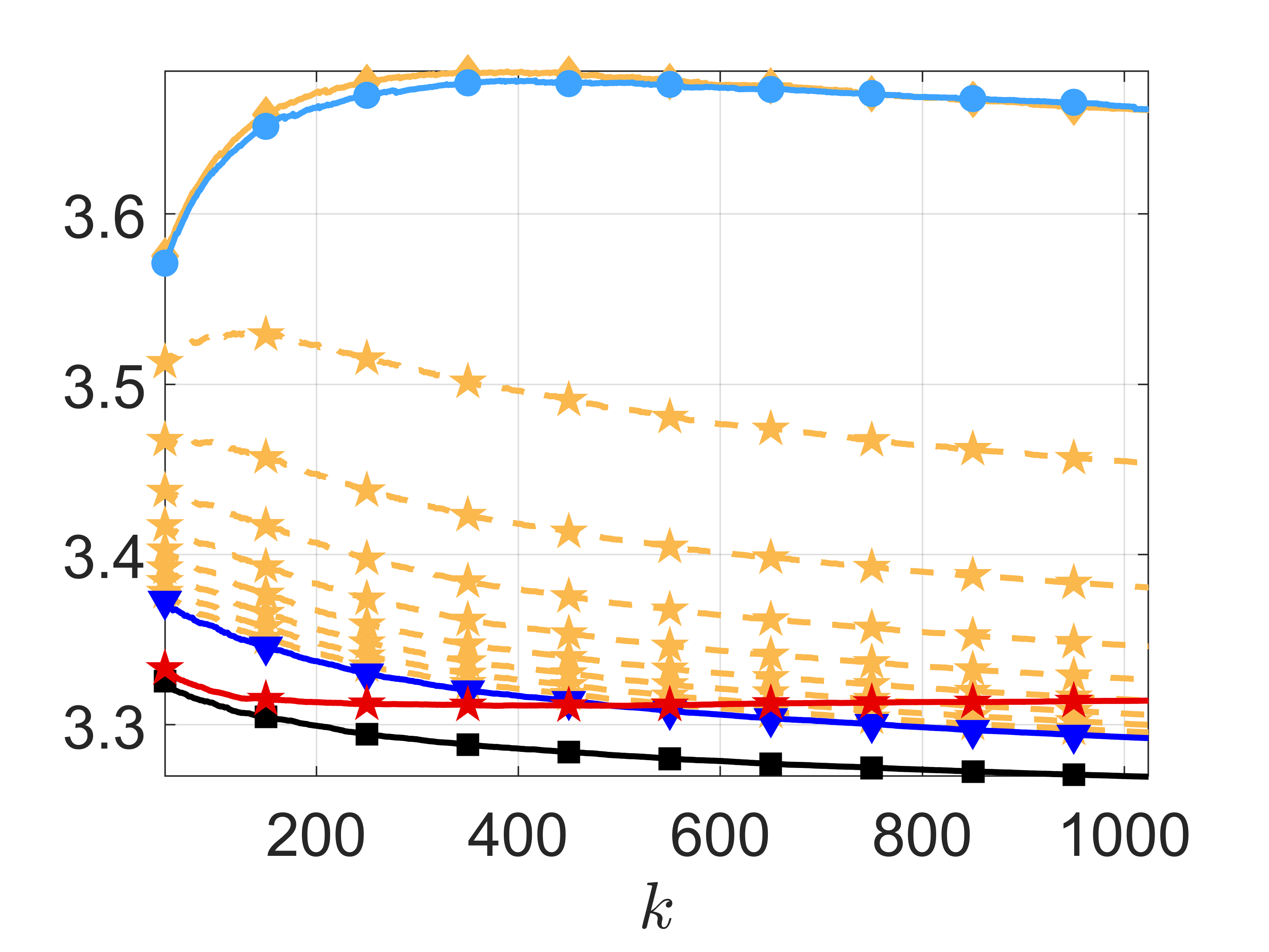}
\end{center}
\caption{\small $k^{1/d} E_{\mu_N,10}(\Xb_k)$ (left) and $k^{1/d} E_{\mu_N,2}(\Xb_k)$ (right) versus $k$ for the same designs as in Figure~\ref{F:E2_d30_n1024_proj_d4}.}
\label{F:E2_d30_n1024_proj_b1-10}
\end{figure}

\section{Conclusions}~\label{S:conclusion}

We have proposed a methodology for generating incremental designs in high-dimensional hypercubes $\SC_d=[-1,1]^d$, based on minimizing the $L_s$-mean quantization error of the uniform distribution. By considering spherically symmetric approximations of product measures on the hypercube, we construct an approximation of the mean distortion of a random design with i.i.d.\ points, which can be minimized with respect to their distribution. For design sizes $n$ growing slower than exponentially in the dimension $d$, the optimal distribution is supported on the vertices of a smaller hypercube $\delta_0^*\SC_d$. Sampling these vertices, or a suitable subset of them when $n<2^d$ (e.g., via a fractional factorial design), possibly through a deterministic algorithm such as greedy packing, yields incremental designs with strong quantization performance. Since sampling hypercube vertices necessarily produces designs with poor low-dimensional projections, a perturbation can be applied, with a parameter that sets the trade-off between quantization performance in the full hypercube and in lower-dimensional projections (Section~\ref{S:projections}). The optimal $\delta_0^*$, which depends on $d$, $n$, and $s$, can be determined numerically by minimizing a function expressed as a double integral (see \eqref{D0}); for large enough $n$, an extreme-value approximation $\widehat\delta_0^*$ can be used, with an explicit form when $s=2$ (see \eqref{EV-approx-cube}). When $d$ is small enough that designs of size $n>2^d$ can be considered, we suggest augmenting the $2^d$ vertices of $\frac12\,\SC_d$ with points uniformly distributed in a larger cube $\delta_1^*\SC_d$ (Section~\ref{S:n>2^d}).

Finally, recall that our incremental constructions of designs $\Xb_{1:k}$ require specifying a target maximum design size $n$. One could also adapt the construction to $k$, i.e., use $\delta_0^*(d,k,s)$ and $\delta_1^*(d,k,s)$. Further investigation is needed to assess whether this would improve performance and, if so, whether the gain would justify the increased complexity of the algorithm.  

\bibliographystyle{plain}
\bibliography{high-d-cube} 

\end{document}